\documentclass[10pt, letter, onecolumn, copyright]{arxiv}

\usepackage{kantlipsum, lipsum}
\usepackage{dm-colors}
\usepackage{amsmath}
\usepackage{pstricks, pst-node}
\usepackage{verbatim}
\usepackage{multirow}
\usepackage{scalerel}
\usepackage{booktabs}
\usepackage{enumitem}
\usepackage{xspace}
\usepackage{bm}
\usepackage{bbm}
\usepackage{mathtools}
\usepackage{soul}
\usepackage{epsfig}
\usepackage{graphicx}
\usepackage{amssymb}
\usepackage{colortbl}
\usepackage{csquotes}
\usepackage{etex}
\usepackage{setspace}
\usepackage{colortbl}
\usepackage[symbol]{footmisc}
\usepackage[bibstyle=nature,citestyle=numeric-comp,%
            natbib=true,backend=biber,maxbibnames=99,%
            giveninits=false]{biblatex}
\usepackage{nameref}
\usepackage{varioref}
\usepackage[pagebackref=false,breaklinks=false,%
            colorlinks=true,bookmarks=true,citecolor=ourdarkblue,%
            urlcolor=ourdarkblue,linkcolor=ourdarkblue]{hyperref}
\usepackage[noabbrev]{cleveref}
\usepackage{etoc}

\graphicspath{{figures/}}

\newcommand{\one}[1]{\mathbbm{1}_{[#1]}}

\def\eg{{\em e.g.,}}

\def\vs{{\em vs.~}}

\def\etal{{\em et al.}\xspace}

\title{Robust~and~Efficient~Medical Imaging~with~Self-Supervision}

\author[$\ast$, $\ddagger$, 1]{Shekoofeh Azizi}
\author[$\ast$,1]{Laura Culp}
\author[$\ast$,1]{Jan Freyberg}
\author[1]{Basil Mustafa}
\author[1]{Sebastien Baur}
\author[1]{Simon Kornblith}
\author[1]{\\Ting Chen}
\author[1]{Patricia MacWilliams}
\author[1]{S. Sara Mahdavi}
\author[1]{Ellery Wulczyn}
\author[1]{Boris Babenko}
\author[1]{Megan Wilson}
\author[1]{\\Aaron Loh}
\author[1]{Po-Hsuan Cameron Chen}
\author[1]{Yuan Liu}
\author[1]{Pinal Bavishi}
\author[1]{Scott Mayer McKinney}
\author[1]{Jim Winkens}
\author[1]{\\Abhijit Guha Roy}
\author[1]{Zach Beaver}
\author[2]{Fiona Ryan}
\author[1]{Justin Krogue}
\author[3]{Mozziyar Etemadi}
\author[1]{Umesh Telang}
\author[1]{\\Yun Liu}
\author[1]{Lily Peng}
\author[1]{Greg S. Corrado}
\author[1]{Dale R. Webster}
\author[1]{David Fleet}
\author[1]{Geoffrey Hinton}
\author[$\dagger$,1]{\\Neil Houlsby}
\author[$\dagger$, $\ddagger$, 1]{Alan Karthikesalingam}
\author[$\dagger$,1]{Mohammad Norouzi}
\author[$\dagger$, $\ddagger$, 1]{Vivek Natarajan}

\affil[1]{Google Research, }
\affil[2]{Georgia Institute of Technology, }
\affil[3]{Northwestern University}

\renewcommand{\correspondingauthor}[1]{$\ast$~Equal contributions. %
                                       $\dagger$~Equal advising. \\%
                                       $\ddagger$~Corresponding author(s): \{shekazizi, alankarthi, natviv\}@google.com }

\begin{abstract}

Recent progress in Medical Artificial Intelligence (AI) has delivered systems that can reach clinical expert level performance. However, such systems tend to demonstrate sub-optimal ``out-of-distribution'' performance when evaluated in clinical settings different from the training environment. A common mitigation strategy is to develop separate systems for each clinical setting using site-specific data~\cite{finlayson2020clinician}. However, this quickly becomes impractical as medical data is time-consuming to acquire and expensive to annotate~\cite{willemink2020preparing}. Thus, the problem of ``data-efficient generalization'' presents an ongoing difficulty for Medical AI development. \textcolor{black}{Although progress in representation learning shows promise, their benefits have not been rigorously studied, specifically for out-of-distribution settings. To meet these challenges, we present \textit{REMEDIS}, a unified representation learning strategy to improve robustness and data-efficiency of medical imaging AI. \textit{REMEDIS} uses a generic combination of large-scale supervised transfer learning with self-supervised learning and requires little task-specific customization. We study a diverse range of medical imaging tasks and simulate three realistic application scenarios using retrospective data. \textit{REMEDIS} exhibits significantly improved in-distribution performance with up to 11.5\% relative improvement in diagnostic accuracy over a strong supervised baseline. More importantly, our strategy leads to strong data-efficient generalization of medical imaging AI, matching strong supervised baselines using between 1\% to 33\% of retraining data across tasks. These results suggest that \textit{REMEDIS} can significantly accelerate the life-cycle of medical imaging AI development thereby presenting an important step forward for medical imaging AI to deliver broad impact.}

\end{abstract}

\begin{document}

\maketitle

\begin{refsection}

\section{Introduction}
Artificial Intelligence (\textit{AI}) methods based on deep learning~\cite{lecun2015deep} have delivered impressive results across medical imaging modalities, including but not limited to radiology~\cite{yala2019deep,wu2019deep,mckinney2020international,rajpurkar2018deep}, dermatology~\cite{esteva2017dermatologist,liu2020deep}, pathology~\cite{bera2019artificial,rakha2021current,wulczyn2021interpretable}, and ophthalmology~\cite{gulshan2016development,de2018clinically}, with AI models demonstrating the potential to match the performance of clinical experts in disease classification tasks. When deployed in healthcare systems, such task specific and custom designed AI systems promise to aid care-givers and improve health outcomes~\cite{zhou2021review}. 

\paragraph{Generalization remains a key translational challenge for medical imaging applications.}
Medical AI systems can be evaluated and deployed in either \textit{in-distribution(ID)} or \textit{out-of-distribution (OOD)} settings. In the controlled ID setting, evaluation of an AI model is performed in a dataset similar to the one in which it was trained, while in the OOD setting the model is evaluated or deployed in a new clinical environment that differs from the training data.  AI models frequently exhibit excellent performance in ID settings, but fail to maintain this expert-level and therefore clinically applicable performance in OOD settings. Distribution shifts are common in deployment and can have far-reaching consequences, with model accuracy shown to degrade in previously-unseen environments in multiple different applications for AI in medical imaging~\cite{condon2021replication,zech2018variable,zhang2021empirical}. Furthermore, model performance and calibration may degrade to a greater extent in underrepresented subgroups, which can propagate existing health disparities~\cite{seyyed2020chexclusion,kadambi2021achieving,pierson2021algorithmic}. The ability of medical AI models to maintain performance by efficiently generalizing to clinical settings not seen during training is necessary for their safe and effective deployment at scale~\cite{AIinHealthcare, kelly2019key,roberts2021common,van2021artificial,freeman2021use}. Rigorous evaluation of medical AI models, therefore, requires assessment of their performance in OOD settings, to guard against ``under-specification'' resulting in unanticipated poor performance during deployment~\cite{d2020underspecification}.

\begin{figure*}[t]
\small
    \centering
    \includegraphics[width=1.0\textwidth]{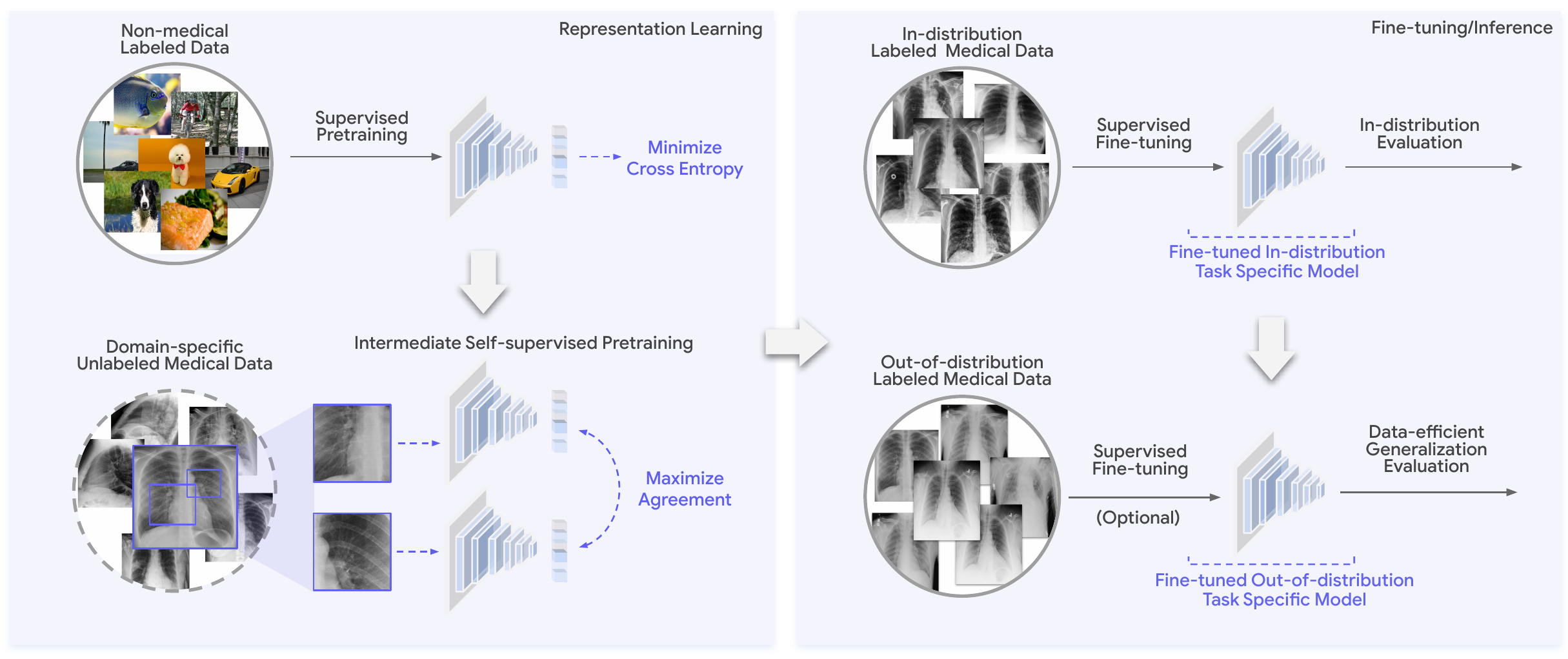}
    \vspace{0pt}
    \caption{\textbf{Overview of our proposed approach for developing robust and efficient medical imaging AI.} REMEDIS starts with representations initialized using large-scale natural image pretraining following the Big Transfer (BiT) method~\cite{kolesnikov2019big}. We then adapt the model to the medical domain using intermediate contrastive self-supervised learning without using any labeled medical data. Finally, we fine-tune the model to specific downstream medical imaging AI tasks. We evaluate the AI model both in an in-distribution (ID) setting and in an out-of-distribution (OOD) setting to establish the data-efficient generalization performance of the model.}
    \vspace{-0pt}
    \label{fig:method-overview}
\end{figure*}

\paragraph{Data-efficient generalization.}
Despite this unmet need for medical AI generalization, few practical solutions currently exist. Adapting to new clinical deployment settings by retraining or developing the AI model from scratch with data from the new distribution is perhaps the most favored approach~\cite{condon2021replication,finlayson2020clinician,futoma2020myth}. However, this may be prohibitively expensive or impractical, due to the requirement for acquiring and annotating large volumes of medical data for each new type of distribution shift, for example, the use of new imaging equipment or deployment in a new clinic~\cite{willemink2020preparing}. In turn, this significantly slows down the life-cycle of medical imaging AI development and deployment and presents an important barrier to their widespread adoption.

To formulate this problem more concretely, we utilize the notion of \textit{data-efficient generalization}, capturing the AI model's ability to generalize to new deployment distributions with significantly reduced need for expert annotated data from the new clinical setting. This is measured as: (1) improvement in zero-shot generalization to OOD settings (assessing performance in an OOD evaluation set, with zero access to training data from the OOD dataset), and (2) significant reduction in the need for annotated data from the OOD settings to reach performance equivalent to clinical experts (or threshold demonstrating clinical utility); while maintaining or improving in-distribution performance.

\paragraph{The advent of self-supervision for developing AI systems.}
The desire to reduce the reliance on hard-to-acquire labeled data for developing AI systems~\cite{li2006one,zhu2003semi} and improving their OOD generalization~\cite{cohn1994improving,sutton1996generalization} has represented a long-standing challenge for the wider AI community. The recent development and use of self-supervised learning techniques in diverse applications across computer vision~\cite{doersch2015unsupervised,doersch2017multi,gidaris2018unsupervised,pathak2016context,larsson2017colorization}, natural language understanding~\cite{devlin2018bert,brown2020language}, and speech recognition~\cite{baevski2019effectiveness} indicates its broad effectiveness towards tackling this challenge. These methods use various pretext tasks to train models to produce high-quality representations without using any label information. Contrastive self-supervised learning, in particular, has emerged as a strong approach in computer vision where models are trained by aligning representations of multiple views of the same instance created via data augmentation or other means~\cite{chen2019self,he2019momentum,he2020momentum,grill2020bootstrap,chen2020simple}. On the popular and competitive natural images benchmark, ImageNet~\cite{deng2009imagenet}, models developed using such contrastive self-supervised methods are starting to approach the performance of fully supervised methods~\cite{he2016deep,touvron2021training}. Furthermore, contrastive self-supervised learning provides additional benefits like improved robustness~\cite{liu2020hybrid} and OOD detection performance~\cite{winkens2020contrastive,shen2022connect,haochen2022beyond}. \textcolor{black}{Although progress in self-supervised learning shows promise for medical AI applications, the effect has not been rigorously studied for data-efficient generalization to out-of-distribution clinical settings. Most prior works often only evaluate on a limited number of tasks and feature custom design choices which makes it difficult to broadly apply them in practice. Furthermore, it is unclear how these methods interact or can be combined with other representation learning strategies.}      

\paragraph{Self-supervision for data-efficient generalization.}
Motivated by the need for data-efficient generalization of medical imaging AI and building on top of the progress in self-supervised learning for tackling this problem, we present \emph{Robust and Efficient Medical Imaging with Self-supervision (REMEDIS)}, a unified transfer learning strategy for developing robust medical imaging AI with minimal customization across multiple domains. We present robust evidence of the efficacy of our approach across multiple clinical tasks and settings. The key insight of this strategy is to learn transferable and generalizable visual representations that can be further fine-tuned and deployed for the downstream medical imaging task using limited labeled data from the clinical deployment setting. Unlike current standard transfer learning strategies which rely on standard supervised representation and modality or task-specific design choices, REMEDIS benefits from both large scale supervised representation and task specific self-supervised representations in a unified framework and with minimal customization across domains.  

The standard approach for learning transferable representations in computer vision involves supervised pretraining~\cite{huh2016makes,kolesnikov2019big} on large-scale natural image datasets that range from a million images in ImageNet dataset~\cite{deng2009imagenet} to several hundred million~\cite{sun2017revisiting} and beyond~\cite{mahajan2018exploring}. While these representations demonstrate strong transfer learning performance on downstream natural image tasks~\cite{zhai2019visual} and can even perform well on medical imaging tasks~\cite{mustafa2021supervised}, they tend to be sub-optimal for the medical imaging domain given the large distribution shift from natural images~\cite{raghu2019transfusion}. To alleviate this discrepancy, further supervised pretraining using medical data could be employed, but, as discussed previously, this is challenging as acquiring annotations for medical data is expensive and time-consuming. Self-supervised learning, however, does not require annotations and unlabeled medical data is often more easily available (although a considerable effort is still needed to acquire, clean, pre-process and harmonize the data~\cite{willemink2020preparing}).

Following this reasoning, our method leverages a combination of both large-scale supervised pretraining on natural images~\cite{kolesnikov2019big} as well as intermediate contrastive self-supervised learning~\cite{chen2020simple} on unlabeled domain-specific medical data to learn transferable and generalizable representations for medical images~\cite{hendrycks2019using}. These representations can be used for developing medical imaging AI by fine-tuning them on task-specific labeled medical data. The proposed strategy introduce minimal changes to the standard transfer learning workflow while drastically improving representation by wisely utilizing the pool of available labeled and unlabeled data. Figure~\ref{fig:method-overview} shows the overview of our proposed strategy for developing medical imaging AI systems that exhibit strong data-efficient generalization.

\begin{figure*}[!t]
\small
    \centering
    \includegraphics[width=1.0\textwidth]{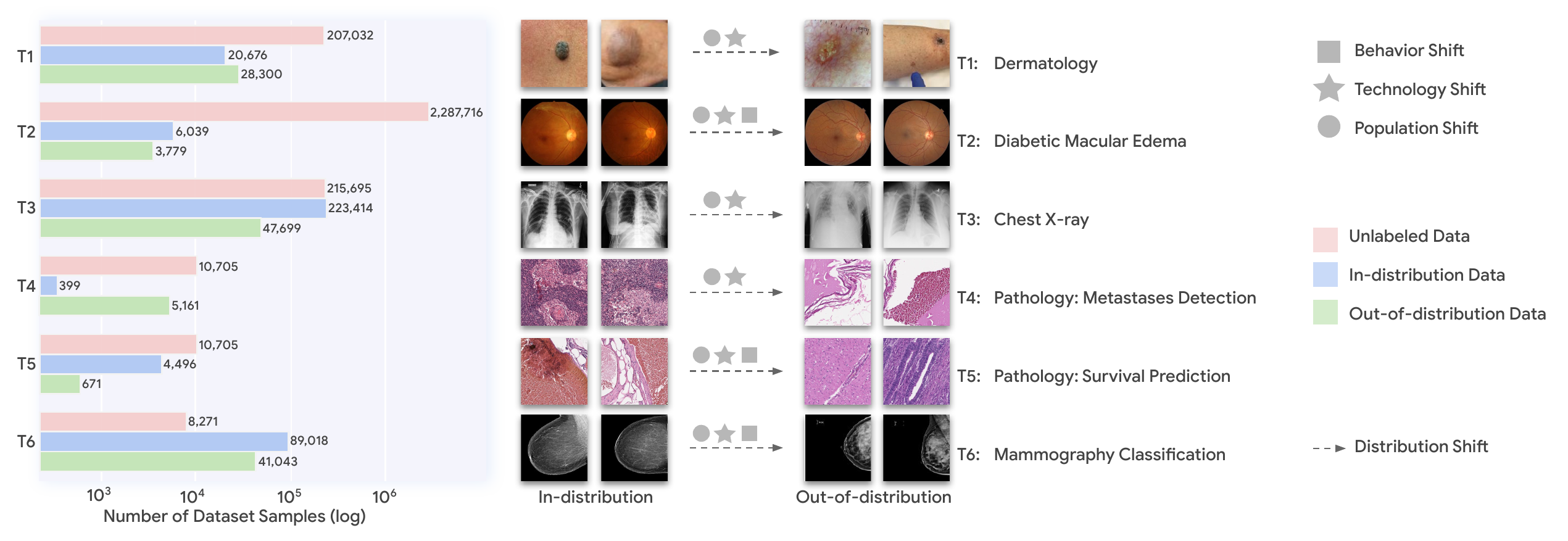}
    \vspace{+6pt}
    \caption{\textbf{Overview of clinical settings for evaluating data-efficient generalization of medical imaging AI.} We evaluate our self-supervision based representation learning method REMEDIS as well as baseline AI models on five different domains, containing six tasks and encounter a wide and complex variety of distribution shifts in these clinical settings as detailed above.}
    \vspace{-0pt}
    \label{fig:data-dist-shift}
\end{figure*}

\paragraph{Summary of key contributions.} 

The value of exploiting unlabeled data via self-supervised learning to improve in-distribution performance has been successfully studied in several individual medical imaging domains like pathology~\cite{li2021sslp,srinidhi2021improving}, dermatology~\cite{azizi2021big}, and chest X-ray~\cite{sowrirajan2021moco} interpretation. 

\textcolor{black}{However, these prior works study self-supervised learning in isolation from other representation-learning techniques, without giving consideration to how they might be combined with other representation learning techniques.
They often rely on task-specific design choices which makes it difficult to broadly apply them in practice. Furthermore, the evaluation protocols are limited and fail to comprehensive consider and evaluate data-efficient generalization.}

\textcolor{black}{For example,  Azizi~\etal~\cite{azizi2021big} propose a custom approach to contrastive self-supervised learning by leveraging multi-view images to create natural augmentation pairs. However, their approach is not generally applicable and evaluation is limited to two medical imaging tasks without consideration for data-efficient generalization to OOD settings. Another closely related work is~\cite{zhang2021empirical} which investigates the related topic of domain generalization in clinical settings. Once again this study is restricted to only two medical tasks and the authors use synthetic datasets for their experiments. The methods considered are more focused on learning invariant predictors by pooling together data from different domains and less on representation learning strategies for transfer learning. Furthermore, their focus is on domain generalization without taking the data-efficiency aspect into account. Perhaps, closest in spirit to our study is~\cite{zhou2019models}, where the authors propose a new self-supervised approach and demonstrate its potential across 5 3D medical imaging tasks. However, they fail to consider the interaction of self-supervised learning with other representation learning strategies nor do they rigorously evaluate their approach on data-efficient generalization.}   

\textcolor{black}{In contrast to prior works, our study is the first to propose a unified strategy that leverages large-scale supervised pretraining and intermediate self-supervised learning and can be applied across multiple medical imaging modalities without domain-specific customization. Given that a key part of our method is representation learning from unlabeled data using contrastive self-supervision, our method is particularly well-suited to the medical imaging AI setting given the challenges of expert labeling and the relative abundance of unlabeled medical images. Across six different medical imaging tasks and 15 different evaluation sets, we show that our strategy significantly improves in-distribution performance over strong supervised AI approaches, with up to 11.5\% relative improvement in diagnostic accuracy.}

We conceptualized the notion of ``data-efficient'' generalization to new clinical deployment settings as an important unmet need for medical AI, and designed rigorous evaluations benchmark to study this using retrospective data. AI models developed using REMEDIS deliver significant improvements in data-efficient generalization performance when evaluated in realistic out-of-distribution settings (new and previously unseen clinical environments). Models trained using our strategy needed only 6\% to 33\% of the amount of retraining data to match the performance of a strong supervised baseline and a widely-used supervised AI models that were provided access to all of the available training data from new clinical settings. These improvements in performance in new environments would result in savings of thousands of valuable clinician-hours that would otherwise be needed for medical data acquisition (years) and annotation (days), thereby potentially accelerating the life-cycle for medical imaging AI systems development, deployment and democratization. We also believe our study is the first to rigorously demonstrate empirically the effectiveness of contrastive self-supervised towards data-efficient OOD generalization both in natural image or medical imaging settings. Concurrent to our work,~\cite{shen2022connect,haochen2022beyond} have provided a more theoretical grounding for our observations.

Beyond the comprehensive experimental results presented, the approach and insights described here have been integrated in several of Google's medical imaging research projects such as dermatology~\cite{DermAI} and mammography~\cite{MammoAI}, among others. \textcolor{black}{In addition to open sourcing the code used for developing REMEDIS as well as other associated details, we provide a comprehensive Appendix that can be used as an independent guide for building on our results.} We hope this enables the medical AI community to replicate our study, derive further scientific value and realize positive clinical impact.


\begin{figure*}[t]
    \centering
    \includegraphics[width=1.0\textwidth]{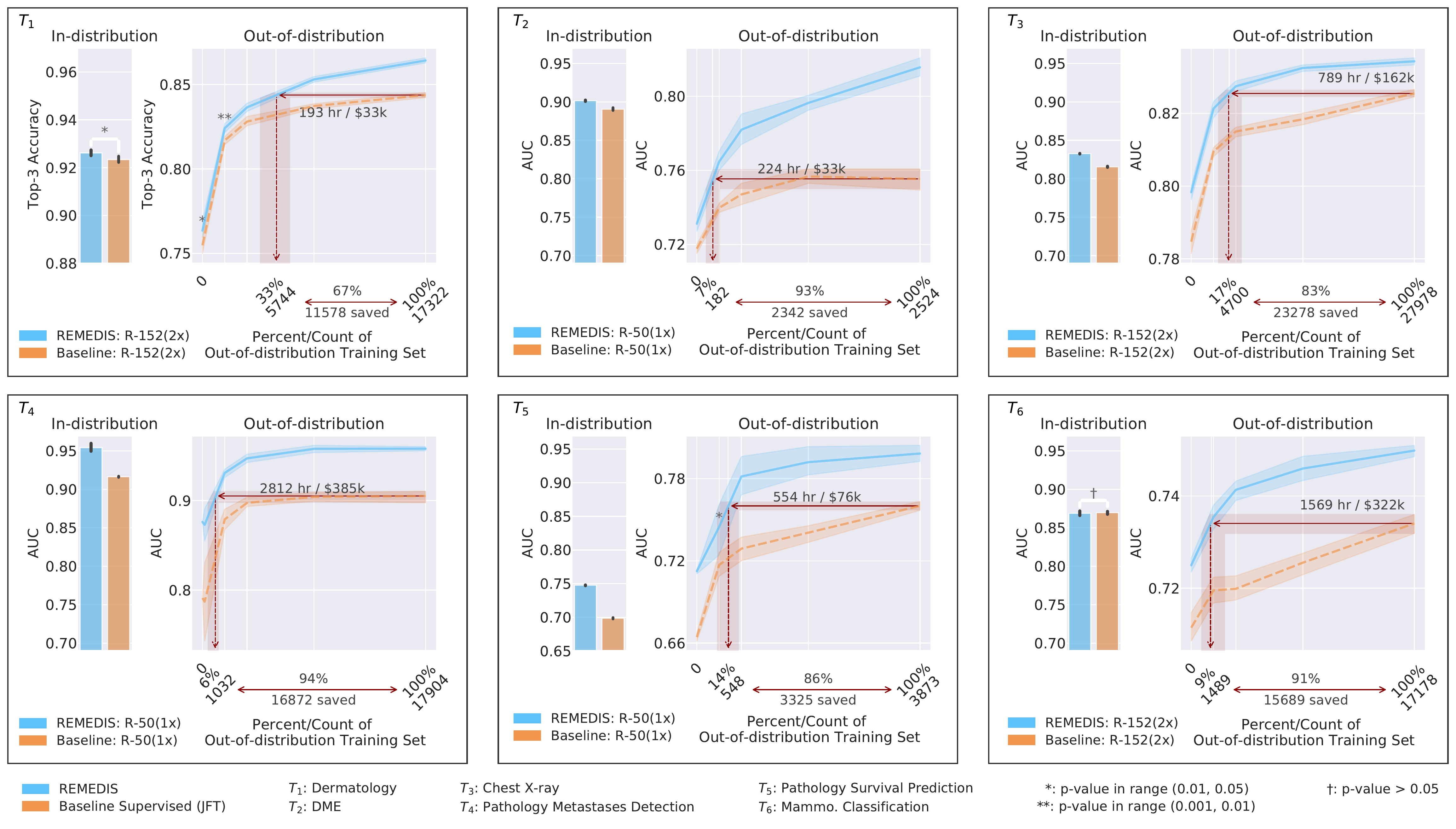}
    \vspace{-0pt}
    \caption{\textcolor{black}{\textbf{Data-efficient generalization results}}. Overview of the results demonstrating overall performance and data-efficient generalization of our proposed self-supervised learning method, REMEDIS as well as the \textcolor{black}{strong supervised baseline pretrained on JFT-300M} for the dermatology condition classification ($T_1$), diabetic macular edema classification ($T_2$), chest X-ray condition classification ($T_3$), pathology metastases detection ($T_4$), pathology colorectal survival prediction ($T_5$), and mammography classification task ($T_6$). We observed significantly improved out-of-distribution generalization and significant reduction in need for labeled medical data when using our proposed approach. 95\% confidence intervals were calculated by running each label fraction and experiment up to ten times and intervals are shown using the shaded area and error bars. A two-sided $t$-test is also done for each label fraction as well as when computing the in-distribution results. If no * is shown, the $p$-value is less than 0.001, otherwise, the $p$-value is as indicated. The red lines indicate the amount of data that REMEDIS needs to match the highest supervised AI baseline performance when simulated in a new OOD clinical deployment setting and summarize the amount of annotated data and clinician hours potentially saved by using REMEDIS for each medical task considered.}
    \vspace{-0pt}
    \label{fig:main-results}
\end{figure*}

\section{A Unified Framework for Robust Medical Imaging}
The goal of our method is to learn a predictor for each domain-specific medical task with low prediction error on both the in-distribution and the out-of-distribution data. Since it has been shown pretraining on massive unlabeled datasets potentially improves accuracy under distribution shift~\cite{hendrycks2019using,hendrycks2020pretrained}, here we focused on predictors that leverage these pretrained representations and further fine-tuned using the labeled data. In the representation learning step, we trained an encoder $f(\cdot)$ to produce representations by minimizing some loss; cross-entropy loss (a multi-class generalization of logistic loss) for supervised pretraining or a contrastive loss for self-supervised pretraining. Then we trained a classification head $g(\cdot)$ to map representations to the medical task-specific label space. The final classifier $h=g\circ f$ was composed of the encoder learnt in the representation learning step followed by the classification head. Under this setup, given labeled training examples $\{ (x_{1},y_{1}), ... ,(x_{n},y_{n})\}$ sampled from the labeled medical dataset $D$, the predictor $h(\cdot)$ maps each input example $x$ to the corresponding class label $y$. The visual representation learning procedure that we consider consisted of two pretraining steps: first, supervised pretraining using large-scale labeled natural images, then, contrastive self-supervised pretraining using unlabeled domain-specific medical images. 

\paragraph{Supervised Pretraining.} Models trained for classification using large-scale natural image datasets such as ImageNet~\cite{russakovsky2015imagenet} are commonly used for transfer learning. It has been found that using generic and large-scale supervised pretrained models can have various benefits such as speeding up training or improving downstream task performance~\cite{alzubaidi2020optimizing,graziani2019visualizing,raghu2019transfusion,mustafa2021supervised}. Big Transfer (BiT)~\cite{kolesnikov2019big} scaled this process up and in conjunction with subtle architectures changes~\cite{wu2018group} and more recent training procedures, improved the performance of transfer learning, achieving state of the art results on several downstream transfer learning tasks~\cite{zhai2019visual}. To exploit these benefits, we initialized the backbone encoder with weights from BiT models trained on the JFT~\cite{sun2017revisiting} dataset by minimizing a supervised cross-entropy loss.
Given deployment settings constrain the size of the AI model (in terms of number of model parameters) that can be used, it is important that our proposed approach works when using both small and large model architecture sizes. To study this in detail, we considered two ResNet architectures with commonly used depth and width multipliers, ResNet-50 (1$\times$) and ResNet-152 (2$\times$) as the backbone encoder networks~\cite{he2016deep}. The pretrained encoder network obtained from this step is indicated by $f_\phi(\cdot)$ and was further fine-tuned in an additional pretraining step using medical domain data.

\paragraph{Contrastive self-supervised pretraining.} To learn visual representations effectively from unlabeled medical images, we adopted SimCLR~\cite{chen2020simple,chen2020big}, a self-supervised learning algorithm based on contrastive learning. Intuitively, SimCLR learns representations by maximizing agreement~\cite{becker1992self} between differently augmented views of the same training example via a contrastive loss in a hidden layer of a feed-forward neural net.  Given a randomly sampled mini-batch of images, each image $x_i$ is augmented in two different ways, creating two views of the same example $x_{2k-1}$ and $x_{2k}$. The two images are mapped via an encoder network $f_\theta(\cdot)$ to generate representations that are transformed again with a non-linear transformation network, yielding representations $z_{2k-1}$ and $z_{2k}$ that are used for computing the contrastive loss objective. With a mini-batch of encoded examples, the contrastive loss between a {\it positive} pair of examples $i, j$ (different augmentations of the {\it same} image) is given as follows:

\vspace{6pt}
\begin{equation}
\label{eq:method_nt_xent}
    \ell^{\mathrm{NT}\text{-}\mathrm{Xent}}_{i,j} = -\log \frac{\exp(\mathrm{sim}(z_i, z_j)/\tau)}{\sum_{k=1}^{2N} \one{k \neq i}\exp(\mathrm{sim}(z_i, z_k)/\tau)}~,
\end{equation}
\vspace{6pt}

Where $\mathrm{sim}(\cdot,\cdot)$ is cosine similarity between two vectors, and $\tau$ is a scalar denoting the temperature. The pretrained encoder network obtained after this step of intermediate self-supervision is indicated by $f_\theta(\cdot)$. There were multiple hyper-parameters influencing the contrastive learning procedure including the type of optimizer, learning rate, weight decay, temperature, training epochs, batch size, selection of negative examples (i.e. $z_i$ and $z_k$) as well as data augmentation strategies and details are provided in the Methods section.

\emph{Fine-tuning and Evaluation:} The encoder network $f_\theta(\cdot)$ obtained from the self-supervised learning step was further fine-tuned using annotated medical images for the domain-specific medical task. For each task we initialized a classifier head $g(.)$ to map representations to the domain-specific label space and trained using the cross-entropy loss. The final classifier $h=g\circ f$ consisted of the encoder with the classification head. Given the importance of data-efficient generalization for medical imaging AI, we studied this in detail by considering three scenarios selected to simulate potential clinical deployment settings:

\begin{enumerate}[leftmargin=1.5em,rightmargin=0em]
    \item In-distribution fine-tuning and performance evaluation: The model $h(.)$ was evaluated on in-distribution test samples $D_{in}^{test}$.
    \item Zero-shot out-of-distribution performance evaluation: The model $h(.)$ was evaluated on out-of-distribution test samples $D_{out}^{test}$ without any further retraining / fine-tuning using OOD data.
    \item Data-efficient out-of-distribution fine-tuning and performance evaluation: The model $h(.)$ was further fine-tuned using some fraction or the whole $D_{out}^{train}$ to obtain $\bar{h}(.)$. $\bar{h}(.)$ was then evaluated on out-of-distribution test samples $D_{out}^{test}$.  
\end{enumerate}


\section{Clinical Evaluation Settings}
AI model development and evaluation were performed for six tasks spanning five medical imaging modalities (clinical dermatology photography, fundus imaging, digital pathology, chest radiography, and mammography). Predictive models in each task were built using the domain-specific unlabeled dataset $D_{u}$, the in-distribution dataset $D_{in}$, and up to two out-of-distribution datasets $D_{out}$. All of the in-distribution and out-of-distribution datasets were further split into train, validation, and test sets (with the test set simulating a deployment setting). Here we provide an overview of each task and the corresponding datasets, and they are further described in the Methods section and in Table~\ref{tab:dataset-fingerprints}.

\vspace{+3pt}
\paragraph{Task 1: Dermatology condition classification.} We compared performance to a previously-published dermatology condition classification AI~\cite{azizi2021big,liu2020deep} ($T_1$), trained to identify various types of skin conditions from digital camera images. The in-distribution dataset $D_{in}^{T_1}$ comprised 20,676 unique cases collected and de-identified by a US based tele-dermatology service. This dataset consisted of skin conditions images taken using consumer-grade digital cameras and included extensive variation in pose, lighting, camera focus, target body part, and backgrounds such as variations in clothing or environment. Each case included between one to six images after removal of cases with the occurrence of multiple skin conditions or ungradable images, with ground truth aggregated from a panel of several US-board certified dermatologists~\cite{liu2020deep}. AI models were trained to identify the most common 26 skin conditions out of 419 unique conditions in the dataset, with the remaining conditions being grouped into an additional \textit{`Other'} class. We examined model generalization using an out-of-distribution dataset, $D_{out}^{T_1}$ consisting of 28,300 unique cases collected from a separate source (Australia based skin cancer clinics), primarily focused on skin cancers and where the ground truth labels were obtained from biopsies~\cite{azizi2021big,liu2020deep}. Self-supervised pretraining leveraged a total of 207,032 unlabeled images from $D_{u}^{T_1}$ drawn from the same settings as the in-distribution dataset.

\vspace{+3pt}
\paragraph{Task 2: Diabetic Macular Edema classification.} Diabetic Macular Edema (DME) is distinguished by thickness of the central area of the retina due to the accumulation of intraretinal fluid. While it is possible to screen for DME using color fundus photographs (CFP) by detecting hard exudates near the fovea as a surrogate for the presence of fluid, extracting the thickness directly from a three-dimensional optical coherence tomography (OCT) volume has become the gold standard for making a diagnosis~\cite{virgili2015optical}. Nevertheless, the use of OCT machines for DME diagnosis world-wide is limited due to high cost~\cite{word2019vision}. In this task we followed the approach of~\cite{liu2022deep} to leverage a dataset of paired CFP and OCT data, and trained a model that takes a CFP as input, and predicts central retinal thickness (CRT) measured from the corresponding OCT. Specifically, CRT was defined as Early Treatment Diabetic Retinopathy Study (ETDRS) zone 1/central sub-field thickness (CST)~$\geq$~300{\textmu}m~\cite{brown2004detection,sadda2006automated}. \textcolor{black}{For pretraining purposes, we used the de-identified and unlabeled dataset $D_{u}^{T_2}$ from EyePACS Inc., which included 2,287,716 fundus images from 308,507 patients. Hispanic is the most prevalent race/ethnicity within this dataset population.} The in-distribution dataset $D_{in}^{T_2}$ collected in Thailand, included 6,039 CFPs from 4,035 patients. We also used a primary de-identified out-of-distribution dataset, $D_{out}^{T_2}$, to investigate the generalization of our proposed strategy under distribution shift. Unlike $D_{in}^{T_2}$, this dataset was collected in Australia and includes a total of 3,779 CFPs from 879 patients. Additionally, we used a secondary out-of-distribution dataset consisting of 909 fundus images from 323 patients collected in India for evaluating the zero-shot out-of-distribution performance evaluation.

\vspace{+3pt}
\paragraph{Task 3: Chest X-ray classification.} The chest X-ray condition classification task ($T_3$) involves multi-label classification of chest X-ray (CXR) images. Three publicly available datasets were used for training and evaluation purposes in this task: CheXpert~\cite{irvin2019chexpert}, MIMIC-CXR~\cite{johnson2019mimic}, and ChestX-ray14~\cite{wang2017chestx}. In particular, we used the combination of the training split of MIMIC-CXR~\cite{johnson2019mimic} and CheXpert~\cite{irvin2019chexpert} as $D_{u}^{T_3}$. MIMIC-CXR~\cite{johnson2019mimic} consisted of 215,695 radiographic studies collected at the Beth Israel Deaconess Medical Center in Boston, MA. Each study contained one or more views, so we sampled from these images during pretraining (preferentially sampling Posterior Anterior (PA)/Anterior Posterior (AP) views over lateral views if available). 
We use CheXpert~\cite{irvin2019chexpert} as $D_{in}^{T_3}$. This dataset was a large open source dataset of 224,316 de-identified CXRs from 65,240 unique patients. The ground truth labels for the training data were automatically extracted from radiology reports. The radiologist report was then mapped to a label space of 14 radiological findings. We predicted the five most prevalent pathologies used by~\cite{irvin2019chexpert}: atelectasis, consolidation, pulmonary edema, pleural effusion, and cardiomegaly~\cite{irvin2019chexpert}. We modelled each finding as an independent binary prediction.  Furthermore, following previous work~\cite{neyshabur2020being,raghu2019transfusion,mustafa2021supervised,azizi2021big}, in order to facilitate a robust comparison of REMEDIS to standard approaches, we defined a custom development subset of the CheXpert dataset which differed from the original set~\cite{irvin2019chexpert}. We also used ChestX-ray14 which was collected at the National Institutes of Health Clinical Center, MD, USA as $D_{out}^{T_3}$. The images in $D_{out}^{T_3}$  were annotated using a similar technique to CheXpert by extracting common findings from radiologist reports and it consisted of 47,699 CXRs.

\vspace{+3pt}
\paragraph{Task 4: Pathology metastases detection.} In the lymph node metastases detection task ($T_4$), the goal was to detect cancer metastases in digital whole-slide images of lymph node histology slides. The models were trained in a weakly-supervised manner, using only case-level labels and without any local annotations. In order to make case-level predictions, embeddings from $2^{14}$ = 16,384 patches per case were combined via an attention layer~\cite{ilse2018attention}. A random sample of 50M patches from 10,705 cases (29,018 slides) spanning 32 ``studies'' (cancer types) from The Cancer Genome Atlas (TCGA) was used for self-supervised pretraining ($D_{u}^{T_4}$). Breast lymph node slides from the CAMELYON16 challenge~\cite{bejnordi2017diagnostic} were used for model development and in-distribution evaluation ($D_{in}^{T_4}$). Lymph node slides from 5,161 stage II and III colorectal cancer cases (36,520 slides) collected between 1984-2007 from the Institute of Pathology and the BioBank at the Medical University of Graz were used for out-of-distribution evaluation ($D_{out}^{T_4}$). This dataset is further described in~\cite{wulczyn2021interpretable}, however here cases were not excluded based on having insufficient tumor content in the primary tissue slides, and we focused on the lymph node slides instead of the primary tissue.

\vspace{+3pt}
\paragraph{Task 5: Pathology colorectal survival prediction.} The objective of the colorectal cancer survival prediction task ($T_5$) was to predict 5-year disease specific survival (DSS) using digitized whole-slide images of primary colorectal tissue histology slides. Models used the same self-supervised pretraining dataset ($D_{u}^{T_4}$) as the lymph node metastases detection task ($T_4$). Colorectal tissue slides from 4,496 stage II and III colorectal cancer cases (36,841 slides) collected between 1984-2007 from the Institute of Pathology and the BioBank at the Medical University of Graz were used for model development and in-distribution validation ($D_{in}^{T_5}$). A temporal split of 671 cases (6,419 slides) collected between 2008-2013 from the same institution were used for out-of-distribution evaluation ($D_{out}^{T_5}$). This dataset is further described in~\cite{wulczyn2021interpretable}, however only cases not lost to followup for DSS within 5-years were included here.

\vspace{+3pt}
\paragraph{Task 6: Mammography classification.} In the mammography cancer classification task ($T_6$), the goal was to predict whether there will be a biopsy-confirmed cancer occurring in the 39-months following the screening episode, as described in~\cite{mckinney2020international}. We utilised multiple datasets collected in various geographic locations in this task. This included a labeled dataset collected in the UK, a labeled dataset from the US (Northwestern Memorial Hospital), an unlabeled set of images from five clusters of hospitals across five different cities in India (Bangalore, Bhubaneswar, Chennai, Hyderabad, and New Delhi), and another unlabeled set of images collected from Northwestern Memorial Hospital (Chicago, USA). Each of these datasets contained four different images per patient: medio lateral oblique (MLO) and craniocaudal (CC) views and for the left and right breasts. The UK and US datasets are described in more detail in~\cite{mckinney2020international}. The UK dataset was used as the labeled in-distribution data, ($D_{in}^{T_6}$) which included a total of 89,018 cases. The labeled dataset from the US was used as out-of-distribution data, $D_{out}^{T_6}$ which consisted of a total of 41,043 cases. For pretraining, the unlabeled dataset ($D_{u}^{T_6}$) was formed by removing labels from the labeled data from the UK dataset and combining it with the unlabeled data from India. During pretraining, as suggested by~\cite{azizi2021big,vu2021medaug} to improve the positive pair mining procedure, a single image was randomly selected from the four possible views and used to generate a positive pair with augmentation.

Put together, the medical imaging domains and tasks described above comprise a comprehensive clinical evaluation setup to rigorously evaluate the data-efficient generalization capability of REMEDIS and the supervised AI baseline.


\begin{table}[tbh!]
\centering
\caption{\textcolor{black}{\small{\textbf{Comparison of REMEDIS with supervised baselines.}}} This table compares the in-distribution and out-of-distribution performance of REMEDIS with the strong supervised baseline pretrained on JFT-300M and the standard supervised baseline pretrained on ImageNet-1K images for three evaluation scenarios. The results reported are either the average AUC or Top-3 accuracy with 95\% confidence intervals in parentheses. If no * is shown, REMEDIS significantly outperform the baseline with the $p$-value less than 0.001, otherwise * shows a $p$-value less than 0.05 and \protect\footnotemark[2] shows a non-significant improvement.}
\centering
\vspace{+6pt}
\label{tab:performance-main}
\footnotesize
\begin{tabular}{l@{\hspace{0.7em}}|@{\hspace{0.7em}}l@{\hspace{0.7em}}|@{\hspace{0.7em}}c@{\hspace{0.5em}}c@{\hspace{0.7em}}c@{\hspace{0.5em}}}
\toprule
\rowcolor{ourlightgray}
   \textbf{Tasks} &                 \textbf{Method} & \textbf{In-distribution}  &           \textbf{Out-of-dist. (0\%)} &    \textbf{Out-of-dist. (100\%)}  \\\midrule
            Task 1 &          Supervised (ImageNet) &       0.900 (0.897,0.903) &                   0.738 (0.734,0.743) &                0.839 (0.838,0.840)  \\ [3pt]
      (Top-3 Acc.) &               Supervised (JFT) &     ~0.923 (0.922,0.925)\footnotemark[1] &   ~0.755 (0.750,0.760)\footnotemark[1]& 0.844 (0.842,0.845)  \\ [3pt]  
                   &                        REMEDIS &       0.926 (0.925,0.928) &                   0.763 (0.760,0.769) &                0.864 (0.863,0.866)  \\ [3pt]\midrule
           Task 2  &          Supervised (ImageNet) &       0.887 (0.886,0.887) &                   0.685 (0.682,0.688) &                0.761 (0.759,0.764)  \\ [3pt] 
             (AUC) &               Supervised (JFT) &       0.891 (0.889,0.892) &                   0.718 (0.715,0.720) &                0.755 (0.750,0.761)  \\ [3pt] 
                   &                        REMEDIS &       0.902 (0.900,0.902) &                   0.731 (0.727,0.736) &                0.816 (0.811,0.821)  \\ [3pt]\midrule
           Task 3  &          Supervised (ImageNet) &       0.818 (0.818,0.819) &                   0.786 (0.783,0.788) &                0.812 (0.807,0.817)  \\ [3pt]
             (AUC) &               Supervised (JFT) &       0.816 (0.815,0.816) &                   0.785 (0.781,0.788) &                0.825 (0.824,0.826)  \\ [3pt]
                   &                        REMEDIS &       0.833 (0.832,0.833) &                   0.798 (0.796,0.800) &                0.835 (0.834,0.836)  \\ [3pt]\midrule
           Task 4  &          Supervised (ImageNet) &       0.856 (0.851,0.864) &                   0.757 (0.755,0.758) &                0.892 (0.886,0.895)  \\ [3pt]
             (AUC) &               Supervised (JFT) &       0.916 (0.916,0.917) &                   0.791 (0.790,0.792) &                0.905 (0.897,0.911)  \\ [3pt]             
                   &                        REMEDIS &       0.954 (0.950,0.960) &                   0.876 (0.876,0.876) &                0.958 (0.956,0.960)  \\ [3pt]\midrule
           Task 5  &          Supervised (ImageNet) &       0.714 (0.712,0.715) &                   0.649 (0.645,0.655) &                0.725 (0.719,0.729)  \\ [3pt]
             (AUC) &               Supervised (JFT) &       0.699 (0.698,0.699) &                   0.664 (0.661,0.667) &                0.760 (0.757,0.763)  \\ [3pt]             
                   &                        REMEDIS &       0.748 (0.747,0.748) &                   0.712 (0.710,0.714) &                0.798 (0.792,0.804)  \\ [3pt]\midrule
           Task 6  &          Supervised (ImageNet) &       0.852 (0.848,0.856) &                   0.700 (0.697,0.702) &                0.727 (0.725,0.728)  \\ [3pt]  
             (AUC) &               Supervised (JFT) &     ~0.869 (0.866,0.872)\footnotemark[2] &    0.711 (0.709,0.715) &                0.734 (0.732,0.736)  \\ [3pt]
                   &                        REMEDIS &       0.870 (0.868,0.872) &                   0.725 (0.724,0.726) &                0.750 (0.749,0.751)  \\ [3pt]
\bottomrule
\end{tabular}
\end{table}

\section{Experiments \& Results}

In each of these settings, we compared REMEDIS to baseline models that had utilized a standard paradigm of supervised transfer learning to demonstrate clinician-level (or otherwise clinically applicable) performance. \textcolor{black}{This includes strong supervised models that have been pretrained on JFT-300M dataset and the standard supervised models that have been pretrained on ImageNet-1K dataset and further finetuned for the specific medical imaging task.} To focus on our primary objective of improving data-efficient generalization, we tested REMEDIS and baseline AI models in previously-unseen datasets from a different clinical setting to that in which the AI system was originally trained.  

Each specific modality and task included an unlabeled pretraining dataset ($D_u$), an in-distribution dataset ($D_{in}$), and one or more out-of-distribution datasets ($D_{out}$) which reflected a variety of realistic distribution shifts due to data acquisition devices or clinical demographics~\cite{finlayson2020clinician} (Figure~\ref{fig:data-dist-shift} and Table~\ref{tab:dataset-fingerprints} has more details). Examining performance in these previously-unseen clinical settings enabled a rigorous test of model robustness to multiple distribution shift scenarios (comprising 12 large datasets with extensive variation in image size, label spaces, class distributions among others). These tasks and datasets embodied many common characteristics of medical imaging such as class label imbalance, variation of pathologies of interest from small local lesions to more global abnormalities, and other image characteristic variations.

\paragraph{Self-Supervision leads to statistically significantly improved data-efficient generalization}
\textcolor{black}{Fig.~\ref{fig:main-results}, Fig.~\ref{fig:appendix-main-results-img}, and Table~\ref{tab:performance-main} show an overview of the results demonstrating data-efficient generalization of our proposed self-supervised based representation learning method, REMEDIS as well as the strong supervised baseline pretrained on JFT-300M and standard supervised baseline pretrained on ImageNet-1K for the dermatology condition classification ($T_1$), diabetic macular edema classification ($T_2$), chest X-ray condition classification ($T_3$), pathology metastases detection ($T_4$), pathology colorectal survival prediction ($T_5$), and mammography classification task ($T_6$). REMEDIS achieves superior out-of-distribution diagnostic performance with significantly reduced requirements for labeled data from new sites compared to baselines. The supervised AI baseline models were pretrained on the either ImageNet-1K~\cite{russakovsky2015imagenet,he2016deep} or JFT-300M~\cite{sun2017revisiting} dataset followed by medical task-specific fine-tuning. More details on the supervised baselines are available in the Appendix A.}

\textcolor{black}{These results indicate that in a setup where AI model has no access to training labels from the new clinical setting, use of REMEDIS leads to a statistically significant ($p<0.05$) improvement compared to both strong and standard supervised baseline when evaluated on out-of-distribution test dataset, improving top-3 accuracy or AUC in six different and challenging medical image analysis tasks. REMEDIS exhibits significantly improved out-of-distribution performance with up to 10.7\% relative improvement in diagnostic accuracy over a strong supervised baseline and up to 15.8\% relative improvement in diagnostic accuracy over a standard supervised baseline when there is no access to retraining data in new clinical setting (0\% / zero-shot out of distribution data regime). Furthermore, the previous best performance was matched with access to 1.4\% (0.0\%, 4.9\%) to 33.2\%  (25.7\%, 39.3\%) of the labels amongst all of these task indicating achieving the same accuracy as baseline specialized models using 3-100x less data. This improved data efficiency manifests in 193 to 2,878 clinician hours saved translating to years of data collections as well as \$32K to \$394K saving in annotation cost and the total of more \$1M annotation cost saving.}

\begin{table}[t]
\centering
\caption{\textcolor{black}{\small{\textbf{Data required to achieve clinically applicable performance.}}} The table illustrates the amount of data required to achieve clinical applicable accuracy for our method (REMEDIS) and the supervised baseline in the out-of-distribution (OOD) clinical setting. (\protect\footnotemark[2]) only achieves the lower range of clinician performance reported in Table~\ref{tab:clinical-equivalent-performance} and (\protect\footnotemark[3]) shows the additional data needed to achieve the upper-bound of clinician performance (T1 achieves average clinician performance with 0\% of the data needed).}
\vspace{+6pt}
\label{tab:clinical-accuracy}
\renewcommand{\arraystretch}{1.4}
\footnotesize
\centering
\begin{tabular}{l@{\hspace{1.5em}}ccc@{\hspace{0.6em}}} 
\toprule
\rowcolor{ourlightgray} 
\textbf{Tasks}                  &\textbf{T1}\footnotemark[3]      &\textbf{T3} \footnotemark[2]   &\textbf{T5}\footnotemark[3]  \\ \midrule
\rowcolor{ourlightblue}
{REMEDIS}                       & ~~~~~~0.0\% (0)~~~~~~            & ~~~~~~4.2\% (1,179)~~~~~         & ~~~~~~1.0\% (43)~~~~~~~                \\  
\rowcolor{ourlightorange}
{Supervised  (JFT)}             & ~~~~~~0.1\% (17)~~~~~            & ~~~~~~9.4\% (2,630)~~~~~         & ~~~~~~9.9\% (380)~~~~~~               \\  
\rowcolor{ourlightgreen}
{Supervised (ImageNet)}         & ~~~~~~2.1\% (369)~~~~~           & ~~~~~~42.7\% (11,959)~~~~~       & ~~~~~~47.1\% (1,826)~~~~~            \\  

\bottomrule
\end{tabular}
\end{table}

\paragraph{Self-Supervised medical AI requires fewer labels to reach clinically applicable performance}
Obtaining an accurate measure of expert clinician performance can be challenging for several reasons. For example, in Task 2, clinicians usually diagnose DME from ocular coherence tomography (OCT) rather than the fundus image utilised by AI. Therefore a measure for clinically applicable performance was not available for all the medical tasks considered in this work (see Table~\ref{tab:clinical-equivalent-performance}). However, wherever available, we observed that self-supervised models required significantly fewer labels to reach clinically applicable performance range necessary for safe deployment (see Table ~\ref{tab:clinical-accuracy}). Moreover, we observed that in certain cases like chest X-ray interpretation, the baseline models were unable to reach clinically applicable performance even with access to all the labels from the new distribution as shown in~\ref{tab:clinical-accuracy}.

\paragraph{Self-Supervision leads to better in-distribution performance}
\textcolor{black}{When compared to the strong and standard supervised training baselines, REMEDIS not only exhibits significant improvement in out-of-distribution performance and data-efficient generalization, it also led to significant improvement in in-distribution performance in five out of six tasks as seen in Figure~\ref{fig:main-results} and Table~\ref{tab:performance-main}. In particular, for the dermatology task, the in-distribution top-3 accuracy improved modestly but statistically significant ($p<0.05$) from 0.900 (0.897, 0.903) of the standard supervised baseline and 0.923 (0.922, 0.925) of the strong supervised baseline to 0.926 (0.925, 0.928) using our strategy. For the task of predicting Diabetic Macular Edema from fundus images, we observed a significant improvement ($p<0.001$) in AUC from 0.891 (0.889, 0.892) of the strong supervised baseline to 0.902 (0.900,0.902). The improvements were more pronounced in the chest X-ray interpretation task with AUC improving from 0.816 (0.815, 0.816) of the strong supervised baseline to 0.833 (0.832, 0.833). Similarly, in the pathology metastases detection task, we observed a significant increase in AUC from 0.856 (0.851,0.864) of the standard supervised method and 0.916 (0.916, 0.917) of the strong supervised baseline to 0.954 (0.950, 0.960). For the pathology survival prediction task, the AUC improved from 0.699 (0.698, 0.699) of the strong supervised baseline to 0.748 using REMEDIS (0.747, 0.748). Finally for the mammography classification task, we observed an improvement in AUC from 0.869 (0.866,0.872) to 0.870 (0.868, 0.872) when compared to the strong supervised baseline, however this is not falling in significant improvement range. However, when compared to the standard supervised baseline we observe a significant improvement with ($p<0.001$) using REMEDIS.}

The fact that REMEDIS improved not only the OOD generalization performance but also led to improvements in in-distribution performance suggests that the benefits to OOD generalization do not come up at the expense of in-distribution and that the learnt representations are stronger across the board.

\paragraph{Clinical impact analysis}
As detailed above, for each of the different tasks, REMEDIS led to a significant reduction in the amount of labeled data required for new deployment sites. This could significantly impact the feasibility of site-specific adaptation/retraining of a model, by resolving several significant concerns - the speed of acquiring local labels (which may take years to manifest if based on outcomes), time and cost for curating, cleaning and de-identifying data for research, the machine learning software infrastructure and skills costs for model retraining (without incurring negative transfer phenomena~\cite{wang2019characterizing} that worsen model performance in unexpected ways), and significant savings of valuable clinician hours spent on annotating data. To give a more concrete example in the survival prediction task, collection of 645 examples takes over 5 years, or low incident of positive cases in tasks such as mammography screening significantly slows down the process of data collection to an extent where a proper dataset is collected over 20 years. Therefore, the saved clinician hours can be translated to saving multiple years that it takes to properly recruit patients, collect and curate a medical dataset. Across tasks, we estimate that self-supervision leads to savings of over 5000 clinician annotation-hours alone. This is likely a lower bound since the data is often labeled multiple times by different clinicians for improved label quality (typically 3-10), with further expense and time required for definition and monitoring of labeling practices, and  consensus or adjudication approaches between annotators. \textcolor{black}{As a result, we estimate that for a model that might require \$1M to adapt to a new site, REMEDIS could deliver a significant cost reduction $>50\%$ (Table~\ref{tab:method-clinical-cost})}. 

\section{Discussion}

\paragraph{Related work} 
The problem of AI generalization is long-standing and several modeling approaches have been proposed to tackle it. A recent benchmark paper~\cite{gulrajani2020search} suggested that under careful evaluation settings, AI models developed using Empirical Risk Minimization (ERM)~\cite{vapnik1998statistical,zhang2021empirical,zhang2017mixup} remain a strong baseline for this problem. In recent times, with large-scale compute capabilities, attention has shifted to how large volumes of data can be better leveraged to create more robust and efficient AI~\cite{goyal2021self,brown2020language,kolesnikov2019big}. While supervised learning approaches (especially with weak or noisy labels) have been studied and have yielded good results in domains like computer vision~\cite{mahajan2018exploring,kolesnikov2019big}, self-supervised and semi-supervised approaches that can leverage unlabeled data at scale have also gained popularity.

In particular, recent results suggest that large-scale self-supervised AI models not only led to better in-distribution task performance but also generalize better to OOD settings, and have exciting few-shot or data-efficient learning capabilities~\cite{bubeck2021universal,brown2020language,ericsson2021well}. This key insight formed the basis of our exploration into leveraging large-scale pretraining and self-supervision for developing robust and efficient medical AI, which is particularly important given the safety-critical nature of the field.

While several classes of self-supervised learning methods exist, in this work, we focused on contrastive self-supervised learning methods given they are simple to implement, domain agnostic and have yielded impressive results on natural image benchmarks. There also exists several different classes of contrastive self-supervised learning strategies, but it is not established which techniques work best under which circumstances, and all yield comparable results on natural image benchmarks. It is likely that our results are not specific to SimCLR and that similar results could be obtained with other contrastive approaches such as~\cite{chen2020improved,he2020momentum,mitrovic2020representation,grill2020bootstrap,chen2021exploring}. \textcolor{black}{Effectiveness of these self-supervised learning alternatives in improving the in-distribution performance of medical AI models has been studied in previous works such as~\cite{sowrirajan2021moco,ciga2020self,vu2021medaug,taher2022caid,taher2021systematic}, however, their effectiveness in the out-of-distribution setting has not been explored.} 

More broadly, unlabeled data can also be leveraged without self-supervision, through methods such as self-training~\cite{xie2020self}. These methods require a teacher model trained solely on annotated data, and use predictions made by this model on unlabeled data to train a student model. We compared against this technique (see Figure~\ref{fig:appendix-self-training}), and observed REMEDIS to be better under our evaluation settings.

\textcolor{black}{Furthermore, to reduce reliance on clinical expert annotated datasets, weak supervision methods has been studied where the unlabeled data is annotated with higher-level, noisier labels and often in a programmatic manner ~\cite{dunnmon2020cross} by leveraging auxiliary data modalities such as text reports when available. The use of weak supervision for in-distribution medical image analysis has been explored in multiple previous studies such as~\cite{campanella2019clinical,eyuboglu2021multi,dunnmon2020cross,bakalo2019classification}, however, the effectiveness of this approach for out-of-distribution settings is yet to be studied. Furthermore, it is difficult to generalize the same weak-supervision approach to multiple domains as they make highly task-specific design choices. Nevertheless, we believe weak supervision is a promising approach and should be considered when appropriate and feasible for developing medical imaging AI using unlabeled data.}

\paragraph{Implications for the development of a universally robust and efficient medical AI} 
\textcolor{black}{While the benefit of representation learning and self-supervised learning specifically for medical imaging has been demonstrated in several prior works, these works are often task-specific~\cite{taher2022caid,srinidhi2021improving,srinidhi2022self,li2021domain,sato2022anatomy} and feature domain specific design choices.} 

\textcolor{black}{For example,~\cite{srinidhi2021improving,srinidhi2022self} explored the use of consistency training and leveraging hard-examples in a dynamic curriculum learning setting along with self-supervision. However, this study was restricted to histopathology image analysis. Similarly,~\cite{sato2022anatomy,li2021domain} propose the use of multi-view, multi-style, and anatomy aware augmentation for self-supervised learning. Once again, these studies consider a limited number of modalities and do not consider how they might be combined with other representation learning strategies.}

\textcolor{black}{It is challenging to apply these methods in practice as they rely on several task-specific design choices during pretraining and fine-tuning~\cite{wortsman2021robust,nguyen2020wide} and often a high level of expertise is required to make these models perform well. Furthermore, these studies do not consider the problem of data-efficient generalization and often only feature in-distribution evaluations.} 

\textcolor{black}{To address these challenges we introduce REMEDIS, a unified representation learning strategy that leads to improved in- and out-of-distribution performance in multiple medical imaging tasks. Our method leverages large-scale supervised pretraining and self-supervised learning without additional domain-specific modifications. The general applicability of this method is demonstrated over 6 diverse medical imaging domains using 15 different evaluation datasets simulating real world distribution shift scenarios.} 

\textcolor{black}{To assist medical imaging AI developers and researchers with using REMEDIS, we  characterize the out-of-distribution scenarios with dataset fingerprints and provide a comprehensive guide that we hope will reduce the search space for empirical design choices when developing medical imaging AI.} 

Given the simplicity of the REMEDIS, its minimal modification to the current widespread transfer learning framework, and ubiquity of unlabeled medical imaging data, we expect widespread adoption of such learning strategy for developing medical imaging AI on top of previously standard supervised transfer learning. 

We characterized the impact of the this strategy in terms of valuable clinician hours saved from acquisition and annotating data for medical AI as well as a significant reduction in cost and time for developing these medical AI solutions. The real world impact of REMEDIS is being further assessed in prospective settings in several medical imaging research projects at Google. 

\paragraph{Limitations \& Future Work}
Self-supervised learning remains a relatively new topic in AI, and successful application for a given task can be quite challenging, especially for researchers and developers with limited compute resources. We hope to reduce the barrier to using self-supervised representations for developing medical imaging AI by open-sourcing our code as well as relevant technical details to allow the wider community to replicate and build on our study.

This study was conducted using retrospective data and as such, further rigorous health economic studies are required to quantify the impact of our method, which would account for other hurdles and costs in real-world deployment and generalization of medical AI; for example integration into clinical workflows, infrastructure, and other IT considerations.

\textcolor{black}{While we have covered a large variety of tasks and medical imaging domains, there are many related learning strategies such as weak-supervision, self-training and active training that have not been considered here as they were beyond the scope and motivation of this study.
In general, when appropriate and feasible all of the labeled and unlabeled data should be leveraged to improve the performance of the medical AI models in both in- and out-of-distribution and our goal was to come up with a unified representation learning strategy that allows this. 
Furthermore, despite its strong performance across several diverse tasks and datasets, there may be tasks or domains that benefit from domain-specific design or distribution shifts not covered in our work here. Nevertheless, we believe our method, REMEDIS, can be a good starting point even for such domains and tasks.}     

Another important direction of future research is to quantify the fairness, privacy and ethical impacts of leveraging self-supervised learning on large scale data for developing medical AI. We have presented some initial subgroup analysis of our proposed method in the dermatology and mammography settings in~\ref{fig:appendix-fairness}. However, rigorous research is required to understand the implications along these key axes, design appropriate mitigation strategies and procedural ethics best practices. 

Given the interest and progress in self-supervised learning across the wider AI community, we expect rapid progress in the development of more compute-efficient self-supervised methods. Though in this work, we demonstrated self-supervised learning by focusing on a single architectural family (ResNet), this approach is in principle architecture agnostic and can be applied to other model classes like transformers~\cite{dosovitskiy2020image} as well as with other semi-supervised learning and domain adaptation methods.

Perhaps, most excitingly, we believe this work also lays the foundation for multi-modal representation learning for AI in healthcare. Health data is inherently multi-modal in nature and includes a mix of images, electronic health records, sensors, and wearable data as well as genomics data at both the individual patient as well as the hospital system level. We believe AI systems that leverage this data at scale using self-supervised learning will be the foundation of next generation learning health systems that scale world class healthcare to everyone.

\setlength\bibitemsep{0pt}
\printbibliography
\balance
\clearpage
\end{refsection}

\begin{refsection}
\clearpage

\renewcommand{\thesection}{A.\arabic{section}}
\renewcommand{\thefigure}{A.\arabic{figure}}
\renewcommand{\thetable}{A.\arabic{table}} 
\renewcommand{\theequation}{A.\arabic{equation}} 

\setcounter{figure}{0}
\setcounter{table}{0}
\setcounter{equation}{0}
\setcounter{section}{0}

\paragraph{\LARGE{Appendix A: Development and Evaluation Details}} 
\paragraph{}\noindent 

In the following text, we provide further details on the development and evaluation of our proposed method REMEDIS. Overall, our approach comprises the following steps: 
\begin{enumerate}
    \item Supervised representation learning on a large-scale dataset of labeled natural images
    \item Self-supervised contrastive representation learning on an unlabeled dataset of in-distribution medical images
    \item Supervised fine-tuning on labeled in-distribution medical images
\end{enumerate}

To rigorously evaluate the data-efficient generalization of the AI models, we further fine-tune using labeled data from the out-of-distribution setting and assess performance. Figure~\ref{fig:method-overview} shows the summary of our proposed method, REMEDIS and Figure~\ref{fig:method-protocol} summarizes the evaluation setup. In the following sections, we provide details on the experimental setups and medical imaging AI development methodology as well as the clinical evaluation setups.


\section{Experimental Setup}
\label{method:experimental-setup}
In the following section, we lay out the experimental setup and explain the design choices made in our study. This includes the pretraining procedure, the choice of base AI network architecture, data preprocessing and augmentation strategies, hyper-parameter search procedure, and fine-tuning setup.

\subsection{Contrastive pretraining}
For contrastive pretraining, we build on SimCLR, which proposes a simple approach for contrastive learning~\cite{chen2020simple} for images. We performed a disjoint hyper-parameter tuning procedure to select factors influencing the quality of the learned representation, which we measured by the model performance in the downstream tasks using the validation set of $D_{in}$. We considered various factors influencing the performance including the network architecture, the initial model pretrained on labeled natural images, the contrastive learning hyper-parameters, and the augmentation strategy used in contrastive learning. 
For each of these factors, we conducted a comprehensive hyper-parameter search as follows. First, we pretrained a model on $D_{u}$ given the target hyper-parameters, then we measured the performance of the pretrained model on the validation set of the in-distribution dataset $D_{in}$ when fine-tuned on the corresponding training set. For the selection of architecture and pretrained model ($f_\phi$), and also the augmentation strategy, we used the SimCLR default learning rate, batch size, and temperature settings as described below. 

\paragraph{Base network and pretrained models.}
In our experiments, we used a standard ResNet architecture~\cite{azizi2021big,chen2020simple,chen2020big,kolesnikov2019big,mustafa2021supervised}, with two different model architecture sizes to ensure the observed phenomenon are disentangled from the model size and number of parameters in the network. In particular, we considered two ResNet architectures with two commonly used depths and width multipliers (hidden layer widening factors) as the backbone networks: ResNet-50 (1$\times$) and ResNet-152 (2$\times$). Following the SimCLR~\cite{chen2020simple} method for contrastive pretraining, we used two fully connected layers to map the 2048-dimensional output of each ResNet to a 128-dimensional representation embedding space. We performed SimCLR pretraining on $D_{u}^{T_n}$ where $n=\{1,2, ..., 6 \}$ indicates different medical imaging tasks. 

In our proposed method, REMEDIS, we initialized the backbone ResNet architecture with weights from Big Transfer (BiT)~\cite{kolesnikov2019big} pretrained models. In addition to the model architecture, BiT models vary based on the pretraining dataset: BiT-S, BiT-M and BiT-L, where S(mall), M(edium) and L(arge) indicate if the pretraining was done on ILSVRC-2012 (ImageNet-1K)~\cite{russakovsky2015imagenet}, ImageNet-21K~\cite{deng2009imagenet} or JFT~\cite{sun2017revisiting}, respectively. The BiT-L family resulted in the best performance on each domain-specific validation set, and hence, was selected as our main backbone model. The BiT models all use a ResNet-v2 architecture~\cite{he2016identity}, which replace all Batch normalization~\cite{ioffe2015batch} layers with Group normalization~\cite{wu2018group} and use Weight standardization~\cite{qiao2019micro} in all convolutional layers. This setup differs from the standard SimCLR setup, and to incorporate BiT pretrained models into the SimCLR-based pretraining, we reused pretrained weights of convolution layers while Batch normalization was replaced with Group normalization via default initialization. 

In addition to the above setup used across our key experiments, we have included detailed ablation studies in the Supplementary Material section where we considered setups both with and without initialization from pretrained weights and ran several other experiments to understand our method and results comprehensively.

\begin{figure*}[t]
\small
    \centering
    \includegraphics[width=0.8\textwidth]{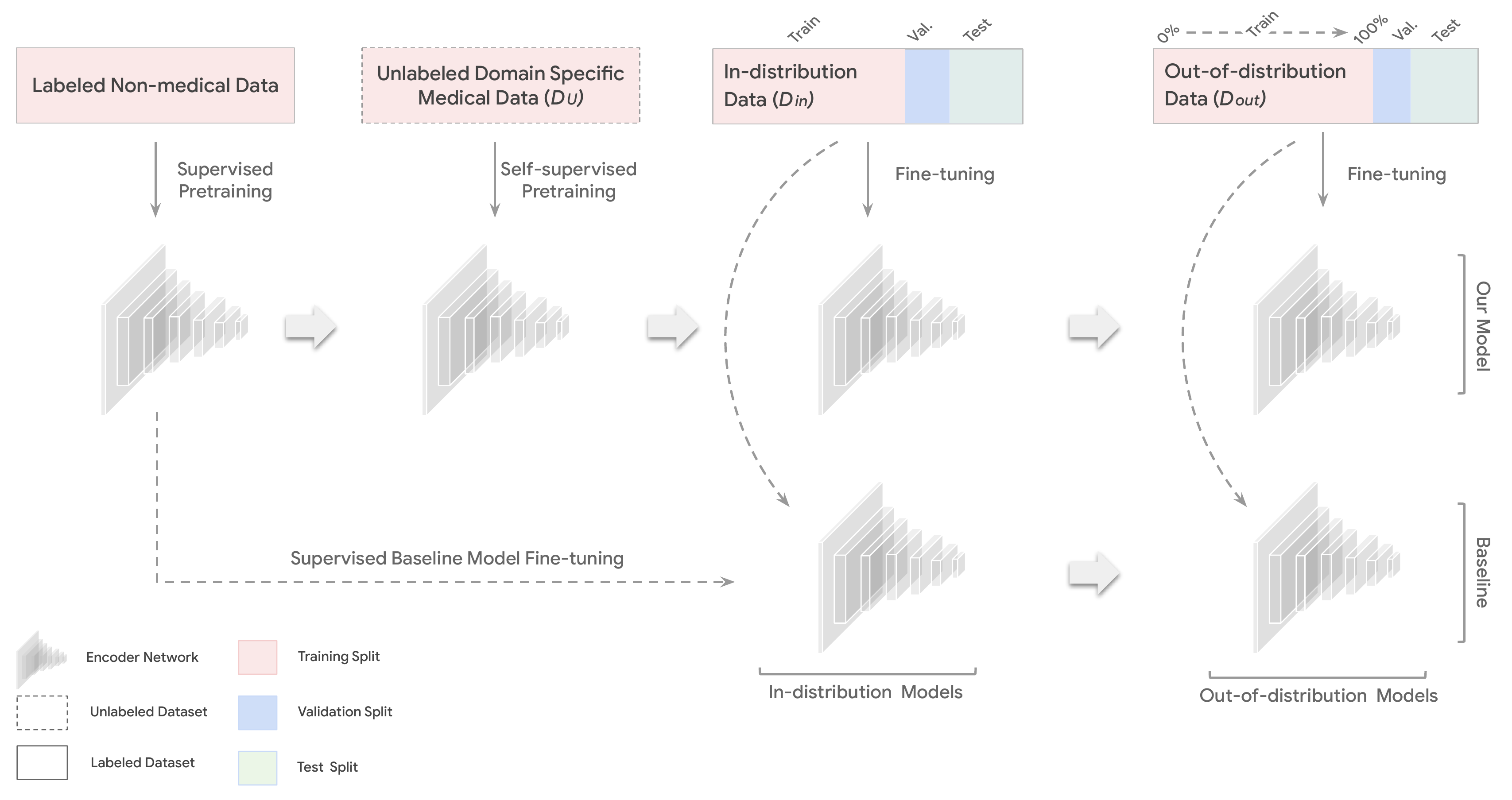}
    \vspace{6pt}
    \caption{\small{\textbf{Overview of our experimental setup for the development of REMEDIS and the baseline AI models across the various medical imaging tasks.} In particular, we detail the different stages in which unlabeled and labeled (both in-distribution and out-of-distribution) are used for model development and evaluation.}}
    \vspace{-0pt}
    \label{fig:method-protocol}
\end{figure*}

\paragraph{Pretraining hyper-parameters.}
The contrastive learning procedure is influenced by multiple hyper-parameters, including the choice of optimizer, learning rate, weight decay, temperature, training epochs, and batch size. We based our training on SimCLR~\cite{chen2020simple}, and used the LARS optimizer~\cite{you2017large} to stabilize training during pretraining as suggested by~\cite{chen2020simple}. We also used the default weight decay of $10^{-6}$ and trained all of the models for 1000 epochs. For each task, we investigated pretraining models with learning rate (${lr}$) in $\{0.1, 0.3\}$ and temperature ($\tau$) in $\{0.1, 0.2\}$, and also with the largest possible batch size in $\{1024, 2048, 4096\}$ that is compatible with memory constraints of our hardware infrastructure. Our experiments showed that in all tasks, 1000 epochs using a learning rate of 0.3 and temperature 0.1 led to a domain-specific pretrained model with optimal performance. Since contrastive pretraining often leads to spiky learning curves, after pretraining the model for 1000 epochs, we selected our final checkpoint by considering all of the checkpoints in a window of $\left[M(1-0.001M), M\right]$ where $M$ is the max iteration, and picking the checkpoint with the minimum contrastive loss. Table~\ref{tab:pretraining-param} shows the final selected hyper-parameter for each task based on the performance of the models on the validation set of the downstream task. 

\begin{table*}[t]
\centering
\caption{\textbf{Pretraining hyper-parameter details.} We pre-trained the models using the following hyperparameter ranges for self-supervised learning with learning rate (${lr}$) in $\{0.1, 0.3\}$, temperature ($\tau$) in $\{0.1, 0.2\}$, and batch size ($B$) in $\{1024, 2048, 4096\}$. We utilized random cropping (C), random color distortion (D), rotation (R), random Gaussian blur (G), histogram equalization (H), and elastic deformation (E) as the data augmentation strategies. We use LARS optimizer~\cite{you2017large} and our experiments suggest that in all tasks, pretraining for 1000 epochs using a ${lr}=0.3$ and $\tau=0.1$ tends to lead to optimal performance.}
\vspace{6pt}
\label{tab:pretraining-param}
\footnotesize
\begin{tabular}{c|c|c|c|c}
\toprule
\rowcolor{ourlightgray} 
\textbf{Tasks}&          \textbf{Augmentations}& \textbf{Max Iteration ($M$)}&   \textbf{Batch Size ($B$)}&   \textbf{Architectures}  \\ \midrule
$T_1$&               C, D, R, G&                        202K&               1024&    ResNet-50 (1$\times$) and ResNet-152 (2$\times$)  \\[3pt]
$T_2$&               C, D, R, G&                      2,229K&               1024&    ResNet-50 (1$\times$) and ResNet-152 (2$\times$)  \\[3pt]
$T_3$&            C, D, R, G, H&                        210K&               1024&    ResNet-50 (1$\times$) and ResNet-152 (2$\times$)  \\[3pt]
$T_4$~\&~$T_5$&      C, D, R, G&                      1,220K&               4096&    ResNet-50 (1$\times$) Only                        \\[3pt]
$T_6$&         C, D, R, G, H, E&                        302K&               1024&    ResNet-50 (1$\times$) and ResNet-152 (2$\times$)  \\[3pt]
\bottomrule
\end{tabular}
\end{table*}

\paragraph{Pretraining data augmentations details.}
In our default contrastive pretraining setting, we utilized random cropping (C), random color distortion (D), rotation (R), and random Gaussian blur (G) as the data augmentation strategy. 

Due to the grayscale nature of radiology images (i.e. mammography and chest x-ray images), for these images we opted for stronger data augmentation to reduce the chances of overfitting. We further improved the final performance by incorporating histogram equalization~\cite{castro2018elastic} and elastic deformation~\cite{castro2018elastic,ronneberger2015u} in addition to our default data augmentation strategy. For these grayscale images, histogram equalization (H) corrects the contrast, ensuring that all further contrast changes start from a uniform place and makes color distortion work better. Furthermore, elastic deformation (E) can realistically occur during breast cancer screening and in general for most human organs during imaging when undergoing any internal or external pressure~\cite{castro2018elastic}. By including these augmentation strategy, the network learns invariance to such deformations, without the need to see these transformations in the image corpus.  

As studied in~\cite{azizi2021big,chen2019self}, we used a standard Inception-style random cropping procedure explained in~\cite{szegedy2015going} as one of the fundamental preprocessing steps for contrastive learning. In all pretraining experiments, images were randomly cropped and resized to 224$\times$224 pixels. This image size is mainly used for pretraining due to memory constraints of the underlying hardware to help increase the mini-batch size. Previous studies~\cite{azizi2021big} suggest pretraining with 224$\times$224 images does not have a substantial impact on the final performance. 

In the pathology tasks, to capture details specifically present in high-resolution pathology slides, we obtained patches from various magnification levels. Additionally, this was followed by a random horizontal left-to-right flip with a 50\% probability. We applied random rotation by angle $\delta\sim~U(-45^{\circ}, 45^{\circ} )$ and random color distortion with maximum strength $1.0$ that included random brightness, contrast, saturation, and hue changes. We blurred the image 50\% of the time using a Gaussian kernel with $\sigma\in[0.1, 2.0]$ and size of 10\% of the image height and width. Additional details about the selection of batch size, learning rate, and augmentations are provided in Table~\ref{tab:pretraining-param}.

\subsection{Fine-tuning and Evaluation Protocol}
Prior works in transfer learning and semi-supervised learning primarily use linear probing for evaluation of the learned representations~\cite{wortsman2021robust,tripuraneni2020theory,du2020few}. End to end fine-tuning is not commonly considered given the computation costs incurred during evaluation. 

In general, obtaining an effective fine-tuned model requires an in-depth understanding of training dynamics and goes beyond solely minimizing the loss function using the labeled data. Fine-tuning and training from scratch both optimize the same training loss but differ in their initial weights. However, the selection of learning hyper-parameters, various output layer/head options, and different gradient flow choices can dramatically affect the performance of the final model. To this end, we considered a detailed fine-tuning and evaluation protocol as explained below with a visual overview in Figure~\ref{fig:method-protocol}.

\paragraph{Fine-tuning hyper-parameters.} After obtaining the pretrained model for each task, we further fine-tuned the model end-to-end using the training set of the labeled in-distribution medical data. Following the approach described by~\cite{azizi2021big,chen2020simple,chen2020big}, we initialized the weights of the network used for the downstream task from the weights of the pretrained network obtained in the previous step. For every combination of pretraining strategy and downstream task, an extensive hyper-parameter search was performed. This included the choice of optimizer, learning rate, weight decay, decay steps, and decay factor. In particular, as recommended in much previous research, we tried both the Adam~\cite{kingma2014adam} optimizer or standard stochastic gradient descent (SGD) with Nesterov Momentum~\cite{loshchilov2016sgdr,goyal2017accurate} utilizing a linear or decaying learning rate. For all tasks, we also performed data augmentation and preprocessing to achieve the best performance after downstream fine-tuning. However, unlike the pretraining step, we did not study the effect of each data augmentation separately in the fine-tuning step and mainly followed the standard augmentation that has been used in prior published work for these tasks~\citep{liu2020deep, mckinney2020international,wulczyn2021interpretable}.   

We performed model selection based on the model performance in the validation set of the in-distribution labeled medical data, and the final performance was reported using the test set of the labeled in-distribution data. We also followed a standard early stopping procedure to avoid overfitting~\cite{bengio2017deep}. To estimate the variability around the model performance, and investigate any statistically significant improvement, the chosen hyper-parameters were used for the training and testing of ten model runs (unless otherwise specified), and task performance was reported based on mean and standard deviations of the performance across these model runs. Details of the fine-tuning procedure for each task are explained in the following paragraphs and summarized in Table~\ref{tab:finetuning-param}.

\emph{Dermatology condition classification:}  For dermatology condition classification, we used the Adam optimizer with a linear learning rate and we fine-tuned all models for 150K steps with a batch size of 16. During fine-tuning, we augmented the dermatology images by performing random color distortion, crops with resizing to 448$\times$448 pixels, blurring, rotation, and random flipping. We selected the learning rate and weight decay after a grid search of seven logarithmically spaced learning rates between 10$^{-6.0}$ and 10$^{-3.0}$ and three logarithmically spaced values of weight decay between 10$^{-6.0}$ and 10$^{-4.0}$, as well as no weight decay.

\emph{Diabetic Macular Edema classification:}  In the DME classification task, we used an SGD optimizer with momentum parameter 0.9 and an exponential learning rate schedule, and we fine-tuned all models for a maximum of 1000 steps with batch size 8. The final checkpoint selection was based on the performance of the model on the validation set. During the training, we used an augmentation strategy consisting random color distortion, crops with resizing to 587$\times$587 pixels, blurring, and random flipping. We tuned the initial learning rate and weight decay by performing a grid search of seven logarithmically spaced learning rates in 10$^{-4.0}$ and 10$^{-0.5}$ and three logarithmically spaced values of weight decay between 10$^{-5.0}$ and 10$^{-3.0}$, as well as no weight decay. We set the decay rate as 0.1 of the initial learning rate. 

\vspace{2pt}
\emph{Chest X-ray classification:}  We used the Adam optimizer with an exponential learning rate decay for the Chest X-ray classification task. We trained all models in this task up to a maximum of 250K steps using a batch size 64. We pre-processed and augmented the chest X-ray images by applying random cropping, scaling to 224$\times$224 pixels, rotation up to 15$^{\circ}$, and random color distortions. We selected the initial learning rate and weight decay after a grid search of seven logarithmically spaced learning rates in 10$^{-8.0}$ and 10$^{-2.0}$ and three logarithmically spaced values of weight decay between 10$^{-6.0}$ and 10$^{-4.0}$, as well as no weight decay.

\emph{Pathology tasks:}  For both pathology tasks, we used the Adam optimizer with a linear learning rate, and we fine-tuned all models for a maximum of 25,000 steps with batch size 8. We tuned the learning rate using a grid search of five logarithmically spaced learning rates in 10$^{-7.0}$ and 10$^{-3.0}$, and we used no weight decay.

\emph{Mammography classification:}  For the breast cancer classification task, we used the SGD optimizer with a momentum parameter of 0.9 and exponential learning rate schedule, and we fine-tuned all models for a maximum of 100,000 steps with a batch size of 1. During training we used an augmentation strategy consisting of random color distortion, resizing to 2048$\times$2048 pixels, random flipping, and elastic deformation. We performed a grid search of five logarithmically spaced initial learning rates between 10$^{-4.0}$ and 10$^{-2.0}$ and three logarithmically spaced values of weight decay between 10$^{-6.0}$ and 10$^{-4.0}$, as well as no weight decay and considered learning rate decay steps in $\{10K, 25K\}$.

\begin{table*}[tbh!]
\centering
\caption{\textbf{Fine-tuning hyper-parameter search details.} In the fine-tuning step, we investigated learning rate (${lr}$), weight decay ($w$), the choice of optimizer and the decay step. In each case, we performed a grid search of logarithmically spaced samples for learning rate and weight decay and performed model selection based on the performance on the validation set in both in-distribution and out-of-distribution settings.}
\vspace{+3pt}
\label{tab:finetuning-param}
\scriptsize
\begin{tabular}{c|c|c|c|c|c}
\toprule
\rowcolor{ourlightgray} 
\textbf{Tasks}&    \textbf{Optimizer}&                          \textbf{Learning rate (${lr}$)} &                              \textbf{Weight decay ($w$)}&          M\textbf{ax steps ($M$)}&    \textbf{Decay step}     \\ \midrule
$T_1$&                   Adam  Linear&        7 samples $\in$ $[10^{-6.0},10^{-4.0}]$  &        $\{0, 10^{-6.0}, 10^{-5.0}, 10^{-4.0}\}$&                    150K &       N/A          \\ [3pt] 
$T_2$&                SGD Exponential&        8 samples $\in$ $[10^{-4.0},10^{-0.5}]$  &        $\{0, 10^{-5.0}, 10^{-4.0}, 10^{-3.0}\}$&                      1K &   $\{50, 100\}$    \\ [3pt]
$T_3$&                SGD Exponential&        5 samples $\in$ $[10^{-5.0},10^{-2.0}]$  &                   $\{0, 10^{-6.0}, 10^{-5.0}\}$&                    250K &   $\{10K, 25K\}$   \\ [3pt]
$T_4$~\&~$T_5$&           Adam Linear&        4 samples $\in$ $[10^{-7.0},10^{-4.0}]$  &                                             N/A&                    25K  &       N/A          \\ [3pt]
$T_6$&                SGD Exponential&        5 samples $\in$ $[10^{-4.0},10^{-2.0}]$  &        $\{0, 10^{-6.0}, 10^{-5.0}, 10^{-4.0}\}$&                    100K &   $\{10K, 25K\}$   \\ 
\bottomrule
\end{tabular}
\end{table*}

\begin{table*}[t]
\centering
\caption{\textcolor{black}{\textbf{Fine-tuning hyper-parameter details.}} In our hyper-parameter search, we investigated the  choice of optimizer, learning rate, weight decay, decay step, and the network architecture. The table summarizes the selected hyper-parameters for fine-tuning REMEDIS and both strong and standard supervised baseline models pretrained on JFT-300M and ImageNet-1K dataset for both the in-distribution and out-of-distribution settings. }
\vspace{+3pt}
\label{tab:final-hyperparams}
\renewcommand{\arraystretch}{1.25}
\scriptsize
\begin{tabular}{@{\hspace{0.2em}}l|l|cccccc}
\toprule
\rowcolor{ourlightgray}
\multicolumn{2}{l}{Tasks}                             &$T_1$                  &$T_2$                   &$T_3$                 &$T_4$                    &$T_5$                 &$T_6$              \\\midrule
\multirow{3}{*}{Data}           & Input size          &448$\times$448         &587$\times$587          &224$\times$224        &224$\times$224           &224$\times$224        &2048$\times$2048   \\
                                & Batch size          &16                     &8                       &64                    &8                        &8                     &1                  \\
                                & Shuffle buffer      &256                    &64                      &256                   &N/A                      &N/A                   &256                \\\midrule
\multirow{3}{*}{Optimization}   & Optimizer           &Adam                   &SGD                     &Adam                  &Adam                     &Adam                  &SGD                \\
                                & Schedule            &Linear                 &Exponential             &Exponential           &Linear                   &Linear                &Exponential        \\ 
                                & Max training        &150K                   &100K                    &250K                  &25K                      &25K                   &100K               \\\bottomrule
                                & Architecture        &R-152 (2$\times$)      &R-152 (2$\times$)      &R-152 (2$\times$)      &R-50 (1$\times$)         &R-50 (1$\times$)      &R-152 (2$\times$)  \\     
                                & Learning rate       &0.0003                 &0.3                    &0.001                  &0.0001                   &0.0001                &0.0003             \\ 
Hyper-parameters                & Decay factor        &N/A                    &0.99                   &0.9                    &N/A                      &N/A                   &0.1                \\ 
REMEDIS ($D_{in}$)              & Decay step          &N/A                    &50                     &10K                    &N/A                      &N/A                   &10K                \\ 
                                & Weight decay        &$10^{-5.0}$            &0.0001                 &$10^{-5.0}$            &N/A                      &N/A                   &0.001              \\ \midrule
                                & Architecture        &R-152 (2$\times$)      &R-152 (2$\times$)      &R-152 (2$\times$)      &R-50 (1$\times$)         &R-50 (1$\times$)      &R-152 (2$\times$)  \\ 
                                & Learning rate       &0.0001                 &0.3                    &0.001                  &0.0001                   &0.0001                &0.0001             \\ 
Hyper-parameters                & Decay factor        &N/A                    &0.99                   &0.9                    &N/A                      &N/A                   &0.1                \\ 
REMEDIS ($D_{out}$)             & Decay step          &N/A                    &100                    &10K                    &N/A                      &N/A                   &10K                \\ 
                                & Weight decay        &$10^{-5.0}$            &0                      &$10^{-5.0}$            &N/A                      &N/A                   &0.0001             \\\bottomrule
                                & Architecture        &R-152 (2$\times$)      &R-152 (2$\times$)      &R-152 (2$\times$)      &R-50 (1$\times$)         &R-50 (1$\times$)      &R-152 (2$\times$)  \\ 
                                & Learning rate       &0.0001                 &0.01                   &0.0001                 &$10^{-4.0}$              &$10^{-5.0}$           &0.01              \\ 
Hyper-parameters                & Decay factor        &N/A                    &0.99                   &0.9                    &N/A                      &N/A                   &0.1               \\ 
Strong Baseline($D_{in}$)       & Decay step          &N/A                    &50                     &10K                    &N/A                      &N/A                   &10K               \\ 
                                & Weight decay        &0.0001                 &0.0001                 &$10^{-5.0}$            &N/A                      &N/A                   &0                 \\ \midrule
                                & Architecture        &R-152 (2$\times$)      &R-152 (2$\times$)      &R-152 (2$\times$)      &R-50 (1$\times$)         &R-50 (1$\times$)      &R-152 (2$\times$)  \\ 
                                & Learning rate       &0.001                  &0.1                    &$10^{-5.0}$            &$10^{-5.0}$              &$10^{-7.0}$           &0.0001            \\ 
Hyper-parameters                & Decay factor        &N/A                    &0.99                   &0.9                    &N/A                      &N/A                   &0.1               \\ 
Strong Baseline($D_{out}$)      & Decay step          &N/A                    &100                    &N/A                    &N/A                      &N/A                   &10K               \\ 
                                & Weight decay        &$10^{-6.0}$            &0                      &$10^{-5.0}$            &N/A                      &N/A                   &0.0001            \\ \bottomrule       
                                & Architecture        &R-152 (2$\times$)      &R-152 (2$\times$)      &R-152 (2$\times$)      &R-50 (1$\times$)         &R-50 (1$\times$)      &R-152 (2$\times$)  \\ 
                                & Learning rate       &0.0001                 &0.003                  &0.001                  &$10^{-5.0}$              &$10^{-6.0}$           &0.01              \\ 
Hyper-parameters                & Decay factor        &N/A                    &0.99                   &0.9                    &N/A                      &N/A                   &0.1               \\ 
Standard Baseline($D_{in}$)     & Decay step          &N/A                    &50                     &10K                    &N/A                      &N/A                   &10K               \\ 
                                & Weight decay        &0                      &$10^{-5.0}$            &0                      &N/A                      &N/A                   &0                 \\ \midrule
                                & Architecture        &R-152 (2$\times$)      &R-152 (2$\times$)      &R-152 (2$\times$)      &R-50 (1$\times$)         &R-50 (1$\times$)      &R-152 (2$\times$)  \\ 
                                & Learning rate       &0.001                  &0.1                    &0.001                  &$10^{-5.0}$              &$10^{-6.0}$           &0.0001            \\ 
Hyper-parameters                & Decay factor        &N/A                    &0.99                   &0.9                    &N/A                      &N/A                   &0.1               \\ 
Standard Baseline($D_{out}$)    & Decay step          &N/A                    &100                    &N/A                    &N/A                      &N/A                   &10K               \\ 
                                & Weight decay        &0.001                  &0                      &$10^{-5.0}$            &N/A                      &N/A                   &0                 \\

\bottomrule
\end{tabular}
\end{table*}

\paragraph{Zero-shot out-of-distribution performance evaluation:} To evaluate the robustness of our models to distribution shifts, we considered a generalization setting where the model post pretraining and end-to-end fine-tuning on in-distribution are used to make predictions on the shifted out-of-distribution (OOD) dataset without using any labels from the new dataset. In this setting, the out-of-distribution dataset ($D_{out}$) and the in-distribution dataset ($D_{in}$) have the same label space but the out-of-distribution dataset undergoes distribution shifts due to demographic, technology, or behavioral changes. Towards this end, we directly evaluated the performance of fine-tuned models which include the classification head on the test split of $D_{out}$ for each task and report the performance based on the task metrics. Similar to the in-distribution dataset, to estimate the variability around the model performance we report the mean and standard deviations of 10 different models trained with the selected hyper-parameters in the fine-tuning step (unless otherwise specified). 

\paragraph{Data-efficient out-of-distribution fine-tuning and evaluation:} We also considered the setting in which with site-specific retraining, we can improve the performance of the model on the out-of-domain dataset. For this purpose and to investigate the data-efficiency of our models, we fine-tuned our models on different fractions of out-of-domain labeled training data. To simulate the label scarcity encountered when developing medical imaging AI, we used various fractions of labeled $D_{out}^{train}$ including $\{10\%,20\%,50\%,100\%\}$, and after fine-tuning of the model using these fractions we evaluated the models on the test split of $D_{out}^{test}$. Due to computation constraints, we performed the hyper-parameter selection based on the previously explained fine-tuning protocol only using 100\% of the $D_{out}^{train}$ and selected the hyper-parameters based on the performance of the fine-tuned models on the validation split $D_{out}^{val}$ for each task. This implies that one could reach even higher performance and data efficiency than the numbers we report for smaller data-fraction sizes by performing a more thorough hyper-parameter search. Our experiment also showed that utilizing early stopping is essential for smaller data fractions.

The amount of out-of-distribution labeled data that we use for the adaptation, the severity of the distribution shift, and the architecture of models used can affect the final performance~\cite{wang2018deep}. For these reasons, in each task, we additionally considered three distinct scenarios for initializing the weights of the network: 
\begin{enumerate}
    \item using the domain-specific pretrained model weights and adding a random classification head
    \item starting from the in-distribution fine-tuned model and also keeping the classification head obtained from the in-distribution fine-tuning step
    \item using the in-distribution fine-tuned model and adding a random classification head
\end{enumerate}

We selected the best setup based on the performance on the validation split $D_{out}^{val}$ for a given task to report our results. 

\subsection{Baselines}
\textcolor{black}{For all tasks, we considered three baselines: (1) the prevailing and standard paradigm for developing medical imaging AI using models pretrained on ImageNet-1K using supervised learning, (2) the strong supervised baseline pretrained on JFT-300M images, (3) the task-specific clinically applicable performance wherever available.} 

\paragraph{Supervised baseline:} 
\textcolor{black}{The standard supervised baseline we considered for all the medical imaging tasks in this study is an ImageNet-1K supervised pretrained ResNet. Initialization from a model pretrained on ImageNet-1K is the standard baseline for transfer learning in medical imaging as demonstrated by their use in previously published works~\cite{liu2020deep, mckinney2020international, esteva2017dermatologist, gulshan2016development} and ResNets remain a popular and competitive architecture baseline for computer vision tasks~\cite{raghu2019transfusion,he2016deep,bello2021revisiting}. We also the strong supervised baseline, which is a ResNet model pretrained on JFT-300M that has shown significant improvement over standard baseline for medical image analysis~\cite{mustafa2021supervised}.  For fine-tuning from the supervised pretrained baselines, we followed the exact same protocol as for the models developed with REMEDIS. For fair comparison we performed identical extensive hyper-parameter sweeps as REMEDIS so that the baseline models attain the highest performance on the validation set. Based on our experiments, the supervised baseline models generally require more iterations to fully converge and our chosen maximum step and early stopping mechanism ensure they attain optimal performance.} 

Furthermore, for each baseline as well REMEDIS, we consider two architecture size - ResNet-50 (1$\times$) and ResNet-152 (2$\times$) (except in the pathology domain due to computational considerations). This helps us further understand the influence of the model size and number of parameters in our proposed method as well as the baselines. In addition to the supervised baselines, it is also possible to train medical imaging models by randomly initializing the model weights from scratch and fine-tuning on the downstream task. However, prior work~\cite{mustafa2021supervised,raghu2019transfusion} suggests this is sub-optimal compared to the supervised baseline and hence we did not consider it in this study.


\paragraph{Clinically applicable performance:} We define clinically applicable performance as accuracy demonstrated by expert clinicians in the out-of-distribution setting for the given task at hand, $D_{out}$. Obtaining a measure of clinician expert performance in the various clinical settings considered in this study is challenging for several reasons (\eg~cost or time taken for annotation). As a result, we have a measure for clinically applicable performance only for some of the medical tasks considered in this work as we detail below. Table~\ref{tab:clinical-equivalent-performance} summarizes the clinically applicable performance range calculated for each task.

\begin{itemize}

\item For the dermatology ($T_1$) task, we defined the clinician applicable performance following a one vs. all approach following~\cite{liu2020deep}. This was done by computing the accuracy of the differential diagnosis of skin conditions provided by one US-board certified dermatologist against the aggregated reading from a panel of several US-board certified dermatologists as the ground truth.

\item Similarly, for chest X-ray ($T_3$) classification task, we followed the same strategy to compute the accuracy of one radiologist \vs the rest of the radiologists as suggested in~\cite{rajpurkar2018deep}. 

\item For pathology survival prediction ($T_5$) task, we followed the approach proposed in~\cite{wulczyn2021interpretable}. 

\item For the DME diagnosis ($T_2$) task, the ground truth was collected using the worldwide gold standard OCT machine, thus the clinician expert performance was not applicable. The clinician expert performance was also not available for the pathology metastases detection ($T_4$) out-of-distribution datasets, as well as for the mammography classification ($T_6$) task.

\end{itemize}

\begin{table}[t]
\centering
\caption{\small{\textbf{Clinically applicable performance.} Summary of clinically applicable performance range across the medical imaging tasks considered in this study in the out-of-distribution clinical setting wherever available.}}
\label{tab:clinical-equivalent-performance}
\renewcommand{\arraystretch}{1.4}
\vspace{6pt}
\footnotesize
\centering
\begin{tabular}{c@{\hspace{1.0em}}|c|c@{\hspace{1.0em}}} 
\toprule
\rowcolor{ourlightgray} 
\textbf{Tasks}  &\textbf{Task Name}                    &\textbf{Clinician Performance}            \\[1pt]\midrule
\textbf{$T_1$}  & Dermatology                          & ~~~0.650 (95\% CI 0.545–0.755)           \\[3pt]
\textbf{$T_3$}  & Chest X-ray Classification           & ~~~0.869 (95\% CI 0.843–0.894)           \\[3pt]
\textbf{$T_5$}  & ~~Pathology Survival Prediction~~    & ~~~0.684 (95\% CI 0.639-0.716)           \\[1pt]

\bottomrule
\end{tabular}
\end{table}

\subsection{Implementation Details.}
We build on top of the Tensorflow~\cite{tensorflow2015-whitepaper} implementation of SimCLR~\cite{simclr} as our model pretraining code base. We further integrated an extended data pipeline for each task to be able to read, preprocess and sample images based on patient meta-data for modalities such as pathology, mammography and Chest X-ray. We also implemented extra augmentation strategies relevant to medical imaging data including implementation of functions for elastic deformation, histogram equalization, and random rotation. We pretrained our models using 16 to 256 Google Cloud TPU cores depending on the chosen batch size, memory and the size of unlabeled dataset for each task. With the fixed input image size of 224$\times$224 and using 64 TPU cores, $\approx$24 hours was needed to pretrain a ResNet-50 (1$\!\times$) with batch size 1024 for 1000 epochs in a dataset with 200K examples. However, pretraining using large unlabeled datasets such as colored fundus photos (CFPs) and image patches extracted from digital pathology whole-slide images which includes millions of examples can take up to seven days ($\approx$150 hours) using 256 TPU cores. Once the models are pretrained, they are fine-tuned for the domain-specific downstream tasks. For example, we used the pretrained model obtained from the unlabeled pathology data for both pathology metastases detection and survival prediction tasks.


\section{Clinical Evaluation Settings}
In the following text, we provide details of our clinical evaluation setting including medical imaging tasks description, details of the datasets, description of distribution shifts, definition of clinically applicable performance, and details of the clinical impact analysis of using our proposed method for developing medical imaging AI.

\subsection{Tasks and Datasets}
We investigated 14 distinct datasets across different imaging modalities, dataset sizes, label spaces and class distributions to reflect the heterogeneity of medical imaging problems and evaluate various distribution shift scenarios. Specifically, we considered five popular modalities in medical imaging: dermatology, mammography, digital pathology, fundus imaging, and X-ray as listed in Table~\ref{tab:dataset-fingerprints} and 6 different tasks. These tasks are representative of many common characteristics and challenges of medical imaging (\eg~class label imbalance, variation of pathologies of interest from small local patches to more global patches, and image characteristic variations). Each specific modality and task includes an unlabeled pretraining dataset ($D_u$), an in-distribution dataset ($D_{in}$), and one or more out-of-domain datasets ($D_{out}$) which have been collected under clinical distribution shifts due to new data acquisition devices or different clinical demographics~\cite{finlayson2020clinician}. Table~\ref{tab:dataset-fingerprints} summarizes the task and datasets and Figure~\ref{fig:data-dist-shift} shows visual samples of each task.

\begin{table*}[t]
\centering
\caption{\textbf{Dataset fingerprints.} The above table illustrates the size and characteristics of the labeled, unlabeled and out-of-distribution dataset across the different medical imaging tasks we considered in this study.}
\vspace{+3pt}
\label{tab:dataset-fingerprints}
\footnotesize
\begin{tabular}{l|l|c|c|c|c|c|c|c}
\toprule

\rowcolor{ourlightgray} 
                 Tasks&          $D_u$&   \multicolumn{3}{c|}{$D_{in}$}&  \multicolumn{3}{c|}{$D_{out}$}&   Secondary $D_{out}$                               \\ \cline{4-9} 
\rowcolor{ourlightgray} 
                      &               &         Train&      Validation&         Test&        Train&     Validation&       Test&                     \\ \midrule
                 $T_1$&        207,032&        15,340&           1,190&        4,146&       17,322&          4,339&      6,639&               --    \\ [3pt]
                 $T_2$&      2,287,716&         3,874&             973&        1,192&        2,524&            643&        612&               323   \\ [3pt]
                 $T_3$&        215,695&       201,055&           9,027&       13,332&       27,978&         17,723&      1,998&               --    \\ [3pt]
                 $T_4$&         10,705&           216&              54&          129&        2,577&          1,295&      1,289&               273   \\ [3pt]
                 $T_5$&         10,705&         2,236&           1,128&        1,132&          402&            101&        168&               --    \\ [3pt]
                 $T_6$&         77,340&        26,739&          49,831&       12,448&       17,178&          11,551&    12,314&               --    \\ [3pt]
\bottomrule
\end{tabular}

\end{table*}

\paragraph{Task 1: Dermatology condition classification.} The dermatology condition classification task ($T_1$) targets identification of various types of skin conditions from digital camera images. For this task, the experiment setup and dataset of~\cite{azizi2021big,liu2020deep} have been followed. Unlabeled pretraining dataset and in-distribution training dataset were collected and de-identified by a US-based tele-dermatology service with images of skin conditions taken using consumer-grade digital cameras. Intrinsically, these images exhibit variations in pose, lighting and camera focus. Additionally, the target body part and also their backgrounds embody noise artifacts such as variations in clothing. Each case includes between one to six images and during the data preparation step, cases with the occurrence of multiple skin conditions or ungradable images were filtered out (for details, see~\cite{liu2020deep}).  This data was collected in the US, and the ground truth labels were aggregated from a panel of several US-board certified dermatologists who provided a differential diagnosis of skin conditions in each case. As in actual clinical settings, the distribution of different skin conditions is heavily skewed in this dataset, ranging from some skin conditions making up more than 10\% of the training data like acne, eczema, and psoriasis, to those making up less than 1\% like lentigo, melanoma, and stasis dermatitis~\cite{liu2020deep}. To ensure the existence of sufficient data in each category for model development, the most common 26 skin conditions out of 419 unique conditions were identified and the rest were grouped into an additional `Other' class leading to a final label space of 27 classes for this task. The 26 target skin conditions include: Acne, Actinic keratosis, Allergic contact dermatitis, Alopecia areata, Androgenetic alopecia, Basal cell carcinoma, Cyst, Eczema, Folliculitis, Hidradenitis, Lentigo, Melanocytic nevus, Melanoma, Post-inflammatory hyperpigmentation, Psoriasis, Squamous cell carcinoma/squamous cell carcinoma in-situ (SCC$/$SCCIS), Seborrheic keratosis, Scar condition, Seborrheic dermatitis, Skin tag, Stasis dermatitis, Tinea, Tinea versicolor, Urticaria, Verruca Vulgaris, and Vitiligo. 

In total, the in-distribution dataset, $D_{in}^{T_1}$ includes a total of 20,676 unique cases. The final train, validation, and test sets include a total of 15,340 cases, 1,190 cases, and 4,146 cases, respectively.  We also use an additional de-identified out-of-distribution dataset, $D_{out}^{T_1}$, to investigate the generalization performance of our proposed method under distribution shift. Unlike $D_{in}^{T_1}$ this dataset is primarily focused on skin cancers and the ground truth labels are obtained from biopsies~\cite{azizi2021big}. $D_{out}^{T_1}$ was further split to train, validation, and test sets consisting of 17,322 cases, 4,339 cases, and 6,639 cases, respectively. The out-of-distribution dataset, $D_{out}^{T_1}$, was collected by a chain of skin cancer clinics in Australia and New Zealand. This dataset has a much higher prevalence of skin cancers such as Melanoma, Basal Cell Carcinoma, and Actinic Keratosis. For self-supervised pretraining, $D_{U}^{T_1}$, was formed by using a total of 207,032 unlabeled images from $D_{in}^{T_1}$ where we removed the annotations. 




\paragraph{Task 2: Diabetic Macular Edema classification.} Diabetic Macular Edema (DME) is distinguished by thickness of the central area of the retina due accumulation of intraretinal fluid. While it is possible to screen for DME using color fundus photographs (CFP) by detecting hard exudates near the fovea as a surrogate for the presence of fluid, extracting the thickness directly from a three-dimensional optical coherence tomography (OCT) volume has become the gold standard for making a diagnosis~\cite{virgili2015optical}. Nevertheless, the use of OCT machines for DME diagnosis word-wide is limited due to high cost~\cite{word2019vision}. In this task we followed the approach of~\cite{liu2022deep,de2018clinically,varadarajan2020predicting} to leverage a dataset of paired CFP and OCT data, and trained a model that takes a CFP as input, and predicts central retinal thickness (CRT) measured from the corresponding OCT. Specifically, CRT was defined as ETDRS zone 1/central sub-field thickness (CST)~$\geq$~300{\textmu}m~\cite{liu2022deep,brown2004detection,sadda2006automated}.

\textcolor{black}{For the pretraining purposes, we used the unlabeled dataset from EyePACS Inc., $D_{u}^{T_2}$ which includes 2,287,716 fundus images from 308,507 patients where Hispanic is the most prevalent race/ethnicity in this cohort.} The in-distribution  dataset (collected in Thailand) $D_{in}^{T_2}$ includes 6,039 fundus images from 4,035 patients. We split this dataset to train, validation, and test sets including a total of 3,874 images, 973 images, and 1,192 images, respectively.  We also used a primary de-identified out-of-distribution dataset, $D_{out}^{T_2}$, to investigate the generalization performance of our proposed method under distribution shift. Unlike $D_{in}^{T_1}$ this dataset was collected in Australia and includes the total number of 3,779 fundus images from 879 patients. $D_{out}^{T_2}$ was also further divided into the train, validation, and test sets including 2,524 images, 643 images, and 612 images, respectively. Additionally, we used a secondary de-identified out-of-distribution dataset consisting of 909 fundus images from 323 patients collected in India for zero-shot out-of-distribution performance evaluation.


\paragraph{Task 3: Chest X-ray  condition classification.} The Chest X-ray Condition classification task ($T_3$) involves multi-label classification of chest X-ray images among five common findings: atelectasis, consolidation, pulmonary edema, effusion, and cardiomegaly. These were chosen due to their prevalence, in particular in $D_{in}^{T_3}$. Each finding was modeled as an independent binary prediction. Three publicly available datasets were used for training and evaluation purposes: CheXpert~\cite{irvin2019chexpert}, MIMIC-CXR~\cite{johnson2019mimic}, and ChestX-ray14~\cite{wang2017chestx}.

We used the training split of MIMIC-CXR~\cite{johnson2019mimic} as $D_{u}^{T_3}$, a dataset consisting of 215,695 radiographic studies collected at Beth Israel Deaconess Medical Center in Boston, MA, for pretraining. Each study contains potentially multiple views, so we sampled from these images during pretraining (preferentially sampling PA or AP if available). This dataset was combined with the in-distribution dataset during pretraining.

CheXpert~\cite{irvin2019chexpert} is a large open-source dataset of de-identified chest radiograph (X-ray) images (224,316 chest radiographs coming from 65,240 unique patients). The ground truth labels for training data were automatically extracted from radiology reports. The radiologist report was then mapped to a label space of 14 radiological observations. We predicted the five most prevalent pathologies used by Irvin and Rajpurkar \textit{et al.}~\cite{irvin2019chexpert}. Following previous work~\cite{neyshabur2020being,raghu2019transfusion,mustafa2021supervised,azizi2021big}, to facilitate a robust comparison of our method to standard approaches, we defined a custom subset of the CheXpert dataset. For this purpose, the full training set was randomly re-split into training, validation, and test images (see Table~\ref{tab:dataset-fingerprints}). This means the performances of our models are not directly comparable to those reported in~\cite{irvin2019chexpert}. Nonetheless, we believe the relative performance of models is representative, informative, and comparable to~\cite{neyshabur2020being,raghu2019transfusion,mustafa2021supervised,azizi2021big}. ChestX-ray14 is an open-source dataset collected at the National Institutes of Health Clinical Center, MD, USA. The data is labeled in a manner similar to CheXpert by extracting common findings from radiologist reports.

\paragraph{Task 4: Pathology lymph node metastases detection.} In the lymph node metastases detection task ($T_4$), the goal is to detect cancer metastases in digital whole-slide images of lymph node histology slides. The models were trained in a weakly-supervised manner, using only case-level labels and without any local annotations. In order to make case-level predictions, embeddings from $2^{14}$ = 16,384 patches per case were combined via an attention layer~\cite{ilse2018attention}. A random sample of 50M patches from 10,705 cases (29,018 slides) spanning 32 studies from The Cancer Genome Project (TCGA) was used for self-supervised pretraining ($D_{u}^{T_4}$). Breast lymph node slides from the CAMELYON-16 challenge~\cite{bejnordi2017diagnostic} were used for model development and in-distribution evaluation ($D_{in}^{T_4}$). Lymph node slides from 5,161 stage II and III colorectal cancer cases (36,520 slides) collected between 1984-2007 from the Institute of Pathology and the BioBank at the Medical University of Graz were used for out-of-distribution evaluation ($D_{out}^{T_4}$). $D_{out}^{T_4}$ was divided into train, validation, and test sets including 2,577 cases (17,904 slides), 1,295 cases (9,313 slides), and 1,289 cases (9,303 slides) respectively. This dataset is further described in Wulczyn \etal~\cite{wulczyn2021interpretable}, however here cases were not excluded based on having insufficient tumor content in the primary tissue slides. Additionally, we used a fraction of the CAMELYON-17 dataset as an additional out-of-distribution dataset for pathology metastases detection which includes 273 pathology slides. We particularly removed overlapping CAMELYON-16 pathology slides and cases from the original CAMELYON-17 dataset.  

\paragraph{Task 5: Pathology colorectal cancer survival prediction.} The objective of the colorectal cancer survival prediction task ($T_5$) is to predict 5-year disease-specific survival (DSS) using digitized whole-slide images of primary colorectal tissue histology slides. Models were trained in a weakly-supervised manner and used the same self-supervised pretraining dataset ($D_{u}^{T_4}$) as the lymph node metastases detection task ($T_4$). Colorectal tissue slides from 4,496 stage II and III colorectal cancer cases (36,841 slides) collected between 1984-2007 from the Institute of Pathology and the BioBank at the Medical University of Graz were used for model development and in-distribution validation ($D_{in}^{T_5}$). A temporal split of 671 cases (6,419 slides) collected between 2008-2013 from the same institution were used for out-of-distribution evaluation ($D_{out}^{T_5}$). $D_{out}^{T_4}$ was divided into train, validation, and test sets including 402 cases (3,873 slides), 101 cases (913 slides), and 168 cases (1,633 slides), respectively. This dataset is further described in Wulczyn \etal~\cite{wulczyn2021interpretable}, however, only cases not lost to follow-up for DSS within 5-years were included here.

\paragraph{Task 6: Mammography classification.} In the mammography cancer classification task ($T_6$), the goal is to predict whether there will be biopsy-confirmed cancer occurring in the 39-months following the screening episode, as described in~\cite{mckinney2020international}.  We utilized multiple different datasets collected in various geographic locations in this task. This included a labeled dataset collected in the UK, a labeled dataset from the USA (Northwestern Memorial Hospital), an unlabeled set of images from five clusters of hospitals across five different cities in India (Bangalore, Bhubaneswar, Chennai, Hyderabad, and New Delhi), and another unlabeled set of images collected from Northwestern Memorial Hospital (Chicago, USA). Each of these datasets contains four different images per patient and for each breast (left and right) there is a medio lateral oblique (MLO) and craniocaudal (CC) view. The UK and USA datasets are described in more detail in~\cite{mckinney2020international}.

The UK dataset was used as the labeled in-distribution data, ($D_{in}^{T_6}$) which includes the total number of 89,018 cases. The training set of this dataset contains 26,739 cases which are unevenly divided between 4,751 positive cases and 21,988 negatives. We up-sample positive cases with the factor of 10 to balance the distribution of negative and positive cases and further improve the performance. The original validation set includes 49,831 cases with a total of only 992 positive cases.  Also, the test data contains 12,448 cases with 249 positives and 12,199 negatives. The labeled dataset from the US was used as out-of-domain data, $D_{out}^{T_6}$. This dataset which includes the total number of 41,043 cases was split to train, validation, and test set with 27,083 cases, 6,901 cases, and 7,059 cases, respectively. In this dataset positive cases are only 9\% of all of the cases.  For pretraining, the unlabeled dataset ($D_{u}^{T_6}$) was formed by removing labels from the labeled data from the UK dataset and combining it with the unlabeled data from the India. During pretraining, as suggested by Azizi~\etal~\cite{azizi2021big,vu2021medaug} to improve the positive pair mining procedure a single image was randomly selected from the four possible views that were further used to generate a pair.

\subsection{Qualitative Analysis of Distribution Shifts} 

\begin{table*}[t]
\centering
\caption{\small \textbf{Analysis of distribution shifts.} The three most frequent sources of data distribution shifts in medical AI datasets are population shifts, technology shifts, and behavioral shifts, each arising from various influencing factors such as changes in acquisition devices,  disease prevalence, etc~\cite{finlayson2020clinician}. The check-mark (\checkmark) and dash sign ($-$) indicate the existence or absence of the shift factor, and $U$ indicates whether the evidence showing the factor is unknown or undefined in the metadata information associated with the datasets considered for the given medical imaging task. }
\vspace{3pt}
\label{tab:my-table}
\footnotesize
\begin{tabular}{l|l|cccccc}
\toprule
\rowcolor{ourlightgray}
\multicolumn{2}{l|}{Tasks}& $T_1$      & $T_2$      & $T_3$      & $T_4$      & $T_5$      & $T_6$      \\ [2pt]\midrule
Technology Shift & Acquisition device shift                 & \checkmark & \checkmark & U          & \checkmark & \checkmark & \checkmark \\ [4pt]
                 & IT practice, software, terminology shift & \checkmark & \checkmark & \checkmark & \checkmark & --         & \checkmark \\ [4pt]\midrule
Population Shift & Demographic shift                        & \checkmark & \checkmark & \checkmark & --         & --         & \checkmark \\ [4pt]
                 & Clinical setting shift                   & \checkmark & --         & --         & --         & --         & \checkmark \\ [4pt]
                 & Disease prevalence shift                 & \checkmark & \checkmark & --         & \checkmark & U          & \checkmark \\ [4pt]
                 & Seasonal shift                           & \checkmark & U          & U          & --         & --         & --         \\ [4pt]\midrule
Behavior Shift   & Clinical behavior and incentives shift   & U          & \checkmark & --         & --         & \checkmark & \checkmark \\ [4pt]
                 & Patient behavior change                  & U          & U          & --         & U          & \checkmark & U          \\ [4pt]
                 & Clinical practice change                 & U          & U          & --         & U          & \checkmark & U          \\ [4pt]
                 & Clinical nomenclature shift              & U          & U          & --         & U          & \checkmark & U          \\ [4pt]
\bottomrule
\end{tabular}
\end{table*}

In the following section, we provide details of the distribution shifts, including the definition, visual effect of distribution shifts, and distribution shift analysis across the different medical imaging tasks in this study.   

\paragraph{Type of distribution shifts} Distribution shifts, where the training distribution differs from the target test distribution, can substantially degrade the diagnostic performance and model calibration of AI systems deployed in new clinical setups and further negatively affect existing health disparities~\cite{finlayson2020clinician}.  To qualitatively benchmark the extent of the distribution shifts, in each task and dataset, we view the overall data distribution shifts as a mixture of multiple possible changes represented by $S_i$. Distribution shift from a given in-distribution set ($D_{in}$) to a given out-of-distribution set ($D_{out}$) can be represented by $\bigcup_{i\in I}^{} S_i$, where $S=\{ S_1, S_2, ..., S_n \}$ represents the set of possible distribution shift sources, and $I$ is the index-set. Here, we limit $n$ to the three most frequent sources of data distribution shifts - population shifts ($S_1$), technology shifts ($S_2$), and behavioral shifts ($S_3$) as suggested in~\cite{finlayson2020clinician}. A large $\left | I \right |$ represents a severe degree of distribution shift between two datasets. 

Changes in technology, one of the most common causes of data distribution shifts, can consist of new types or different models or brands of data-acquisition devices, changes in the IT practices and infrastructure for the upstream task. Changes in demographics are identified by changes in characteristics of the population in which the model was developed \vs the target test population and includes but is not limited to shifts in age, sex, race, and ethnicity. Behavioral changes arise from changes in patient behavior, clinician behavior, clinical practice, clinical nomenclature, clinical incentives, and finally AI system-induced behavioral changes~\cite{finlayson2020clinician}. For example, adding surgical skin marking can affect the accuracy of the dermatology classifier~\cite{winkler2019association}. 

We focused on these distribution shifts because they collectively capture the structure of most of the shifts in medical applications. However, it is worth noting that identifying and analyzing all of the causal factors in different clinical settings is challenging, and sometimes impossible.  

\emph{Dermatology condition classification:} In the dermatology classification task, we observed both technology shift and population shift between in-distribution and out-of-distribution datasets. Different acquisition devices were used to collect the in-distribution (consumer-grade digital cameras) and out-distribution (tele-dermatology service) images. In addition, IT practice, software, and technology also shifted due to changes in clinical location and setting. Demographic shift, disease prevalence shift, and seasonal shift also occurred since the in-distribution dataset was collected in multiple US clinics whereas the out-of-distribution dataset was collected in Australia and New Zealand. 

\emph{Diabetic Macular Edema classification:} The DME in-distribution and out-of-distribution data were collected using different scanners and in different hospitals and countries, with the former collected in Thailand and the latter in Australia. Due to this demographic shift, the rate of diabetes and the prevalence of diseases differs between the two datasets. Also, clinical incentives to detect DME between these two countries varies which introduces behavioral shifts. So, we observe technology shift, demographic change, and behavioral shift between in-distribution and out-of-distribution datasets in the DME task. These shifts also hold between DME in-distribution and the additional out-of-distribution (collected in India) datasets as detailed in the Supplementary Materials section. 

\emph{Chest X-ray classification:} The Chest X-ray in-distribution (CheXpert, Stanford) and out-of-distribution (ChestX-ray14, Virginia) were collected in different demographic and hospitals and undergo technology shift, and population shifts. However, there is no evidence to show behavior changes between these datasets. 

\emph{Pathology tasks:} In the pathology metastases task, the in-distribution data targets the breast lymph node slides while the out-of-distribution data consists of lymph node slides of colorectal cancer. Also, the data has been collected on different clinical sites and scanned with different devices. Therefore, we observe technology shifts and population shifts between in-distribution and out-of-distribution datasets in the pathology lymph node metastases detection task. The data used for pathology survival prediction in-distribution (collected between 1984-2007) and out-of-distribution (collected between 2008-2013) were also collected in different clinical setups and different times.  

\emph{Mammography classification:} The mammography in-distribution (UK) and out-of-distribution (US) data were collected in different countries and with various scanners. However, factors affecting behavior shift between these datasets are unknown. Therefore, we consider technology shifts and population shifts between in-distribution and out-of-distribution datasets in this task.

\begin{figure*}[!tbh]
    \centering
    \includegraphics[width=0.8\textwidth]{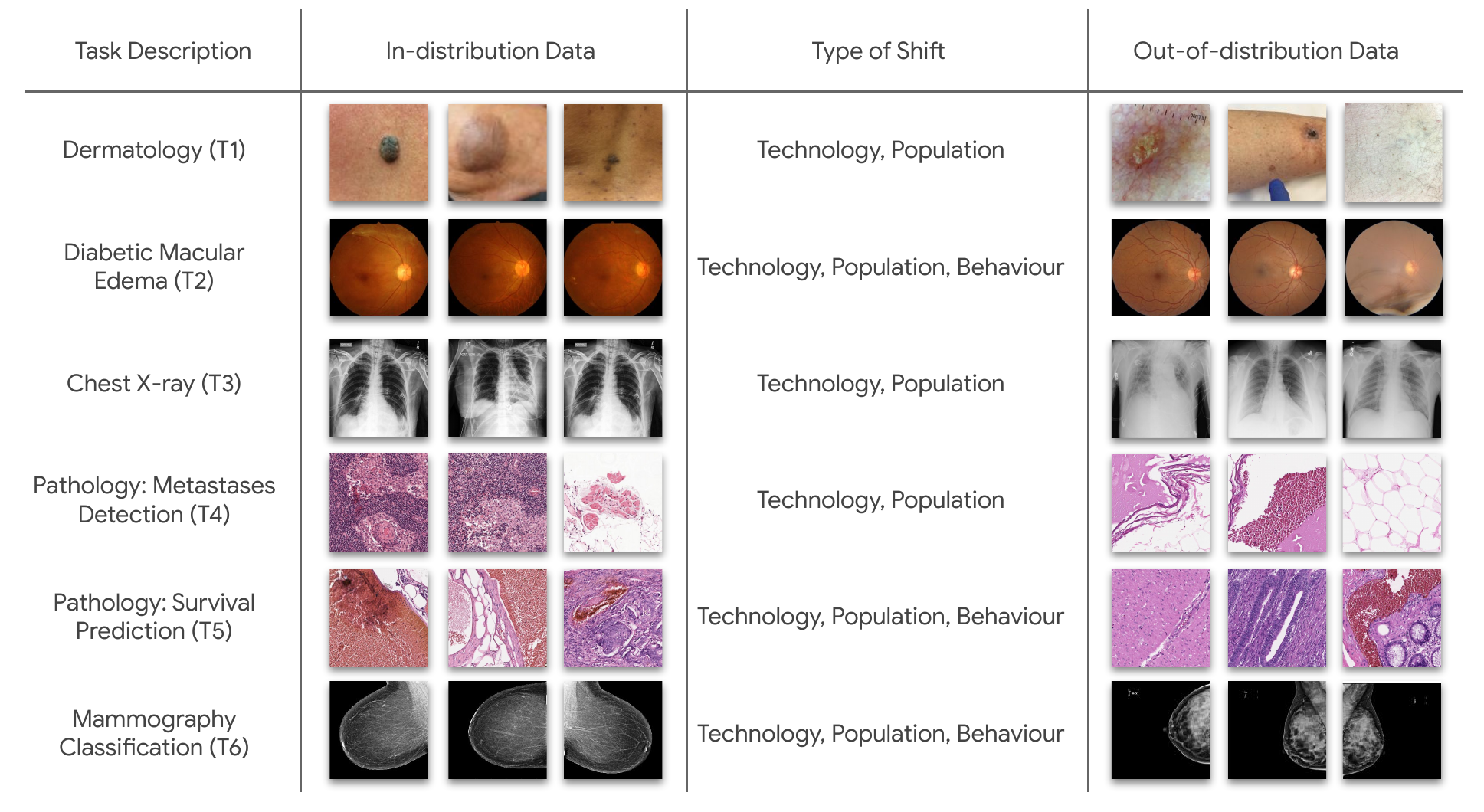}
    \caption{\small \textbf{Visual samples of distribution shifts across the medical imaging tasks considered in this study.} Variation between in-distribution and out-of-distribution data can be visually subtle or pronounced. This variation includes but is not limited to changes in contrast, sharpness or tint, differences in non-linear effects of X-ray sensor construction or in zoom levels. The underlying cause of the distribution shift can be associated with technology shift, demographic shift, or behavioural shift~\cite{finlayson2020clinician}.}
    \label{fig:method-data-shift-samples}
    \vspace{-0cm}
\end{figure*}

\paragraph{Visual effects of distribution shifts} As discussed above, models used in medical AI applications are often trained on data from a small number of hospitals (in-distribution data), but the goal is usually to deploy these models more generally across other hospitals or other varied clinical settings (out-of-distribution). Variations in data collection and processing can degrade model accuracy on data from new clinical deployment settings. This variation can be subtle or visually pronounced, ranging from changes in contrast, sharpness, tint, to non-linear effects of X-ray sensor construction, differences in zoom levels, etc. For example, in histopathology this variation can arise from sources such as slide staining, and image acquisition differences. Figure~\ref{fig:method-data-shift-samples} shows visual samples of in-distribution and out-of-distribution images used in this study. As there is no canonical way to quantify these distribution shifts, we assessed the quality of the dataset shift according to Finlayson \etal~\cite{finlayson2020clinician}. 

\subsection{Clinical Impact Analysis}
Medical data acquisition and annotation is often extremely expensive and time-consuming~\cite{willemink2020preparing}. The cost associated with medical data consists of multiple components such as acquisition cost, handling cost (\eg~image de-identification), curation cost (\eg~ data authentication, archiving, management), and annotation cost. To gain a better understanding of the benefits of our data-efficient generalization strategy, we attempt to collect an estimate of the cost and clinician hours associated with data annotation for each of our target modalities and specifically for the out-of-distribution data. Since the exact cost of data annotation is not available for many datasets considered in this study, we estimated them using public data for US-based clinicians from \href{salary.com}{salary.com website}. Estimates of clinician hours needed for data annotation was also collected from sources relevant to each task such as~\cite{seah2021effect,haygood2009timed,jain2021development}. However, these are approximations, and the exact labeling cost likely includes additional overhead and would largely depend on the experience level of the clinicians providing the labels as well as the choices made by the medical AI developers (\eg~the number of clinicians reviewing each case to generate the gold standard ground truth). In certain cases, like mammography, the ground-truth is derived from real world data like biopsies or outcomes and the clinician annotation is primarily used for comparison. However, such data while being more reliable is once again extremely time-consuming to obtain often spanning many years.

Table~\ref{tab:method-clinical-cost} summarizes clinical costs associated with the acquisition, curation and annotation of each out-of-distribution dataset and in all cases, we focus on train splits. The annotation cost and hours savings are calculated based on the percentage of the data that REMEDIS needs to match the performance of supervised baseline as depicted in Figure~\ref{fig:main-results}. The overall annotation cost of a dataset is equal to (average annotation cost per image)~$\times$~(number of training images) and the overall annotation clinician hours for each dataset is equal to (average annotation time per image)~$\times$~(number of training images). We also report the approximate duration needed to collect each dataset starting from the first patient recruitment.  A similar calculation for acquisition time savings could be done. Since with REMEDIS, a smaller out-of-distribution dataset is sufficient for reaching baseline or clinically-applicable performance, the total data acquisition duration can be significantly reduced. However, this is not shown because the calculation implies multiple assumptions, including a uniform acquisition of different classes or pathologies over time, which may not always be valid.

\emph{Dermatology condition classification:} For dermatology, we used a median diagnosis time of 60 seconds for this particular dataset~\cite{jain2021development}. The average hourly wage of a dermatologist is~\href{https://www.salary.com/research/salary/alternate/dermatologist-hourly-wages}{\$172}, resulting in an average cost per case of \$2.86. The cost of annotation of dermatology out-of-distribution train dataset which consists of 17,322 dermatology cases is estimated at \$49,540 and the total clinician hours associated with this dataset is 289 hours.  In this task, REMEDIS can reach the highest supervised baseline performance using only 33\% of the total training data, which means REMEDIS potentially offers 193 clinician hours and \$33K saving. In addition, $D^{T1}_{out}$ was collected over 9 years between 2007 and 2016 (very roughly estimating the effect of dataset acquisition duration when using REMEDIS, with only 33\% of the data needed for reaching baseline performance, the data collected after the first 3 years may have been sufficient for REMEDIS).

\emph{Diabetic Macular Edema (DME) classification:} For Diabetic Macular Edema (DME) while we have access to expert reader numbers, OCT generated labels were used for the AI model development. This means the cost associated with this task is not comparable to clinician annotator cost in other tasks. Nevertheless, based on ~\href{https://www.webmd.com/eye-health/what-is-retinal-imaging}{WebMD} and~\cite{pugh1993screening} the amount of time that takes to perform an OCT test is 5 minutes and 6.5 minutes, respectively. We estimate an average of 5.75 minutes (345 seconds) per case to obtain an OCT-generated label. The average hourly wage of an ophthalmologist is~\href{https://www.salary.com/research/salary/alternate/ophthalmologist-hourly-wages}{\$147}, resulting in an average cost per case of~\$14. Considering the 2,524 cases in the $D^{T2}_{out}$ train split, the cost of annotation of this dataset is estimated at \$35,336 and with an annotation duration of 242 hours using OCT machines.  In this task, REMEDIS can reach the highest supervised baseline performance using only 7\% of total training data, which means REMEDIS possibly offers 224 clinician hours and \$33K saving. In addition, $D^{T2}_{out}$ was collected over 7 years between 2003 and 2020~\cite{liu2022deep}.

\emph{Chest X-ray condition classification:} For chest X-rays, we use a reported interpretation time of 122 seconds per image~\cite{seah2021effect}, and an average hourly wage of~\href{https://www.salary.com/research/salary/alternate/radiologist-hourly-wages}{\$205} of radiologists, resulting in a cost of \$6.98 per image. The train partition of our out-of-distribution dataset in this task includes 27,978 cases which translate to \$194,369 in cost of collection and 948 clinician hours. REMEDIS can reach the highest supervised baseline performance without any out-of-distribution training data, which means REMEDIS offers 789 clinician hours and \$162K saving. The $D^{T3}_{out}$ was collected over 23 years between 1992 and 2015.

\emph{Pathology tasks:} For the pathology task data, the annotation cost of a single pathology slide is estimated at around~\$23 where the time spent on a single slide is estimated at 10 minutes, and the average salary of a pathologist is estimated at~\href{https://www.salary.com/research/salary/alternate/pathologist-hourly-wages}{\$138} per hour. Based on this, annotation of pathology metastases detection out-of-distribution, $D^{T4}_{out}$, train split which includes 2,577 cases (17,904 pathology slides) is costs around~\$411,792 and requires 2,984 clinician hours. Our method can attain the baseline performance using only 6\% of the data, suggesting that it can contribute to over~\$385K savings in the cost of the annotations. The $D^{T4}_{out}$ which includes lymph node slides from stage II and III colorectal cancer cases was collected over the periods 23 years between 1984 and 2007. In pathology survival prediction, $T_5$, each slide was annotated for the possible clinical outcome and the out-of-distribution dataset which includes colorectal tissue slides from stage II and III colorectal cancer cases was collected over five years between 2008 and 2013. The $D^{T5}_{out}$ train split consists of 402 cases (3,873 pathology slides) and the cost of annotation is around \$89,079 and needs 645 clinician hours for the annotation. REMEDIS can cut this cost to \$13K and offer \$76K saving by using 14\% of the whole data.     

\emph{Mammography classification:}  For the mammography data, the annotation task includes localization and grading of the images. The cost of annotation per hour is estimated based on the average salary of a radiologist, reported as~\href{https://www.salary.com/research/salary/alternate/pathologist-hourly-wages}{\$205} per hour. Moreover, it takes 5-6 minutes to annotate a single mammography image and the annotation cost for a single mammography image is estimated at \$20.5. For the mammography classification out-of-distribution dataset training split, this translates to a total cost of \$352,149 for 17,178 images,  requiring 1,718 clinician hours for annotation. This dataset was collected over 17 years between 2001 and 2018. In this task, REMEDIS can attain the highest supervised baseline performance without using any out-of-distribution example, which means REMEDIS offers 1,569 clinician hours and \$322K saving. 
  
It is worth mentioning that the saved annotation overhead can potentially significantly impact the feasibility of site-specific adaptation/retraining of a model. In addition to the visible cost and clinician hours that hinder medical image data collection and medical AI model adaptation, there exist several significant concerns such as the speed of acquiring local labels which may take years to manifest if based on outcomes. For example in the survival prediction task, collection of 645 examples takes over 5 years, or low incident of positive cases in tasks such as mammography screening significantly slows down the process of data collection to an extent where a proper dataset is collected over 20 years.  Therefore, the saved clinician hours can be translated to saving multiple years that it takes to properly recruit patients, collect and curate a medical dataset.

\begin{table*}[]
\centering
\caption{\small \textcolor{black}{\textbf{Clinical cost analysis of data acquisition and annotation.}} The table below provides a summary of clinical costs associated with the collection of the out-of-distribution dataset for all the medical imaging tasks considered in this study. In all cases, we focus on the train splits of the dataset. The acquisition time for each task approximates the time that it took to collect each dataset starting from the first patient recruitment. The cost and hour savings are calculated based on the percentage of the data that REMEDIS requires to match the performance of \textcolor{black}{the strong supervised baseline pretrained on JFT-300M dataset as depicted in Figure~\ref{fig:main-results}}. The overall annotation cost of a dataset is equal to (average annotation cost per image)~$\times$~(number of training images) and the overall clinician hours for each dataset is equal to (average annotation time per image)~$\times$~(number of training images).}
\vspace{6pt}
\label{tab:method-clinical-cost}
\footnotesize
\renewcommand{\arraystretch}{1.25}
\begin{tabular}{@{\hspace{.5em}}l@{\hspace{.5em}}|@{\hspace{.5em}}cccccc}
\toprule
\rowcolor{ourlightgray}
Tasks                                         & $T_1$            & $T_2$           & $T_3$          & $T_4$            & $T_5$             & $T_6$            \\ [3pt] \midrule
Number of training images                     & 17,322           & 2,524           & 27,978         & 17,904           & 3,873             & 17,178           \\ [3pt] \midrule
Average annotation time per image (second)    & 60               & 345             & 122            & 600              & 600               & 360              \\ [3pt]
Average hourly wage of clinician (\$)         & \$172            & \$147           & \$205          & \$138            & \$138             & \$205            \\ [3pt]
Average cost of annotation per image (\$)     & \$2.86           & \$14            & \$6.95         & \$23             & \$23              & \$20.5           \\ [3pt] \midrule 
Clinical hours cost of dataset (hour)         & 289              & 242             & 948            & 2,984            & 645               & 1,718            \\ [3pt]
Annotation  cost of dataset  (\$)             & \$49K            & \$35K           & \$194K         & \$411K           & \$89K             & \$352K           \\ [3pt]
Acquisition Time (Years)                      & 9                & 7               & 23             & 23               & 5                 & 17               \\ [3pt]
Acquisition Period                            & 2007-2016        & 2003-2020       & 1992-2015      & 1984-2007        & 2008-2013         & 2001-2018        \\ [1pt] \midrule
Percentage of data saved using REMEDIS        & 67\%             & 93\%            & 83\%           & 94\%             & 86\%              & 91\%             \\ [3pt]
Amount of data saved using REMEDIS            & 11,578           & 2,342           & 23,278         & 16,872           & 3,325             & 15,689           \\ [3pt]
Clinical annotation hours saved (hour)        & 193              & 224             & 789            & 2,812            & 554               & 1,569            \\ [3pt]
Annotation cost saved (\$)                    & \$33K            & \$33K           & \$162K         & \$385K           & \$76K             & \$322K           \\ [3pt]
\bottomrule
\end{tabular}
\end{table*}

\section{Additional Related Work}
\label{method:related-work}
Improving the robustness and generalization ability of AI models under distribution shifts is a longstanding challenge that becomes more critical given the progressive adoption of these models in safety-critical settings like healthcare and self-driving cars in recent years. Unlike previous studies, here we specifically consider ``data-efficient generalization'' which targets a more practical (following real-world deployment scenarios) but also challenging setup that aims to generalize to a new distribution by using none or a very few data examples from the target deployment data distribution.
Our proposed method is closely related to transfer learning, self-supervised, and contrastive learning as well as to literature on robustness, domain adaptation and generalization. Here we review prior works that are key for understanding the context of our study.

\paragraph{Robustness and out-of-distribution (OOD) generalization. } 
To tackle the challenges from distribution shift, numerous efforts have been made over the years, resulting in a rich literature. The proposed techniques vary greatly ranging from causality~\cite{scholkopf2021toward} to representation learning~\cite{bengio2013representation} and model-based~\cite{liu2021heterogeneous,robey2021model} techniques. Based on the model development strategy, one can address OOD generalization by improving the underlying representation learning method, introducing a new mapping function between distributions or by formulation as a new optimization problem~\cite{shen2021towards,wang2021generalizing,zhou2021domain}. Nevertheless, a recent paper~\cite{gulrajani2020search} suggested that under careful evaluation settings, models developed using standard Empirical Risk Minimization (ERM)~\cite{vapnik1998statistical,zhang2017mixup} remain a strong baseline for the generalization problem.

Representation learning techniques aim to learn distinct and informative representations~\cite{bengio2013representation,locatello2019challenging} that can improve transfer learning and OOD generalization~\cite{geirhos2018imagenet,geirhos2018generalisation}. For this purpose both conventional disentangled representation techniques based on Variational Autoencoders (VAE)~\cite{kim2018disentangling}, casual representation learning~\cite{scholkopf2021toward} and CausalVAE~\cite{yang2021causalvae} has been studied. However, it remains unclear whether disentangled representation benefits OOD generalization. Some findings suggest that the learned disentangled representation fails to extrapolate to unseen data~\cite{leeb2020structure,geirhos2018imagenet,geirhos2018generalisation}, while multiple studies verify the ability of those representations to generalize under OOD circumstances~\cite{trauble2021disentangled,dittadi2020transfer,andreassen2021evolution}. Nevertheless, the adaptation of these methods in medical image analysis has been limited and yet to be explored.  

There is also rapid development around large-scale pretraining models such as BiT~\cite{kolesnikov2019big} and CLIP~\cite{radford2021learning} to improve the learned representation and consequently generalization.~\cite{andreassen2021evolution} demonstrated that such pretrained models in the middle of fine-tuning as well as zero-shot pretrained models represent an entire class of techniques that exhibit a high amount of Effective Robustness (ER)~\cite{taori2019robustness}. Our method is closely related to this group, however, here we consider learning representation using a combination of large-scale pretraining and self-supervision. Unlike these prior works which use standard, synthetic visual corruption datasets and benchmarks such as ImageNet-C~\cite{hendrycks2019benchmarking} we consider a realistic, real world setup by using datasets that undergo distribution shift in the clinical setting. 

End-to-end learning mechanisms for OOD generalization design various models and learning strategies to address the generalization problem which also includes the category of domain adaptation methods. Representation learning is still a principal component in domain adaptation~\cite{albuquerque2020adversarial,li2018deep,ganin2015unsupervised,ganin2016domain}.  A popular baseline for domain adaptation is domain-adversarial neural network (DANN)~\cite{ganin2015unsupervised,ganin2016domain}. DANN learns representations that are discriminative and invariant to domains by jointly optimizing the underlying features, a label predictor that predicts class labels and is used both in the training and inference phase, and a domain classifier that discriminates between the source and the target domains during training. The representations are trained to confuse the domain classifier using gradient reversal so that domain invariant features are learned. There are multiple variants of DANNs~\cite{albuquerque2020adversarial,li2018deep,shao2019multi} including conditional invariant adversarial network (CIAN)~\cite{li2018deep} which learns class-specific adversarial networks via a conditional invariant adversarial network. Most domain adaptation methods belong to domain alignment where the central idea is to minimize the difference between source and target domains data by learning domain-invariant representations~\cite{motiian2017unified,muandet2013domain,shao2019multi,wang2021generalizing,zhou2021domain}. Some other prior works also study training strategies for domain generalization which includes meta-learning, ensemble learning, unsupervised and semi-supervised domain adaptations~\cite{zhou2021domain}. However, there is limited works investigating domain adaptation and generalization for medical imaging despite it being a critical unmet need and practically the most common solution to this is problem is site-specific data collection and model retraining~\cite{finlayson2020clinician} which as we show can be prohibitively expensive and time consuming for large scale, real world medical imaging AI deployment.\textcolor{black}{Recently, the closely related topic of domain generalization in the medical setting has been studied in [120] where Zhang~\etal introduced a framework to induce synthetic but realistic domain shifts. They benchmark the performance of eight domain generalization methods on multi-site clinical time series and medical imaging datasets. This setup is different from our proposed setup, where we focused on retraining performance in retrospective data mirroring real world clinical settings.} 

\paragraph{Transfer learning.}  Despite the differences in image statistics, scale, and task-relevant features, transfer learning from natural images is commonly used in medical imaging~\cite{esteva2017dermatologist,liu2020deep,mckinney2020international,menegola2017knowledge,wu2019deep,xie2019dual}. Multiple empirical studies suggest that such transfer learning strategy improves performance~\cite{alzubaidi2020towards,graziani2019visualizing,heker2020joint} for medical imaging tasks. However, detailed investigations of this strategy in~\cite{raghu2019transfusion} demonstrate that this does not always improve performance in medical imaging contexts but transfer learning from ImageNet can speed up convergence, and is particularly helpful when the medical image training data is limited. Importantly, the study used relatively small architectures, and found pronounced improvements with small amounts of data especially when using their largest architecture of ResNet-50 (1$\times$)~\cite{he2016deep} (which is the smallest architecture we considered in our study).

Transfer learning from in-distribution data can help alleviate the domain mismatch issue. For example,~\cite{chen2019med3d,heker2020joint,liang2020transfer,pmlr-v102-geyer19a} report performance improvements when pretraining on labeled data in the same domain. However, this approach is often infeasible for many medical tasks in which labeled data is expensive and time-consuming to obtain. Recent advances in self-supervised learning provide a promising alternative enabling the use of unlabeled medical data that is often easier to acquire.

\paragraph{Self-supervised and contrastive learning.}  Preliminary works in self-supervised representation learning focused on the problem of learning representations without labels such that a small linear classifier network operating on these representations could achieve high classification accuracy~ \cite{doersch2015unsupervised,gidaris2018unsupervised,noroozi2016unsupervised,zhang2016colorful}. Contrastive self-supervised methods such as instance discrimination~\cite{wu2018unsupervised}, CPC~\cite{henaff2019data,oord2018representation}, Deep InfoMax~\cite{hjelm2018learning}, Ye \etal~\cite{ye2019unsupervised}, AMDIM~\cite{bachman2019learning}, CMC~\cite{tian2019contrastive}, RELIC~\cite{mitrovic2020representation}, MoCo~\cite{chen2020improved,he2020momentum}, PIRL~\cite{misra2020self}, SimCLR~\cite{chen2020simple,chen2020big} and SwAV~\cite{caron2020unsupervised} among others were the first to achieve linear classification accuracy approaching that of end-to-end supervised training. Recently, these methods have been harnessed to achieve dramatic improvements in label efficiency for semi-supervised learning. Specifically, one can first pretrain in a task-agnostic manner with self-supervised learning using all data, and then perform task-specific fine-tuning on the labeled subset with a standard supervised objective~\cite{chen2020simple,chen2020big,henaff2019data}.~\cite{chen2020big} show that this approach benefits from large (high-capacity) models for pretraining and fine-tuning, but after a large model is trained, it can be effectively distilled to a much smaller model with little loss in accuracy.

Although self-supervised learning has only recently become viable on standard image classification datasets, it has already seen some application within the medical domain. While some works have attempted to design domain-specific pretext tasks~\cite{bai2019self,spitzer2018improving,zhuang2019self,zhu2020rubik}, other works concentrate on tailoring contrastive learning to medical data~\cite{chaitanya2020contrastive,he2020sample,li2021imbalance,liu2019align,zhou2020comparing,soni2021contrastive}. Most closely related to our work,~\cite{sowrirajan2021moco} explore the use of Momentum Contrast (MoCo)~\citep{he2020momentum} pretraining for classification of the CheXpert dataset through linear evaluation.  In addition to task and medical-domain specific contrastive learning, several recent publications investigate semi-supervised learning for medical imaging tasks using self-supervised pretrained models (\eg~\cite{liu2020semi,wang2020focalmix,zhang2020contrastive,truong2021transferable}). When compared to these efforts, our study proposes a novel method, REMEDIS, that leverages both large scale pretraining and self-supervision and we rigorously evaluate and quantify the impact of our proposed method in varied and challenging clinical settings, thereby demonstrating its clear potential for accelerating the life-cycle of medical imaging AI development and deployment and facilitating its widespread uptake in the real world.

\section{Data Availability} 
The datasets from Northwestern Medicine and Apollo Hospitals were used under license for the current study, and are not publicly available. Applications for access to the Optimam database can be made using this \href{https://medphys.royalsurrey.nhs.uk/omidb/getting-access/}{webform}. The de-identified tele-dermatology data used in this study are not publicly available due to restrictions in the data-sharing agreement. 
\textcolor{black}{The unlabeled dataset used for DME classification is de-identified data from EyePACS Inc. Interested researchers should contact \href{jcuadros@eyepacs.com}{J.C.} to enquire about access to EyePACS data and approach the \href{https://www.research.va.gov/resources/ORD_Admin/ord_contacts.cfm}{Office of Research and Development} to enquire about access to VA data. The rest of annotated data for in-distribution and out-of-distribution DME classification tasks are collected at Rajavithi Hospital Thailand and Lions Eye Institute and are not publicly available due to restrictions in the data-sharing agreement.} Data used in evaluation and pretraining of the Chest X-ray condition classification, including \href{https://physionet.org/content/mimic-cxr/2.0.0/}{MIMIC-CXR}, \href{https://stanfordmlgroup.github.io/competitions/chexpert/}{CheXpert}, and \href{https://www.kaggle.com/nih-chest-xrays/data}{ChestX-ray14} are publicly available. Data used for in-distribution fine-tuning and evaluation of pathology metastases detection is publicly available on the \href{https://camelyon16.grand-challenge.org/Data/}{CAMELYON} challenge website. The Cancer Genoma Atlas (TCGA) data has been used for pretraining of both pathology metastases detection and pathology survival prediction task which is available at \href{http://cancergenome.nih.gov}{NIH website}. The rest of the data used in pathology tasks are not publicly available due to restrictions in the data-sharing agreement. Moreover, ImageNet-1K (ILSVRC2012)~\cite{russakovsky2015imagenet} has been used for pretraining of baseline supervised models and ImageNet-21K has been used for pretraining of BiT-M models. Both of these are publicly available at \href{https://image-net.org/download.php}{ImageNet website}. BiT-L models have been trained on the JFT-300M~\cite{sun2017revisiting} dataset and are not publicly available due to restrictions in the data-sharing agreement.

\section{Code Availability} 
Several major components of our work are available in open source repositories such as the \href{https://www.tensorflow.org}{Tensorflow} library. The code-base and pretrained weights used for self-supervised pretraining are available at \href{https://github.com/google-research/simclr}{SimCLR GitHub repository}. The code-base and pretrained weights for BiT models are available at \href{https://github.com/google-research/big_transfer}{Big Transfer GitHub}. All experiments and implementation details are described in sufficient detail in the Method and Supplementary Material sections to support replication with non-proprietary libraries. The code-base used for our comparison to ResNet-RS was based on the~\href{https://github.com/tensorflow/tpu/tree/master/models/official/resnet/resnet_rs}{ResNet-RS GitHub repository}.

\setlength\bibitemsep{0pt}
\printbibliography[heading=subbibliography]
{
\linespread{0.8}\footnotesize
\vspace{5pt}
\textbf{Acknowledgments}  
This project was an extensive collaboration between Google Brain and the Google Health AI Team. We would like to thank Zoubin Ghahramani for his valuable feedback and continuous support through the course of the project. We thank Maithra Raghu, Nenad Tomašev, Jonathan Krause, Douglas Eck, and Michael Howell for their valuable feedback in improving the quality of the work. Additionally, we would like to thank Jakob Uszkoreit, Jon Deaton, Varun Godbole, Marcin Sieniek, Shruthi Prabhakara, Daniel Golden, Dave Steiner, Xiaohua Zhai, Andrei Giurgiu, Tom Duerig, Christopher Semturs, Peggy Bui, Jay Hartford, Sunny Jansen, Shravya Shetty, Terry Spitz, Dustin Tran, Jieying Luo, Olga Wichrowska, and Abbi Ward for their support throughout this project. 

We would like to acknowledge multiple contributors to this international project: Rajavithi Hospital Thailand, Lions Eye Institute and Derbarl Yerrigan Health Service, Western Australia, Stanford Center for Artificial Intelligence in Medicine and Imaging, MIT Laboratory for Computational Physiology and PhysioNet, \href{https://www.nih.gov/news-events/news-releases/nih-clinical-center-provides-one-largest-publicly-available-chest-x-ray-datasets-scientific-community}{NIH Clinical Center}.  We also would like to thank our collaborators at DermPath AI, Apollo Hospitals, and \textcolor{black}{EyePACS} for their support of this work. We would like to thank our collaborators at Northwestern medicine and all members of the Etemadi Research Group for their support of this work. 

The images and data used in this publication are derived from the \href{https://pubs.rsna.org/doi/abs/10.1148/ryai.2020200103?journalCode=ai}{Optimam database} the creation of which was funded by Cancer Research UK. Part of the retinal image data set was provided for the study by Sankara Nethralaya, Chennai, India.

The results published here are in whole or part based upon data generated by The Cancer Genome Atlas (TCGA) managed by the NCI and NHGRI. Information about TCGA can be found at the \href{http://cancergenome.nih.gov}{NIH website}. This study also utilized archived and anonymized pathology slides, clinicopathologic variables, and outcomes from the Institute of Pathology and the Biobank at the Medical University of Graz. Furthermore, the current study also used pathology slides from the \href{https://camelyon16.grand-challenge.org/Data/}{CAMELYON} challenge.

\vspace{5pt}
\textbf{Author Contributions} S.A., J.F., L.C., V.N., N.H., A.K., M.N., S.K., T.C., B.M., P.M., S.S.M., E.W., C.C., and G.H. contributed to the conception and design of the work; S.A., L.C., J.F., V.N., A.K., B.B., P.B., E.W., C.C., Yuan Liu, S.M., A.L., J.W., M.W., Z.B., A.G.R., D.W., L.P., G.C., and J.K., contributed to the acquisition of the data; S.A., L.C., J.F., S.B., B.M., V.N. majorly contributed to the evaluation of the work and S.A., L.C., J.F., V.N., N.H., A.K., M.N., S.B., S.K., T.C., B.B., D.W., D.F., G.C., and M.E. contributed to analysis and interpretation of the data; S.A., L.C., J.F., V.N., N.H., A.K., M.N., S.K., E.W., P.M., S.S.M., and M.E. contributed to drafting and revising the manuscript.

\vspace{5pt}
\textbf{Competing Interests} This study was funded by Google LLC and/or a subsidiary thereof (‘Google’). J.F., L.C., S.A., V.N., N.H., A.K., M.N., B.M., S.B., P.M., S.S.M., S.K., T.C., B.B., P.B., E.W., C.C., Yuan Liu, Yun Liu, S.M., A.L., J.W., M.W., Z.B., A.G.R., U.T., D.W., D.F., L.P., G.C., J.K., and G.H. are employees of Google and may own stock as part of the standard compensation package. M.E. received funding from Google to support the research collaboration. 


\vspace{5pt}
\paragraph{}
}
\balance
\clearpage
\end{refsection}

\begin{refsection}
\clearpage
\onecolumn

\renewcommand{\thesection}{B.\arabic{section}}
\renewcommand{\thefigure}{B.\arabic{figure}}
\renewcommand{\thetable}{B.\arabic{table}} 
\renewcommand{\theequation}{B.\arabic{equation}} 

\setcounter{section}{0}
\setcounter{figure}{0}
\setcounter{table}{0}
\setcounter{equation}{0}

\noindent \textbf{\LARGE{Appendix B: Additional Results}}\\
\normalfont

In the following sections, we report further experiments and analysis to understand the performance of our proposed method, REMEDIS. This includes ablation studies to analyze the benefits of the different components underlying REMEDIS, comparison with several supervised baselines and other approaches to leveraging unlabeled data. 
Given the sensitivity of the medical domain and the importance of ensuring AI development methods do not propagate existing health equity disparities, we conduct a detailed subgroup analysis in the dermatology and the mammography setting.
Furthermore, we also include results on a non-classification task and visualize the learned representations and list detailed t-test statistics for all our experiments.  

\begin{figure*}[!tbh]
    \centering
    \includegraphics[width=0.55\textwidth]{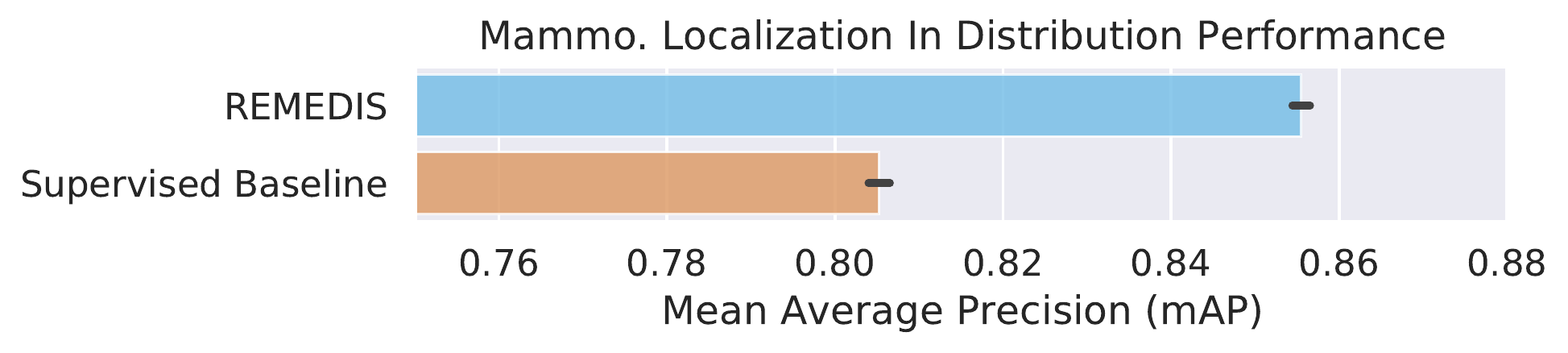}
    \vspace{6pt}
    \caption{\small Mammography localization performance, measured using the mean average precision of the localized cancer. 95\% confidence intervals were calculated by running each label fraction and experiment ten times and is shown with the shaded area and error bars. A two-sided $t$-test was also done, which showed a $p$-value less than 0.001. Specifically, the T-Statistic was 57.14, the $p$-value is 3.54e-52, and the degree of freedom is 57.}
    \label{fig:appendix-localization-detailed}
    \vspace{-0cm}
\end{figure*}

To summarize:

\begin{itemize}[]
    \item Figures: 
    \begin{itemize}
    \item  Figure~\ref{fig:appendix-main-results-img}: Study of the performance of our approach \vs the standard supervised baseline.
    \item  Figures~\ref{fig:appendix-main-results-img} and~\ref{fig:appendix-data-efficiency-all} Study of the performance of our approach \vs supervised pertaining strategies. 
    \item  Figure~\ref{fig:appendix-ablations}: Ablation study of the our proposed approach components and their contribution.  
    \item  Figure~\ref{fig:appendix-ablations-derm-bit}: Ablation of family of BiT~\cite{kolesnikov2019big} models.
    \item  Figure~\ref{fig:appendix-self-training}: Comparison with self-training~\cite{chen2020big, xie2020self}.
    \item  Figure~\ref{fig:appendix-fairness}: Performance analysis across subgroups.
    \item  Figure~\ref{fig:appendix-localization-detailed}: Results on mammography localization task.
    \item  Figure~\ref{fig:appendix-tsne}: t-SNE visualization of the learned representations.
    \item  Figures~\ref{fig:appendix-zero-shot-detailed-jft}~--~\ref{fig:appendix-label-eff-detailed-img}: Detailed results. 
    \end{itemize}
    \item Tables: 
    \begin{itemize}
    \item Tables~\ref{tab:performance-table-best-vs-best}~--~\ref{tab:relative-out-of-distribution-img}: Detailed performance gains.
    \item Tables~\ref{tab:appendix-ablations-stats-all-task-zero-shot}~--~\ref{tab:appendix-ood-stats-img}: Detailed $t$-test statistics for all experiments. 
    \end{itemize}

\end{itemize}

\section{Performance of Standard Supervised Transfer Learning Strategy}
\begin{figure*}[]
\small
    \centering
    \includegraphics[width=1.0\textwidth]{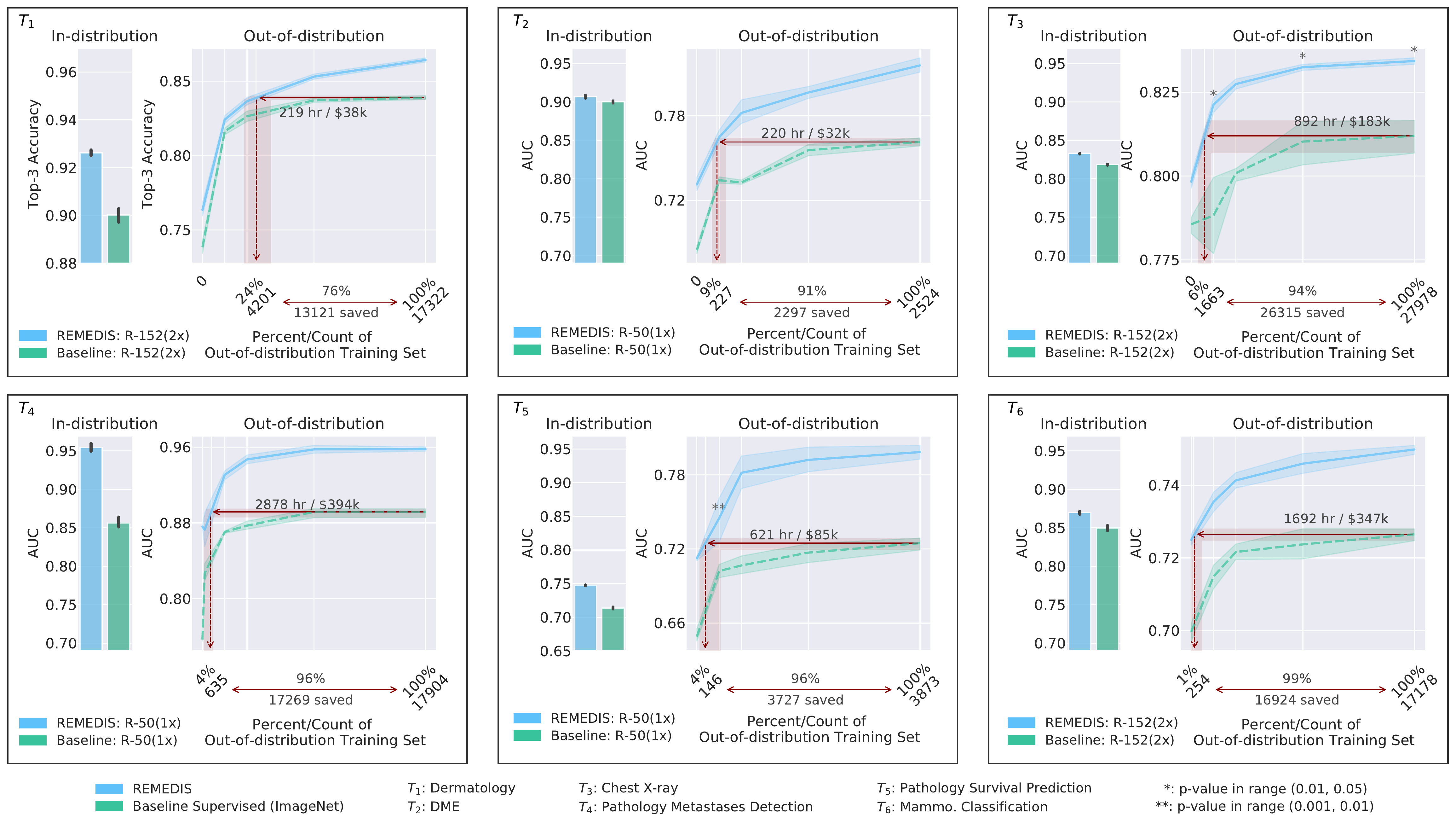}
    \vspace{6pt}
    \caption{\textcolor{black}{\textbf{REMEDIS vs. Standard Supervised Baseline}}. Overview of the results demonstrating overall performance and generalization of our proposed strategy as well as the standard supervised baseline pretrained on ImageNet-1K. We observed significantly improved out-of-distribution generalization and significant reduction in need for labeled medical data when using our proposed approach. 95\% confidence intervals were calculated by running each label fraction and experiment up to ten times and intervals are shown using the shaded area and error bars. A two-sided $t$-test was also done for each experiments. If no * is shown, the $p$-value is less than 0.001, otherwise, the $p$-value is as indicated. The red lines indicate the amount of data that REMEDIS needs to match the highest standard supervised baseline performance when simulated in a new OOD clinical deployment setting and summarize the amount of annotated data and clinician hours potentially saved by using REMEDIS for each medical task.}
    \vspace{-0pt}
    \label{fig:appendix-main-results-img}
\end{figure*}

In addition to the strong supervised baseline discussed in Fig.~\ref{fig:main-results} we also evaluate our method against the standard supervised strategy which is defined as transfer learning using models pretrained on 1M natural images from ImageNet-1K dataset. Figure~\ref{fig:appendix-main-results-img} shows the overview of the results demonstrating overall performance and data-efficient generalization of the proposed self-supervised learning strategy, REMEDIS in compare to the standard supervised baseline pretrained in ImageNet-1K and finetuned for the specific medical task. We observed significantly improved out-of-distribution generalization and significant reduction in need for labeled medical data when using our proposed approach. REMEDIS exhibits significantly improved in-distribution performance with up to 11.5\% relative improvement is diagnostic accuracy in comparison to the standard supervised baseline. Furthermore, REMEDIS leads to strong data-efficient generalization, matching the standard supervised baseline using 1\% to 31\% of retraining data from a new clinical development setting across tasks. This can translate to a significant reduction in the retraining data and time required to deploy medical AI at scale and accelerate the development life-cycle of these AI models. 

\section{Ablation Studies}

\subsection{Contribution of large-scale pretraining data}

In Figure~\ref{fig:main-results}, Figure~\ref{fig:appendix-main-results-img}, Figure~\ref{fig:appendix-label-eff-detailed-jft}, Figure~\ref{fig:appendix-label-eff-detailed-img}, Figure~\ref{fig:appendix-zero-shot-detailed-img}, and Figure~\ref{fig:appendix-zero-shot-detailed-jft}, we report the comparison of REMEDIS with the widely used supervised pretraining baseline. To investigate the contribution of large-scale pretraining data, here, we also separately compares and contrast the overall performance of REMEDIS \vs these baselines. 

\begin{figure*}[]
\small
    \centering
    \includegraphics[width=0.9\textwidth]{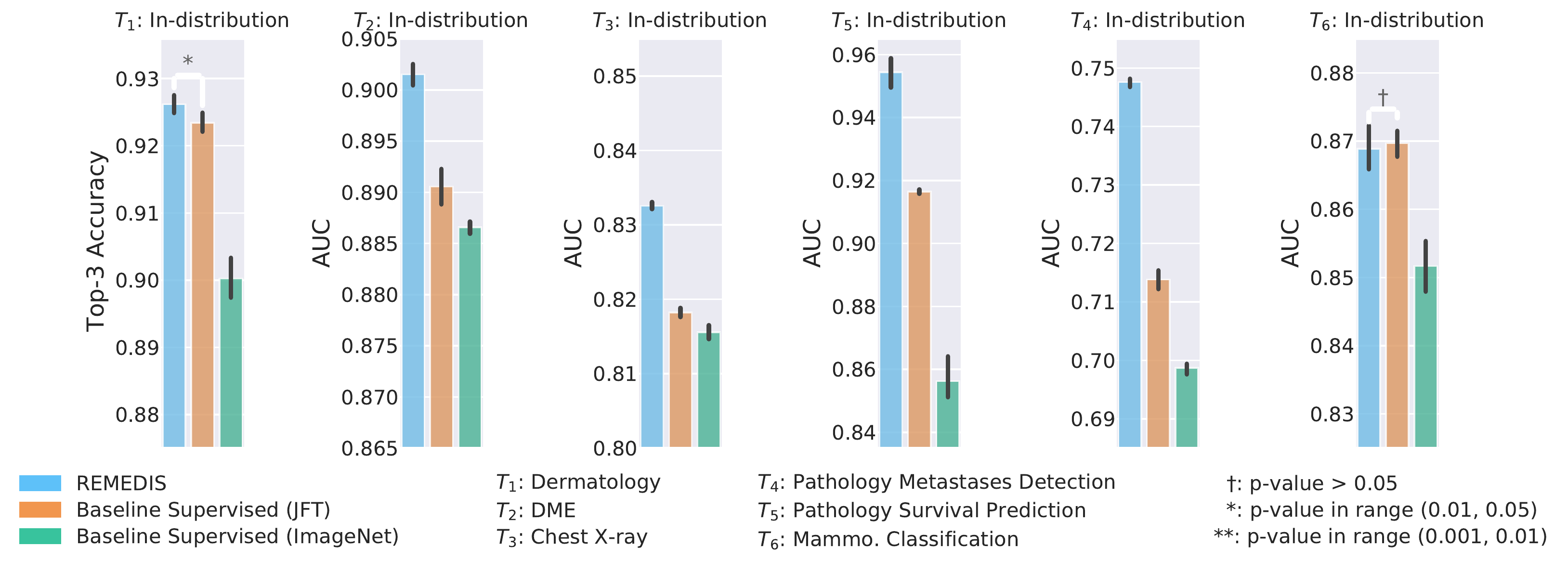}
    \vspace{-0pt}
    \caption{\textcolor{black}{\textbf{Overall in-distribution performance}}. Overview of the results demonstrating overall in-distribution performance of REMEDIS as well as the standard and strong supervised baselines trained using both ImageNet-1K and JFT-300M. We observed significantly improved in-distribution performance using our proposed strategy \vs the standard and strong transfer learning strategies. Moreover, the strong supervised baseline (BiT-L) can provide significantly better in-distribution performance against the standard supervised baseline. 95\% confidence intervals were calculated by running each label fraction and experiment up to ten times and intervals are shown using the error bars. A two-sided $t$-test was also done for each label fraction as well as when computing the in-distribution results. If no * is shown, the $p$-value is less than 0.001, otherwise, the $p$-value is as indicated.}
    \vspace{-0pt}
    \label{fig:appendix-performance-all}
\end{figure*}

The strong supervised baseline and REMEDIS both pretrained on JFT-300M (BiT-L) and relies on large-scale pretraining while the standard supervised baseline has been trained on a much smaller dataset contains only 1M images. Figure~\ref{fig:appendix-performance-all} demonstrates overall in-distribution performance of REMEDIS as well as the supervised baselines trained using both ImageNet-1K and JFT-300M. Moreover, we observe that strong supervised baseline (BiT-L) can provide significantly better in-distribution performance against the standard supervised baseline (BiT-S), shows the benefits of large scale training as it has been reported in~\cite{mustafa2021supervised}. 

\begin{figure*}[]
\small
    \centering
    \includegraphics[width=.9\textwidth]{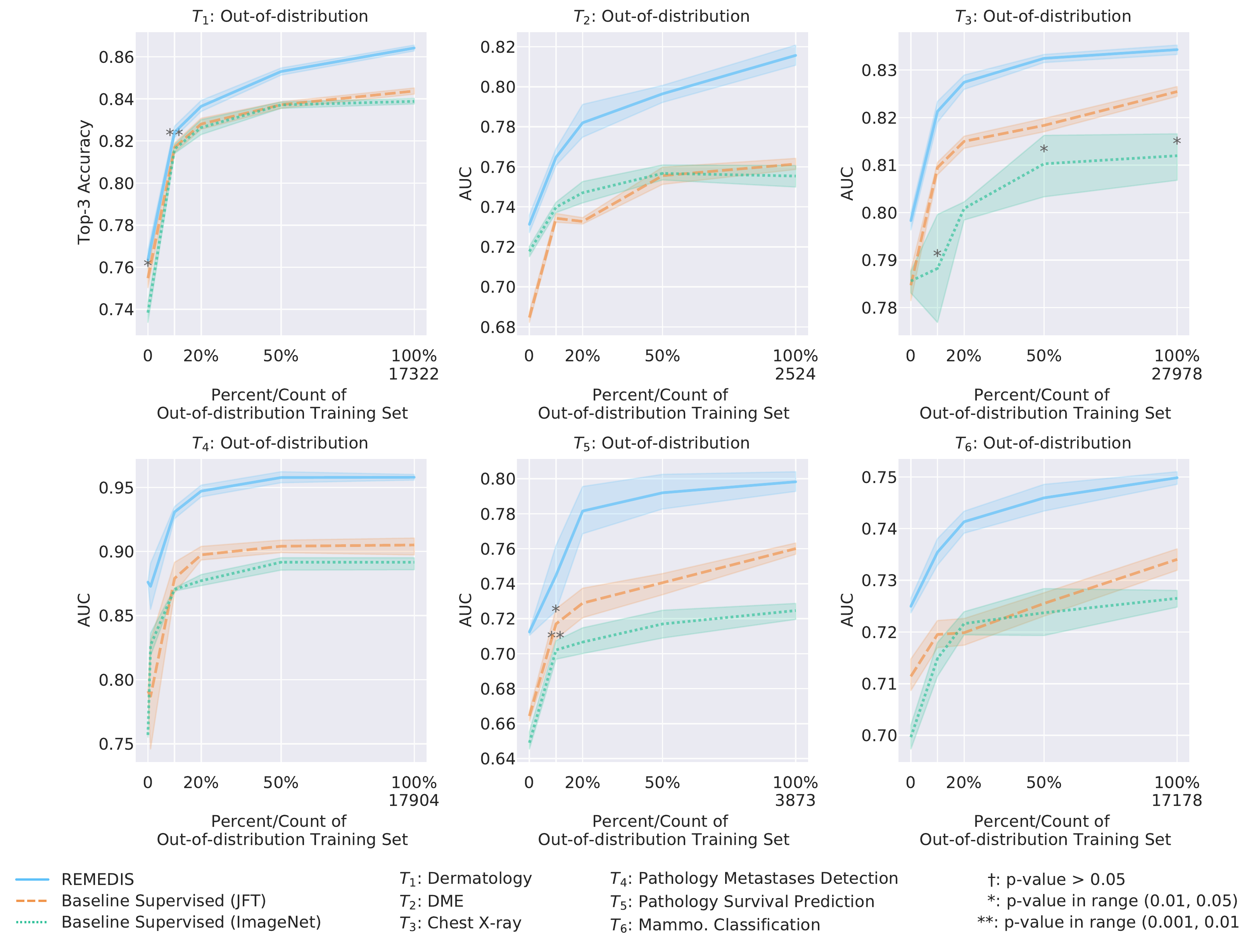}
    \vspace{6pt}
    \caption{\textcolor{black}{\textbf{REMEDIS overall data-efficiency performance}}. Overview of the results demonstrating data-efficient generalization of our proposed self-supervised learning strategy, REMEDIS as well as the standard and strong supervised baseline pretrained using both ImageNet-1K and JFT-300M. We observed significantly improved out-of-distribution generalization and significant reduction in need for annotated medical data when using our proposed approach. Moreover, comparing the data-efficient generalization of the strong and standard pretrained models, often string supervised baseline performs significantly better than the standard supervised baseline in out-of-distribution regime. 95\% confidence intervals were calculated by running each label fraction and experiment up to ten times and intervals are shown using the shaded area. A two-sided $t$-test was also done for each label fraction. If no * is shown, the $p$-value is less than 0.001, otherwise, the $p$-value is as indicated.}
    \vspace{-0pt}
    \label{fig:appendix-data-efficiency-all}
\end{figure*}

As discussed, the large-scale supervised pretraining (BiT-L) represents a strong supervised baseline for medical imaging~\cite{mustafa2021supervised}. Figure~\ref{fig:appendix-data-efficiency-all} shows the overview of the results demonstrating data-efficient generalization of our proposed self-supervised learning strategy, REMEDIS as well as the standard and strong supervised baseline pretrained using both ImageNet-1K and JFT-300M. We observed significantly improved out-of-distribution generalization and significant reduction in need for annotated medical data when using our proposed approach. Moreover, comparing the data-efficient generalization of the strong and standard pretrained models, often string supervised baseline performs significantly better than the standard supervised baseline in out-of-distribution regime. Meanwhile REMEDIS holds a steady significantly better performance against both strong and standard baseline, in some of the tasks such as Dermatology classification ($T_1$) and DME ($T_2$) the superior performance of strong supervise baseline in the data-efficient generalization regime against the standard baseline can shows an unexpected pattern.

\begin{figure*}[!tbh]
\footnotesize
    \centering
    \includegraphics[width=0.9\textwidth]{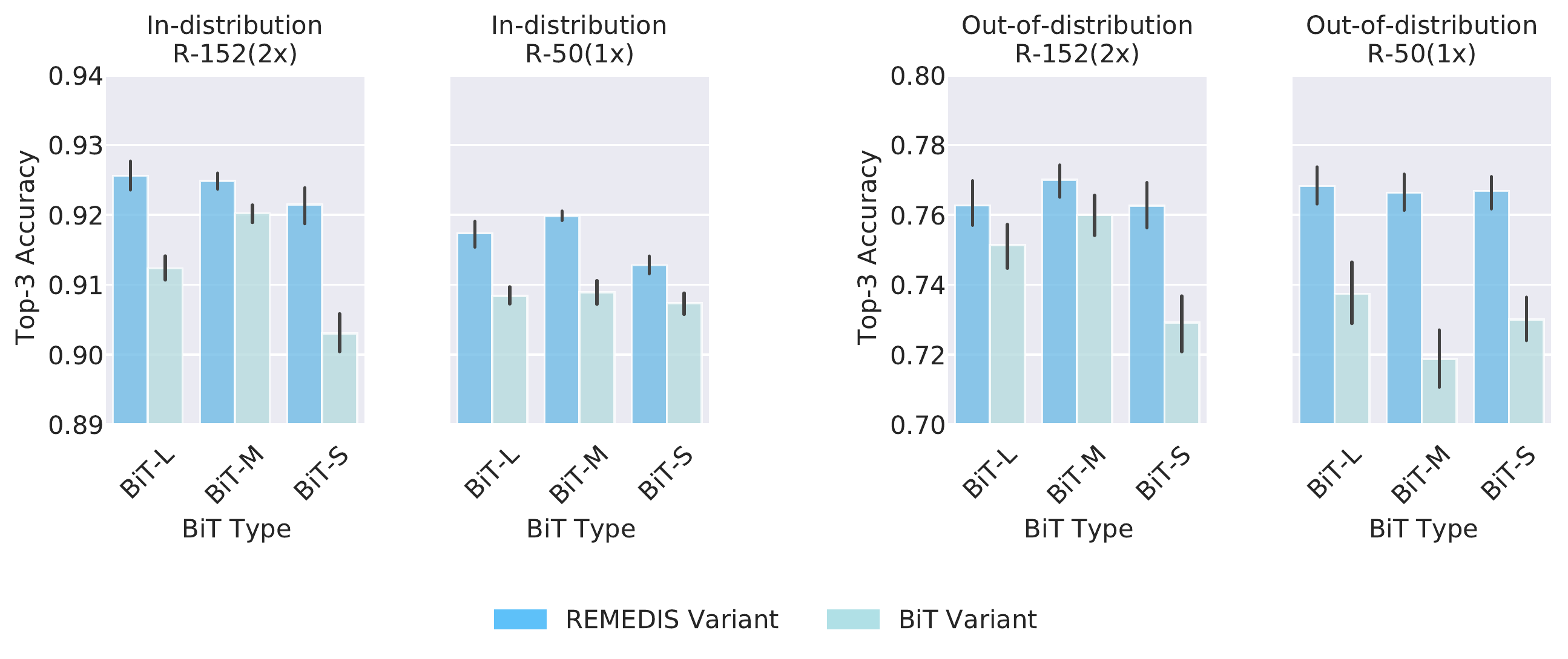}
    \caption{\small Ablation of different BiT models used in both REMEDIS and the BiT Baseline, for the dermatology task, $T_1$. The BiT model type shown is used in both REMEDIS and BiT for this comparison. BiT-L is the largest BiT model, trained on JFT, and is the default BiT model used in REMEDIS. BiT-M is trained on ImageNet 21k, while BiT-S is trained on ImageNet.}
    \label{fig:appendix-ablations-derm-bit}
    \vspace{-0cm}
\end{figure*}

Both Figure~\ref{fig:appendix-performance-all} and Figure~\ref{fig:appendix-data-efficiency-all} indicating that large-scale supervised pretraining is a strong component and is a good starting point for developing medical imaging models~\cite{mustafa2021supervised}. To further investigate the specific significance of large-scale pretraining and how the choice of supervised pretraining impacts REMEDIS, in a new set of experiment we adapt BiT-S, M, and L as our based network and performs the self-supervised training on medical data on top of each of these models. The default REMEDIS method uses BiT-L as the base-network. This direct comparison is only completed on the dermatology task, due to the high computational cost.  The results, shown in figure~\ref{fig:appendix-ablations-derm-bit}, show that the best results tend to be achieved using BiT-L, but that BiT-M can perform competitively. Given that BiT-M is openly available and is not trained on proprietary data, we believe that this is a further good indication that large-scale supervised pretraining is valuable and hope the wider medical AI community leverages this to build medical imaging models. 

\subsection{Contributions of BiT-L and SimCLR}

While our general focus in this study has been to compare REMEDIS with supervised baselines (~Fig.~\ref{fig:main-results} and Fig.~\ref{fig:appendix-main-results-img}), it is also of interest to understand the contributions of the representation learning strategies underlying REMEDIS. To this end, we ran ablation studies in which we investigated and disentangled the contribution of the large-scale pretraining on natural images and self-supervised representation learning on medical images as well as the specific architecture choices. For a fair comparison, in each case we follow the exact same pretraining and fine-tuning protocol as our method to optimize these models.

\begin{figure*}[!tbh]
\footnotesize
    \centering
    \includegraphics[width=0.8\textwidth]{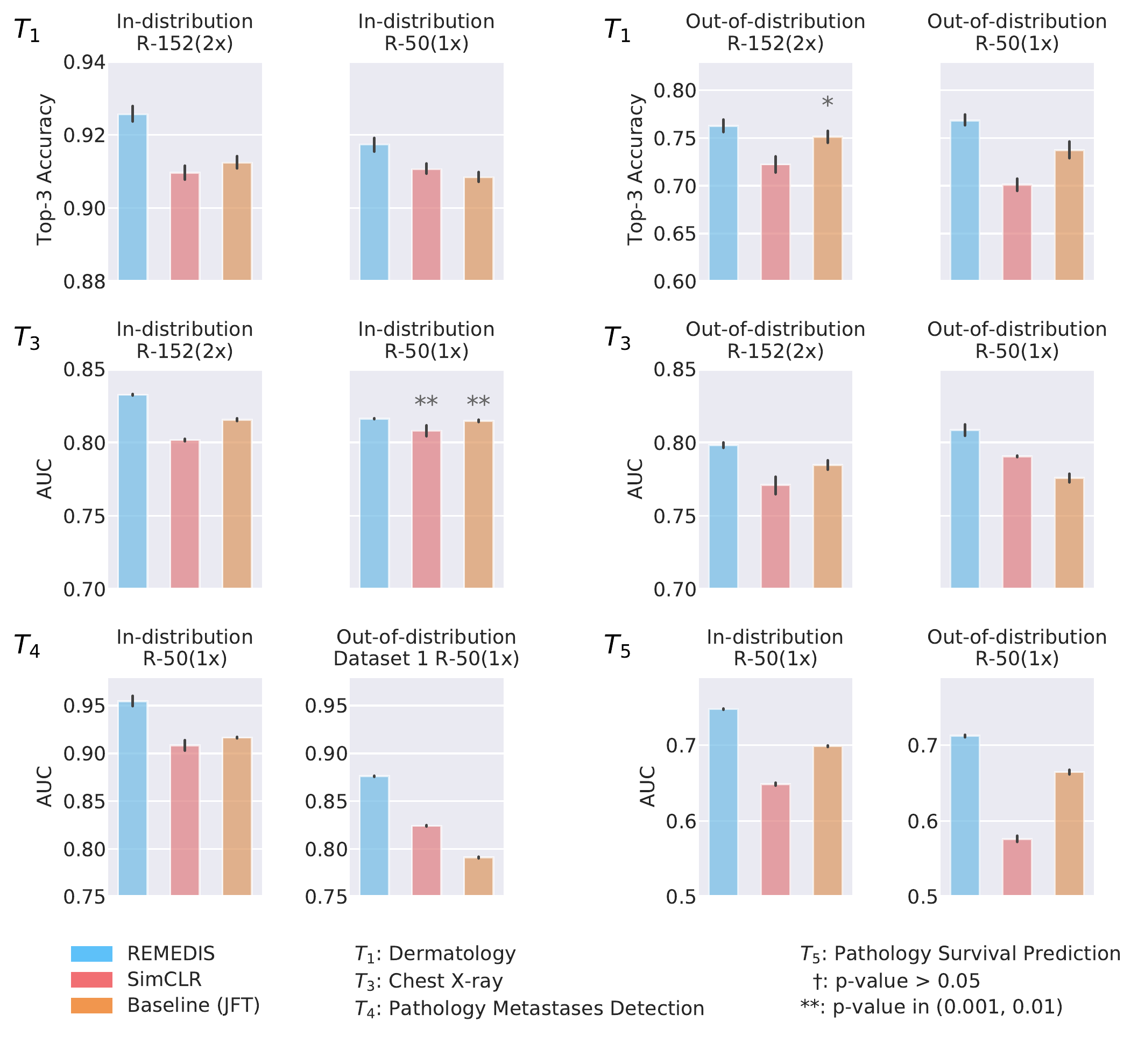}
    \caption{\small Contribution of BiT-L and SimCLR. Both large-scale supervised pretraining and self-supervised pretraining separately provide benefits. A two-sided $t$-test is performed between each baseline model and REMEDIS, and a symbol above the bar being compared to REMEDIS shows the relevant $p$-value range. If no symbol is shown, the $p$-value is less than 0.001. REMEDIS outperforms both of its building components, BiT-L and SimCLR.}
    \label{fig:appendix-ablations}
    \vspace{-0cm}
\end{figure*}

Both SimCLR and BiT-L provide benefits over the widely used supervised ImageNet pretraining baseline for most tasks in both in- and out-of-distribution settings (see figure \ref{fig:appendix-ablations}). While for the larger architecture such as ResNet-152 (2$\times$), BiT can provide performance gains approximately comparable to REMEDIS in some cases, this is not consistent across architectures as well as all the medical imaging tasks considered. This is also aligned with previous observations in~\cite{azizi2021big}. Note that SimCLR results for mammography and DME classification are missing due to the high computational cost.

\section{Comparison with Self-training}
REMEDIS leverages self-supervised pretraining to make use of large amounts of unlabeled medical data for learning high quality representations. However, there are other approaches that enable models to learn from unlabelled data. One such approach is self-training~\cite{xie2020self,chen2020big}. In a typical self-training setting, a teacher network trained on labeled data predicts labels on the unlabeled data. Then, a second student model is fine-tuned on the original labeled data, as well as the predicted labels. We implement this by training a model on $D_{in}$, inferring on $D_{u}$, and then re-training the model on $D_{in}$ and $D_{u}$. We use the predicted probabilities on $D_{u}$ as soft labels, and separately sweep over hyper-parameters for the teacher and student training. We otherwise do not vary the training set-up. Due to computational constraints and early evidence of superiority of REMEDIS, we only report these numbers for dermatology.

\begin{figure*}[!tbh]
    \centering
    \includegraphics[width=0.9\textwidth]{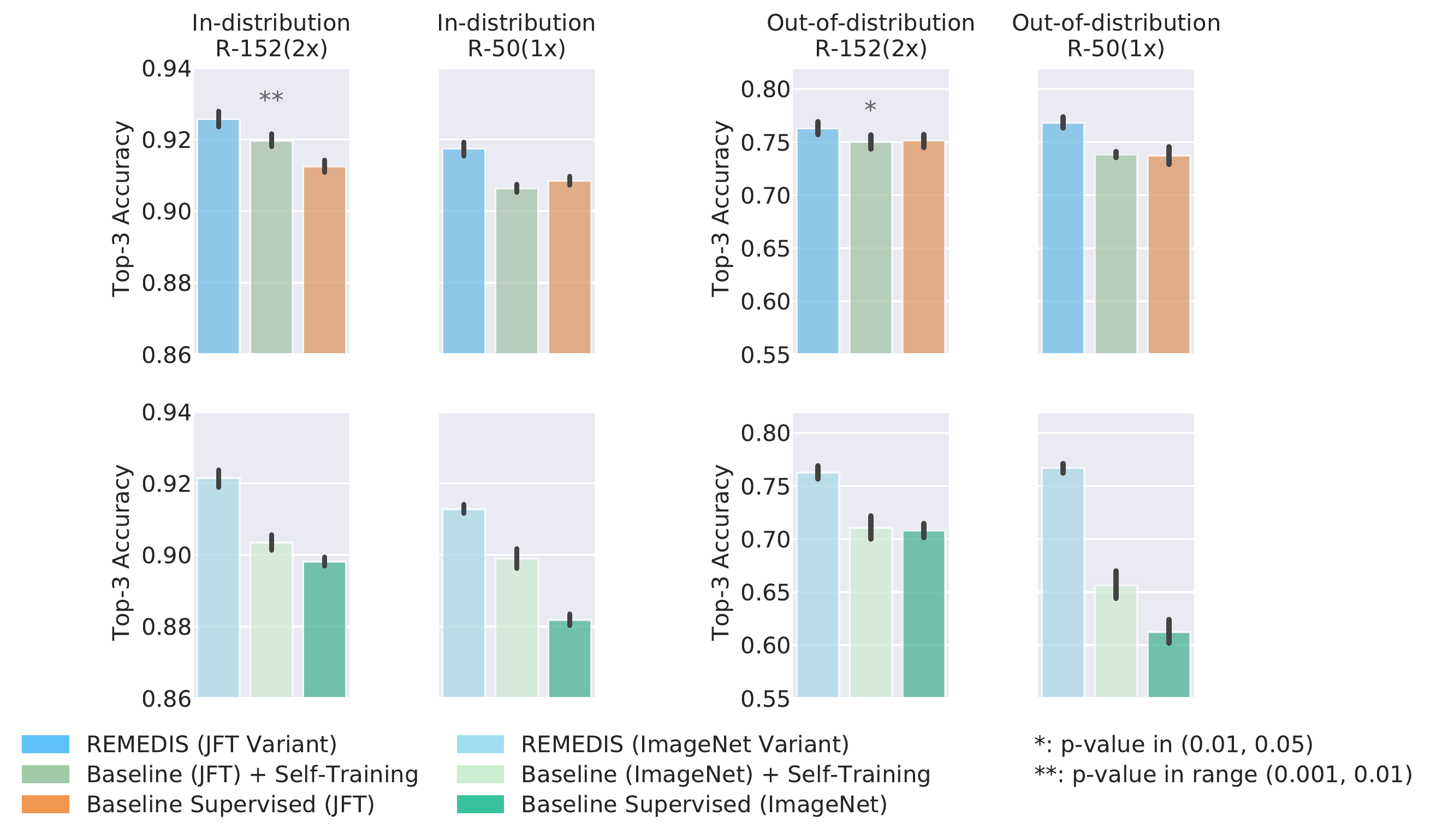}
    \vspace{6pt}
    \caption{\small \textcolor{black}{\textbf{Self-training \vs REMEDIS.}} Comparison of our proposed self supervised method, REMEDIS as well as the strong and standard supervised baseline with self-training approach~\cite{xie2020self} for leveraging unlabeled medical data on the dermatology task. We observe that while self-training can produce high performance models on par with REMEDIS when using large architecture, it does not consistently provide benefits across all the settings considered. This is possibly due to the requirement for a good teacher model when using self-training. Furthermore, self-training requires the representation learning task and the downstream task to be well aligned while contrastive pretraining is agnostic to the downstream task leading to representations that can be generally applied.}
    \label{fig:appendix-self-training}
    \vspace{-0cm}
\end{figure*}

\vspace{6pt}
\textcolor{black}{We performed this self-training cycle using models that start from BiT-L, as the most direct comparison to REMEDIS, as well as with models that start from the standard supervised baseline and the corresponding REMEDIS variant.} The results, shown in figure~\ref{fig:appendix-self-training}, indicate that self-training can produce high quality models on par with REMEDIS especially when using the larger model architecture. However, this is not consistent when using smaller architecture size. Self-training can also degrade performance when applied from the standard supervised baseline, perhaps because self-training relies on the quality of the teacher model. Furthermore, self-training requires the representation learning task and the downstream task to be well aligned while contrastive pretraining is agnostic to the downstream task leading to representations that can be generally applied. Nevertheless, we believe self-training is a promising approach and should be considered when appropriate for developing medical imaging AI using unlabeled data.

\section{Performance Analysis Across Subgroups}
Given the importance of fairness in AI, when using pretrained representations for developing medical imaging AI, we are interested in approximate parity of performance across target subgroups of interest so as to ensure the models are not amplifying existing health disparities. More specifically, for the deployment of such models in clinical settings, it is important to evaluate them comprehensively across protected subgroups. This can be of particular concern when leveraging large-scale pretraining datasets, as they may be biased towards certain subgroups without the knowledge of the model developer~\cite{asano21pass}.

\vspace{6pt}
Thus, we also investigated the performance distribution across different subgroups of interest. We focused on subgroups in the dermatology and mammography task where we have access to metadata to disentangle them. We are particularly interested in how the introduction of large-scale pretraining, or self-supervised pretraining, affects performance across these clinically-relevant subgroups.

\begin{figure*}[!tbh]
    \centering
    \includegraphics[width=0.49\textwidth]{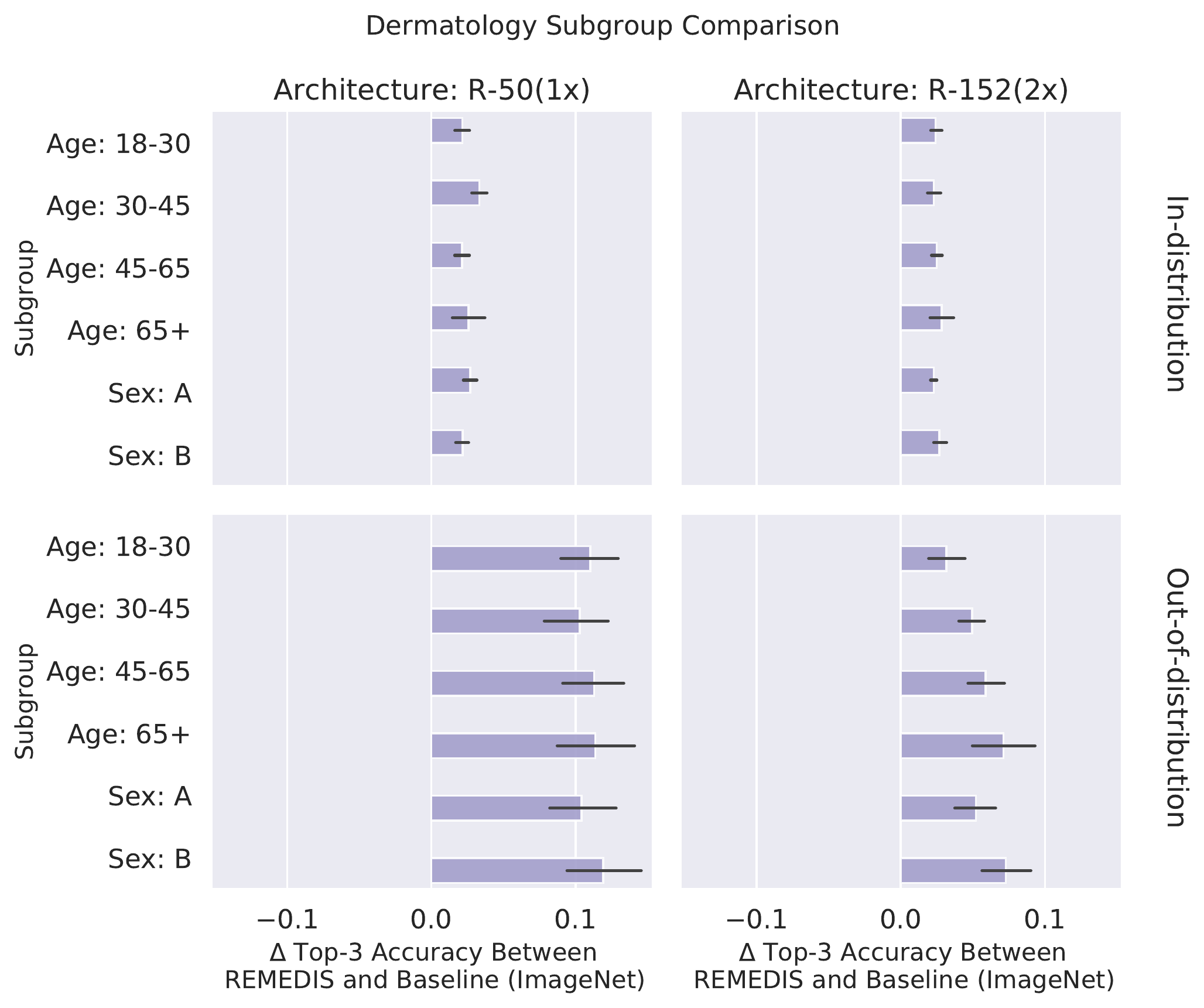} \hfill
    \includegraphics[width=0.49\textwidth]{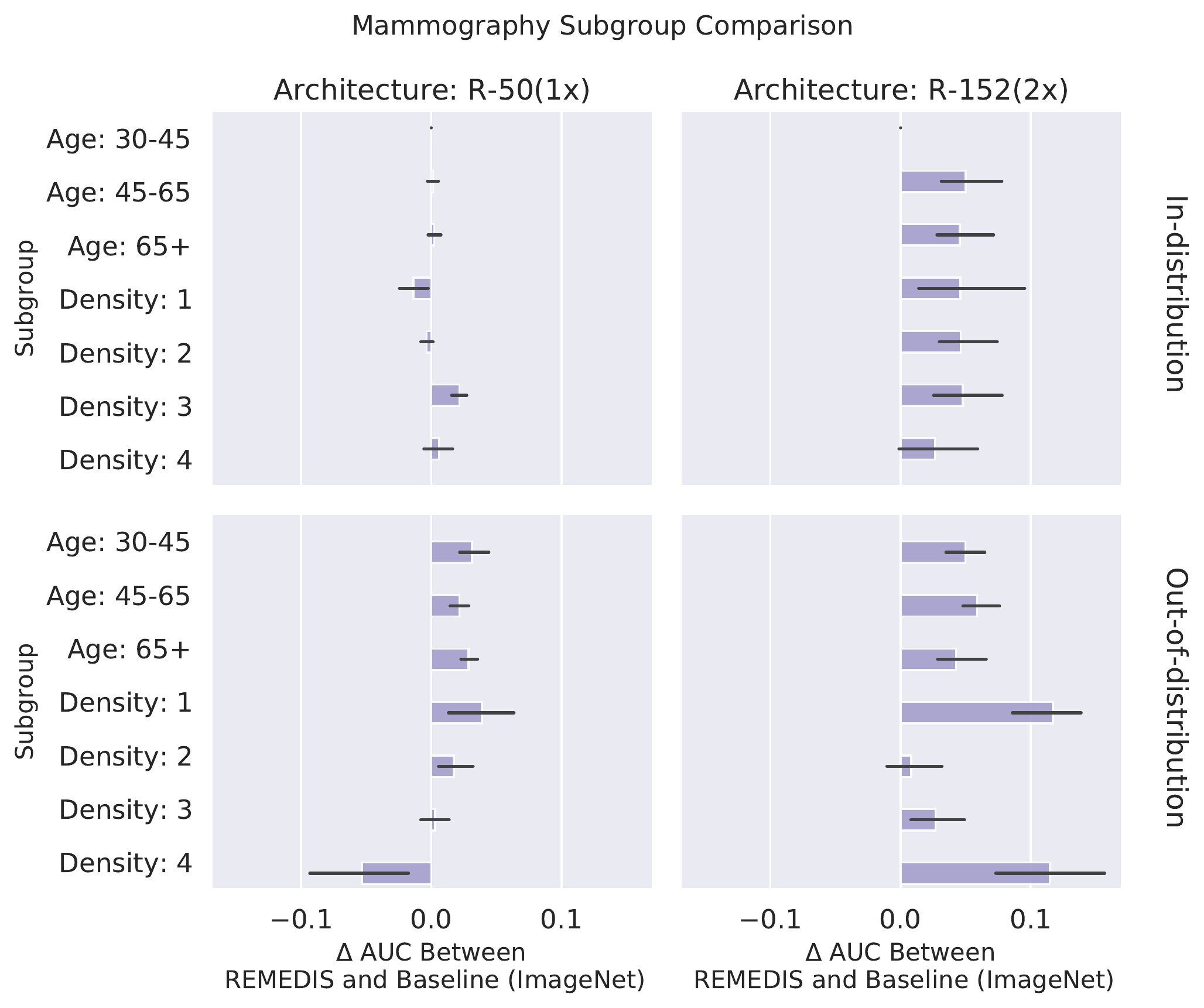} 
    \includegraphics[width=0.49\textwidth]{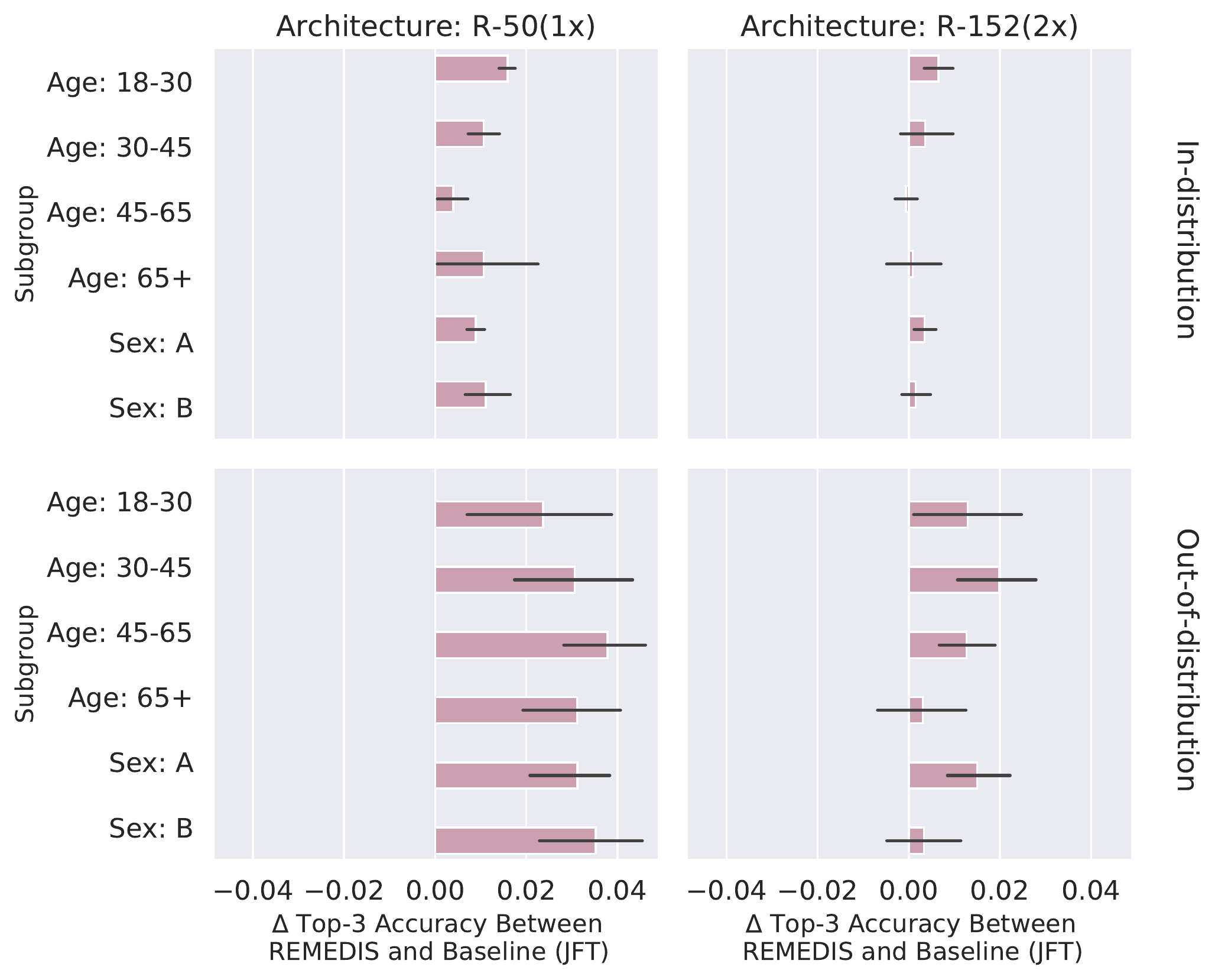} \hfill
    \includegraphics[width=0.49\textwidth]{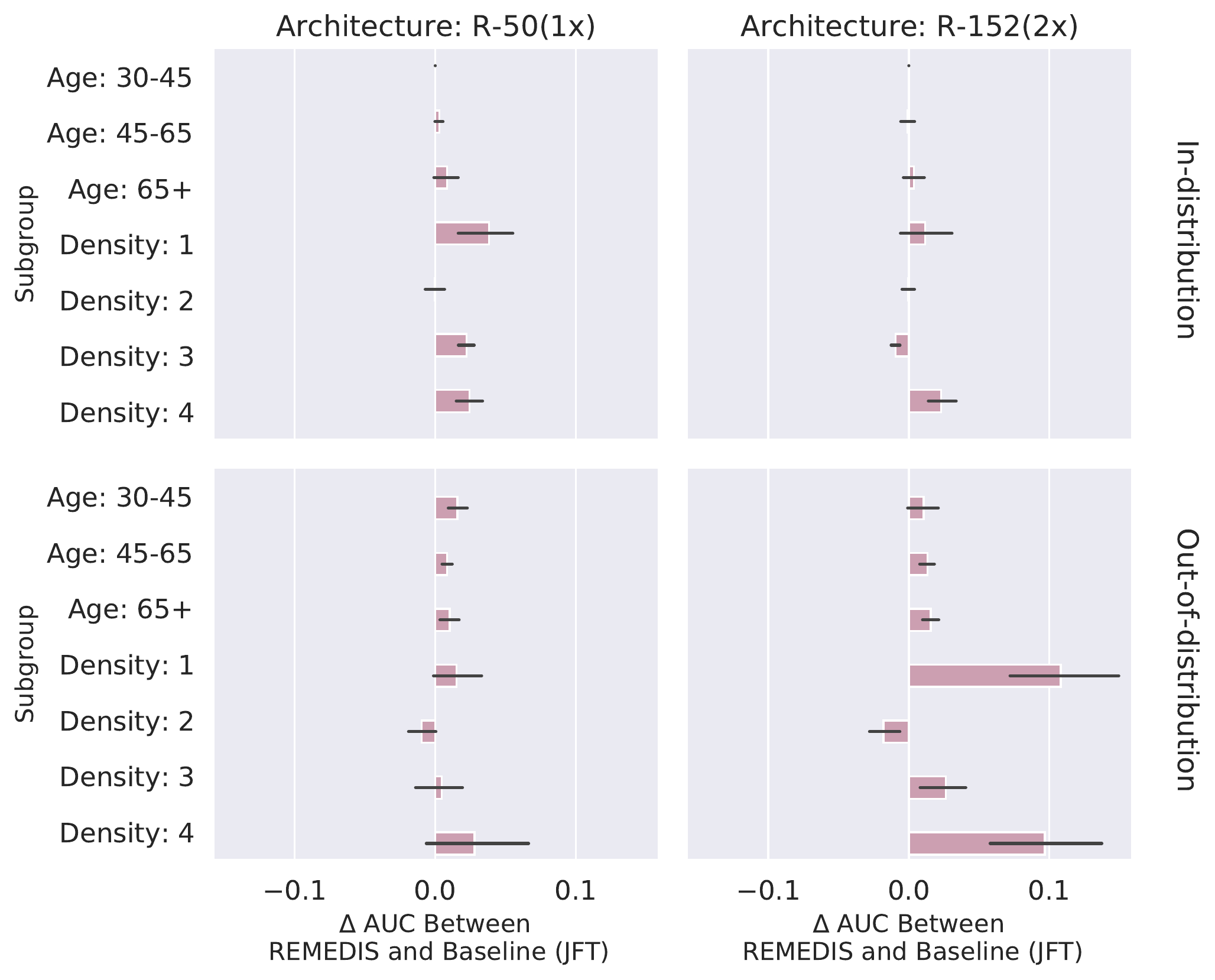}
    \vspace{6pt}
    \caption{\textcolor{black}{\textbf{Performance analysis across subgroups.}} \small Comparison of REMEDIS and the supervised baseline on subgroups of interest in the dermatology and mammography tasks. 95\% confidence intervals were calculated by running each label fraction and experiment ten times and is shown with the error bars. In particular, we note that our method leads to improvement consistently across all subgroups of interest across both the tasks.}
    \label{fig:appendix-fairness}
    \vspace{-0cm}
\end{figure*}

\vspace{6pt}
In dermatology, we established subgroups based on age and biological sex. For sex, we considered a binary setting and compared 2,564 and 1,505 cases for the in-distribution dataset, and 3,153 and 3,486 cases of each sex for the out-of-distribution dataset. For age, we divided the data into four age groups of 18-30, 30-45, 45-65 and 65+ years, which include 1,185, 1,162, 1,495, and 226 cases respectively for the in-distribution data and 186, 702, 2,560, and 3,181 cases for the out-of-distribution dataset. We compared the top-3 accuracy across these groups using the standard and strong supervised baseline and REMEDIS. We observed that while the baseline supervised pretrained model performance drops on some of the subgroups, using intermediate self-supervised pretraining, the model performance is more even across the different subgroups (see figure~\ref{fig:appendix-fairness} (a)). This exploratory experiment suggests that the learnt representations are likely general, and in most of the cases neither picking up spurious correlations during pretraining, nor are they biased towards particular subgroup. 

\vspace{6pt}
For the mammography classification task, we compared subgroups based on age and breast density (which can be correlated with age and ethnicity). The test data is divided into four age groups of 30-45, 45-65 and more than 65+ years of age which include 0, 9,901, and 2,547 cases, respectively, for the in-distribution data and 2,963, 7,109, and 109 cases for the out-of-distribution dataset. The four different density categories include 585, 3,606, 2,314, and 957 cases for the in-distribution dataset, and 109, 612, 741, and 71 cases for the out-of-distribution dataset. Density level of four is associated with a denser breast based on BI-RADS~\cite{american2013acr} assessments. The results, shown in figure~\ref{fig:appendix-fairness} (b), show the distribution of performance across these subgroups. With the exception of a couple of the breast density categories, REMEDIS consistently improves performance. We believe that these subgroup performance disparities are unlikely to be caused by intrinsic biases in the pretraining mechanism, but future work should investigate specific pretraining strategies such as dataset re-sampling to mitigate performance drops.

\section{Mammography Localization Results}
In addition to classification tasks, we also evaluated our method and the baseline on a localization task. For this purpose, we considered the cancer localization task in mammography images ($T_7$). In this task, the goal is to localize cancerous lesions. This task is evaluated using the mean average precision (mAP). Matches between a ground truth and a predicted bounding boxes are considered positive when the intersection-over-union (IOU) is higher than 10\%. The pretraining setup and data were identical to $T_6$'s. Due to the computational complexity of training these models, we report only partial results.  For training the localization model, only positive cases were included. The labels in this task consisted of the coordinates of the bounding boxes derived by human radiologists a-posteriori having access to all mammograms, biopsy results, and radiology text reports. The pretrained CNN backbones were the same as in $T_6$, but an additional feature pyramid network was added on top of the CNN~\cite{tan2020efficientdet}.

\vspace{6pt}
The dataset contains 3,727 cases, including 5,854 ground truth bounding boxes across all mammograms. 2,909 cases were used for training, 158 for tuning, and 660 for test. When using REMEDIS, the model shows a significant improvement over the baseline, moving from a mAP of 0.805 to a mAP of 0.855. We used the Adam optimizer with a exponential learning rate decay in breast cancer localization task where we performed a thorough grid search to select the initial learning rate, decay steps, and decay factor. All of the models were trained for a maximum of 200K steps. For this task, scaled 2048$\times$2048 pixels mammography images go through the augmentation process including random flipping, random shifting, and random color distortion. We selected the learning rate, decay steps, and decay factor after a grid search of three logarithmically spaced learning rates between 10$^{-5.0}$ and 10$^{-4.0}$ and three decay steps in $\{10K, 25K, 50K\}$, and three decay factor in decay steps in $\{0.1, 0.25, 0.5\}$.

\section{t-SNE Visualization of Representations}
\begin{figure*}[!tbh]
\small
    \centering
    \includegraphics[width=0.93\textwidth]{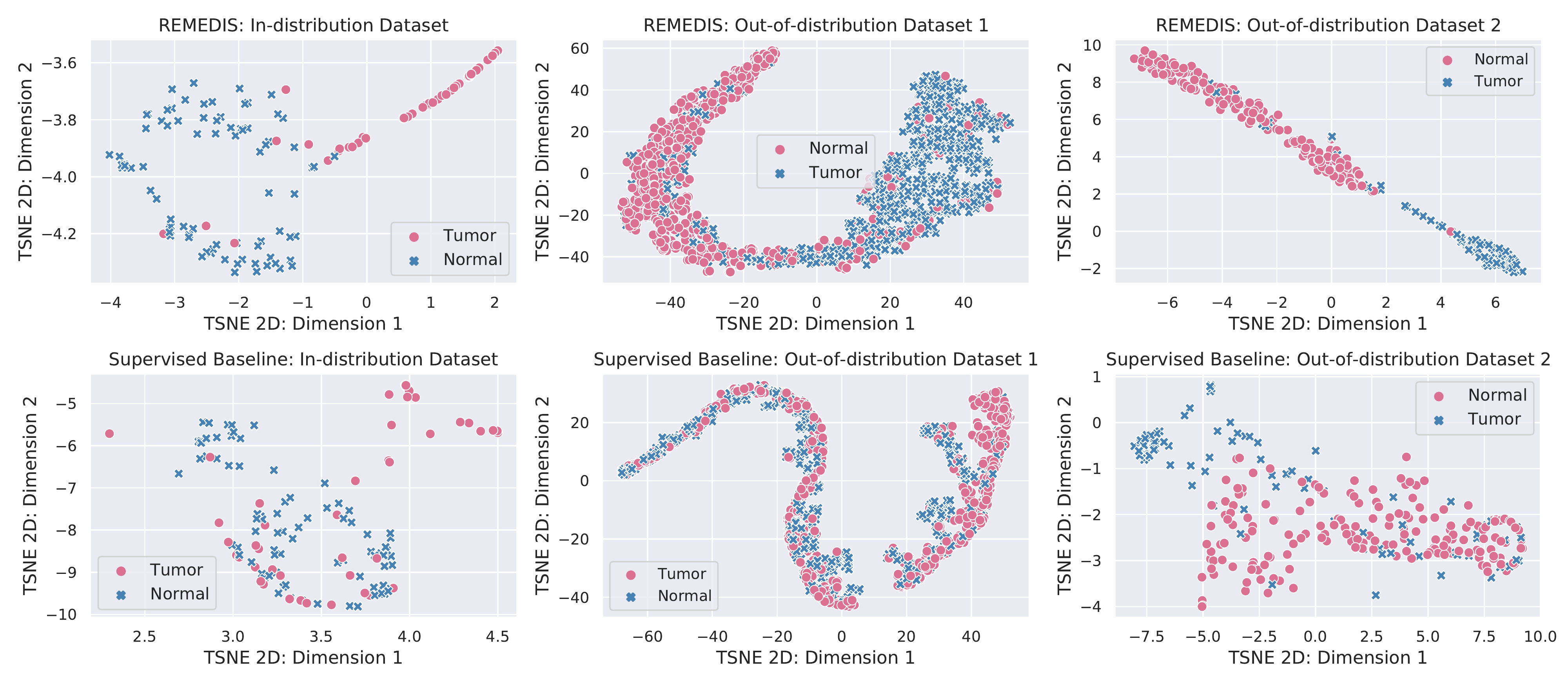}
    \vspace{-0pt}
    \caption{\textbf{t-SNE visualization of representations.} The embedding representations obtained using REMEDIS and supervised baseline models are visualized for both the in-distribution and out-of-distribution datasets of the pathology metastases task. Clusters associated with various classes are better separated in the REMEDIS feature space as compared to the baseline.}
    \vspace{-0pt}
    \label{fig:appendix-tsne}
\end{figure*}

To gain more insight into the high dimensional embedding representations learned by models considered in this study, we use t-SNE visualization~\cite{van2008visualizing}. For this purpose we focus on the pathology metastases task ($T_4$) which has a binary label space and includes two out-of-distribution datasets. The t-SNE visualization of the best REMEDIS and best supervised model representation embeddings obtained from the test in-distribution and out-of-distribution examples of the pathology metastases task ($T_4$) are depicted in Figure~\ref{fig:appendix-tsne}. These models are only fine-tuned with the in-distribution train dataset and not the out-of-distribution data. The binary labels of this task are color-coded. The visualisations (figure~\ref{fig:appendix-tsne}) qualitatively indicate that clusters associated with each class are better separated in the REMEDIS feature space compared to the supervised baseline.

\section{Additional Experimental Results}
\begin{figure*}[!tbh]
    \centering
    \includegraphics[width=0.93\textwidth]{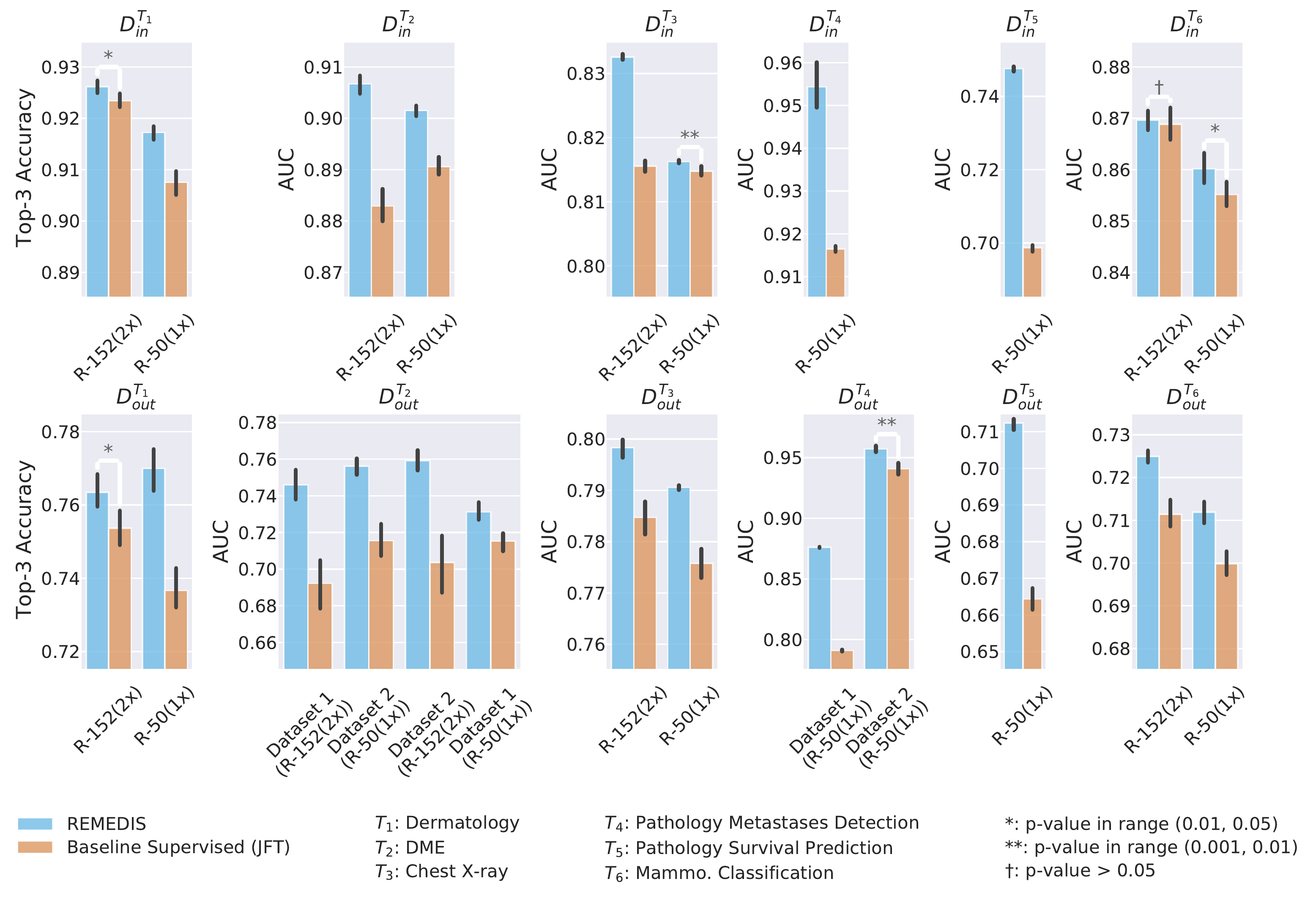}
    \caption{\small{\textcolor{black}{\textbf{Detailed in-distribution and zero-shot out-of-distribution performance for all architectures and all datasets considered in this study.}} We show the superior in-distribution performance of REMEDIS across all tasks and all datasets \vs strong supervised baseline trained on JFT-300M. The 95\% confidence intervals were calculated by running each label fraction and experiment ten times and are shown with the shaded area and error bars. A two-sided $t$-test was also conducted for each pair of results. If no * is shown the $p$-value is less than 0.001, otherwise, the $p$-value is as indicated. Unlike previous visualizations, here we group the results based on the base network architecture not the architecture with the overall best performance.}}
    \label{fig:appendix-zero-shot-detailed-jft}
    \vspace{-0cm}
\end{figure*}

\begin{figure*}[!tbh]
    \centering
    \includegraphics[width=0.93\textwidth]{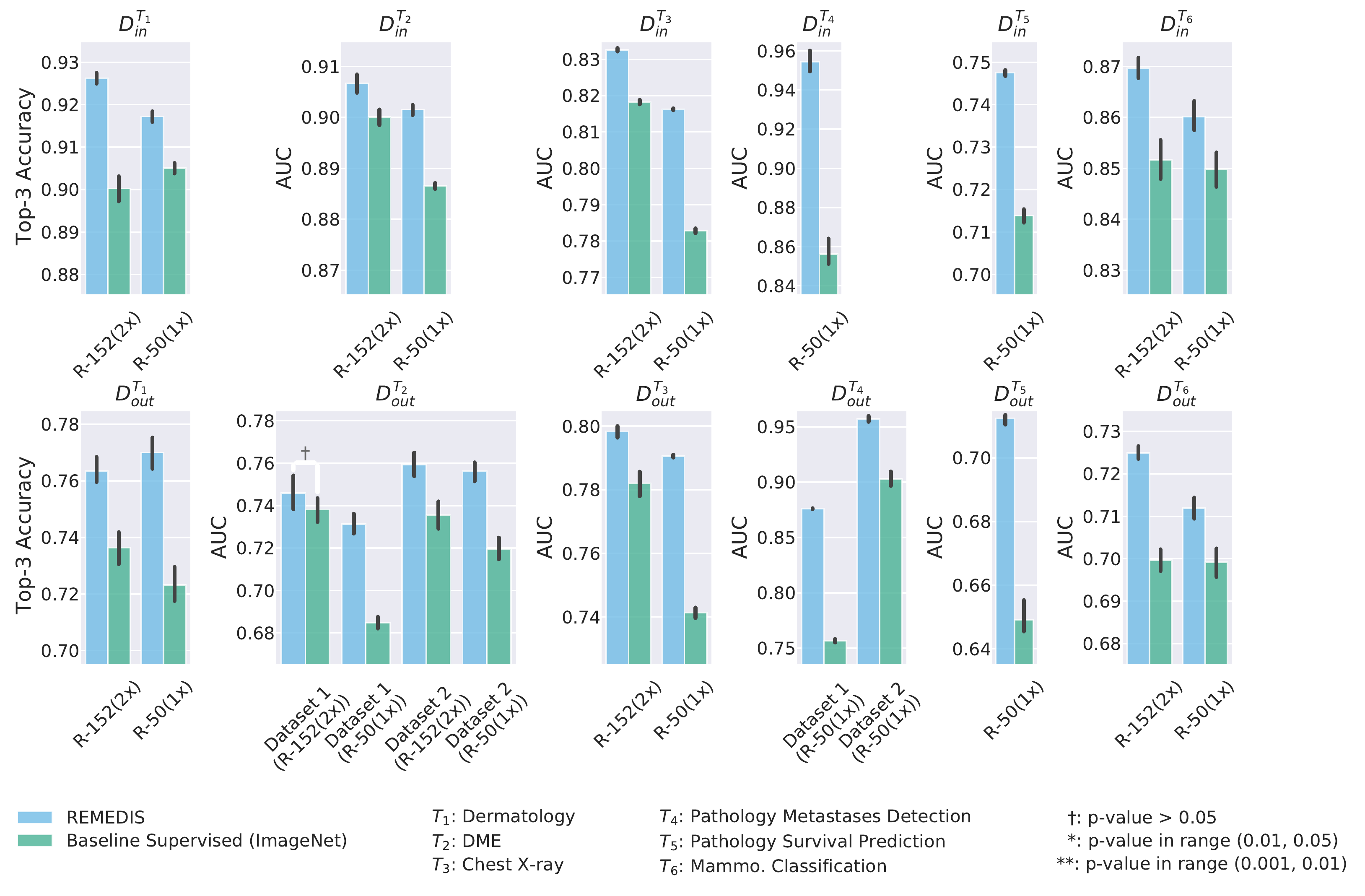}
    \caption{\small{\textcolor{black}{\textbf{Detailed in-distribution and zero-shot out-of-distribution performance for all architectures and all datasets considered in this study.}} We show the superior in-distribution performance of REMEDIS across all tasks and all datasets \vs the standard supervised baseline pretrained on ImageNet-1K. The 95\% confidence intervals were calculated by running each label fraction and experiment ten times and are shown with the shaded area and error bars. A two-sided $t$-test was also conducted for each pair of results. If no * is shown the $p$-value is less than 0.001, otherwise, the $p$-value is as indicated. Unlike previous visualizations, here we group the results based on the base network architecture not the architecture with the overall best performance.}}
    \label{fig:appendix-zero-shot-detailed-img}
    \vspace{-0cm}
\end{figure*}

\begin{figure*}[!tbh]
    \centering
    \includegraphics[width=0.8\textwidth]{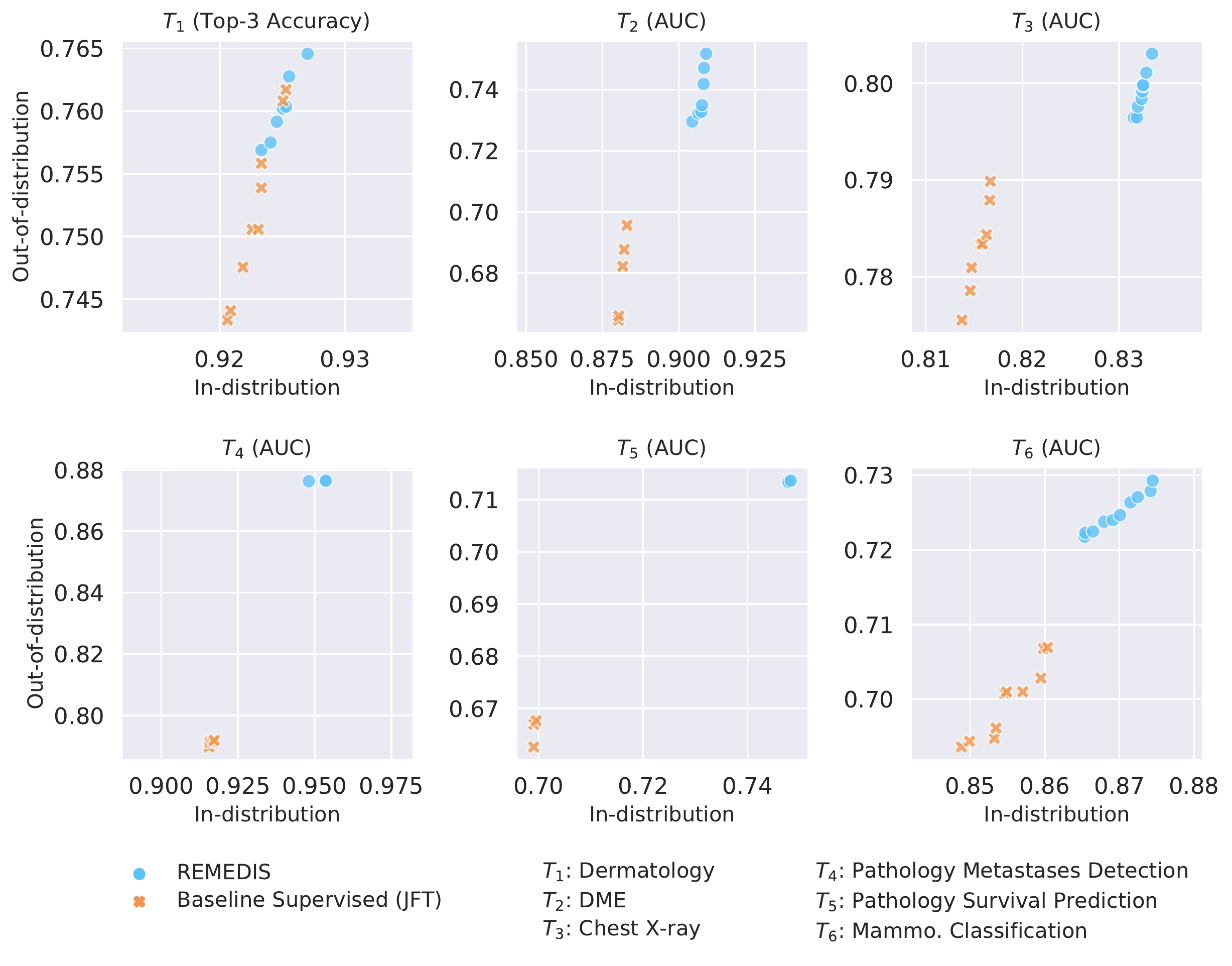}
    \vspace{6pt}
    \caption{\small{\textcolor{black}{\textbf{In-distribution Performance vs. Zero-shot Out-of-distribution Performance.}} We show that REMEDIS produces a consistently superior model performance in comparison to the strong supervised baseline trained using JFT-300M both in- and out-of-distribution. 95\% confidence intervals of our experiments were calculated by running each experiment ten times. Each point in this plot corresponds to one of these repeated runs and its coordinates are obtained by calculating the in-distribution and zero-shot out-of-distribution for the target task. These plots suggest REMEDIS improves out-of-distribution performance without decreasing in-distribution performance and REMEDIS models have higher in-distribution and out-of-distribution performance.}}
    \label{fig:appendix-in-vs-ood}
    \vspace{-5pt}
\end{figure*}

\begin{figure*}[!tbh]
    \centering
    \includegraphics[width=0.95\textwidth]{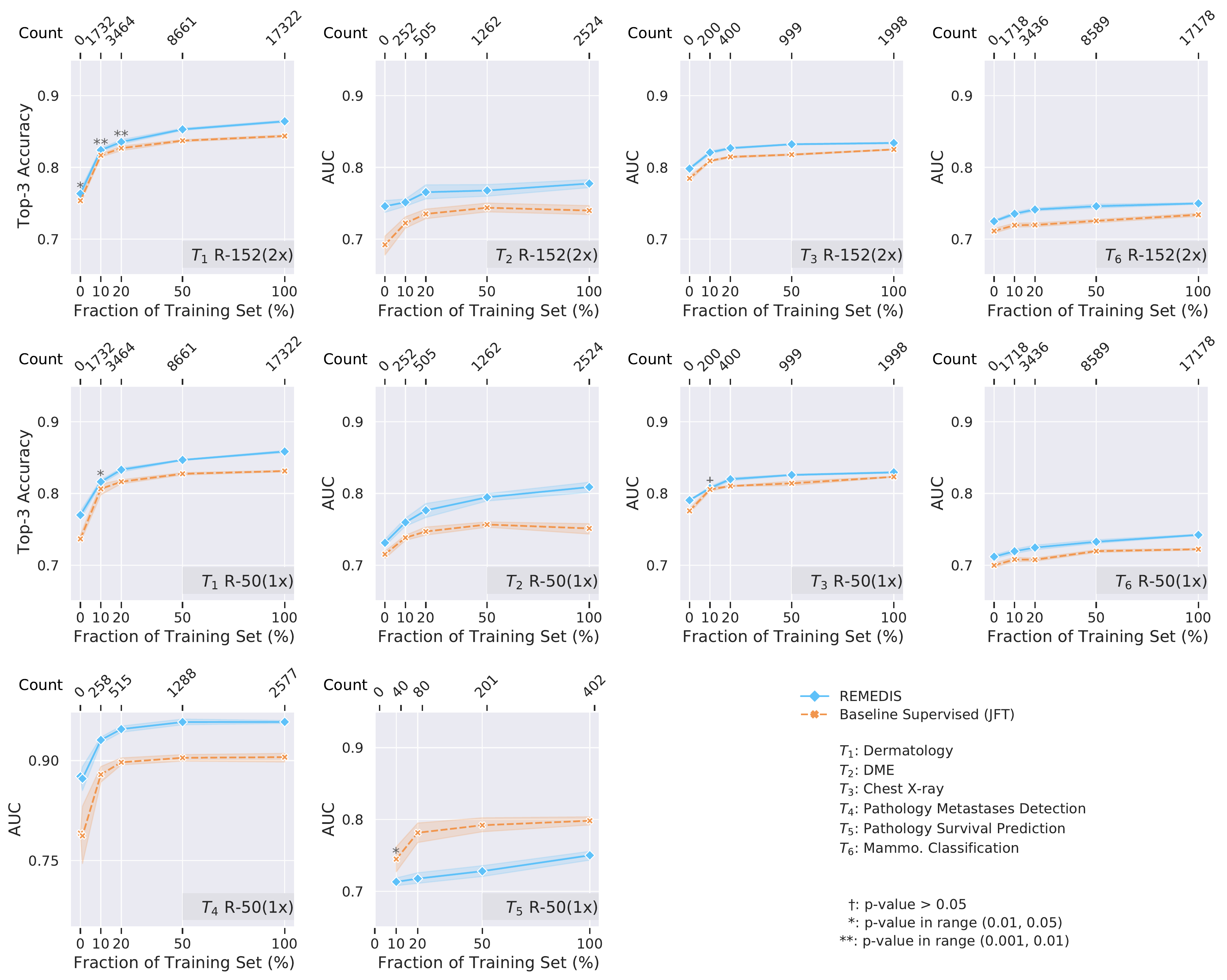}
    \vspace{6pt}
    \caption{\small{\textcolor{black}{\textbf{Detailed generalization results for REMEDIS \vs strong supervised baseline}}. Overview of the results demonstrating data-efficient generalization of our method \vs the strong supervised baseline pretrained on JFT-300M for all tasks and architectures. This includes the dermatology condition classification ($T_1$), diabetic macular edema classification ($T_2$), chest X-ray condition classification ($T_3$), pathology metastases detection ($T_4$), pathology colorectal survival prediction ($T_5$), and mammography classification ($T_6$) as well as two architectures ResNet-50 (1$\times$) and ResNet-152 (2$\times$). In particular, we observe significantly improved out-of-distribution generalization and a significant reduction in the need for labeled medical data when using our proposed approach. 95\% confidence intervals were calculated by running each label fraction and experiment ten times and intervals are shown using the shaded area and error bars. A two-sided $t$-test was also done for each label fraction as well as in-distribution results. If no * is shown, the $p$-value is less than 0.001, otherwise, the $p$-value is as indicated. Unlike previous visualizations, here we scale all of the graphs using a unified range and group the results based on the base network architecture.}}
    \label{fig:appendix-label-eff-detailed-jft}
    \vspace{-0cm}
\end{figure*}

\begin{figure*}[!tbh]
    \centering
    \includegraphics[width=0.95\textwidth]{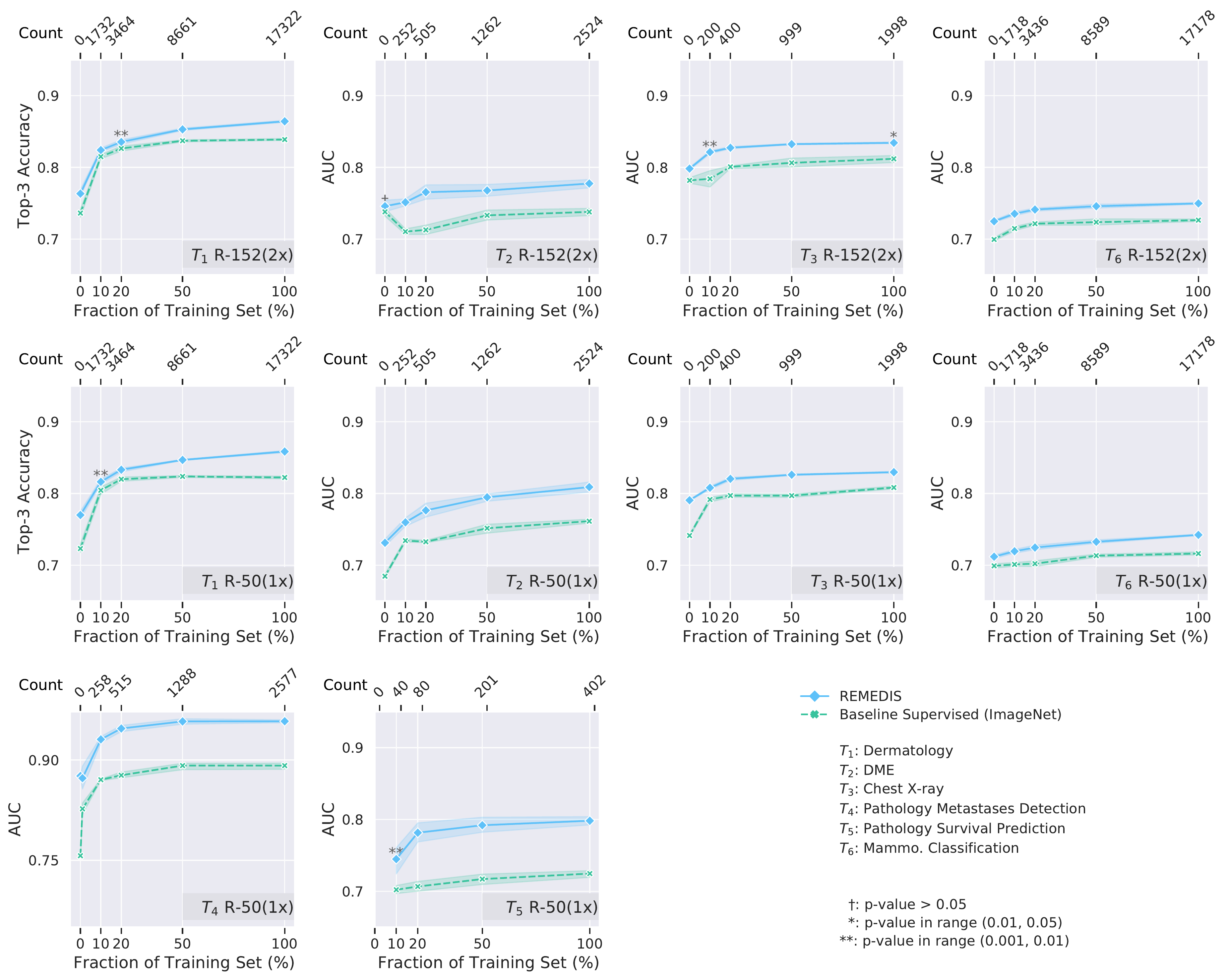}
    \vspace{6pt}
    \caption{\small{\textcolor{black}{\textbf{Detailed generalization results for REMEDIS \vs standard supervised baseline}}. Overview of the results demonstrating data-efficient generalization of our method \vs the standard supervised baseline pretrained on ImageNet-1K for all tasks and architectures. This includes the dermatology condition classification ($T_1$), diabetic macular edema classification ($T_2$), chest X-ray condition classification ($T_3$), pathology metastases detection ($T_4$), pathology colorectal survival prediction ($T_5$), and mammography classification ($T_6$) as well as two architectures ResNet-50 (1$\times$) and ResNet-152 (2$\times$). In particular, we observe significantly improved out-of-distribution generalization and a significant reduction in the need for labeled medical data when using our proposed approach. 95\% confidence intervals were calculated by running each label fraction and experiment ten times and intervals are shown using the shaded area and error bars. A two-sided $t$-test was also done for each label fraction as well as in-distribution results. If no * is shown, the $p$-value is less than 0.001, otherwise, the $p$-value is as indicated. Unlike previous visualizations, here we scale all of the graphs using a unified range and group the results based on the base network architecture.}} 
    \label{fig:appendix-label-eff-detailed-img}
    \vspace{-0cm}
\end{figure*}

The following figures show additional experimental results that compare in-distribution and out-of-distribution performance of REMEDIS \vs the supervised baseline in further detail. Specifically, in this section we investigate the performance of models for both architectures ResNet-50 (1$\times$) and ResNet-152 (2$\times$), and multiple additional out-of-distribution datasets for certain tasks. This section provides followings supplemental figures:

\begin{itemize}[leftmargin=1.5em,rightmargin=0em]
    \item  Figure~\ref{fig:appendix-zero-shot-detailed-jft} provides detailed in-distribution and zero-shot out-of-distribution performance for all datasets grouped by network architectures and compares REMEDIS \vs the strong supervised baseline trained on JFT-300M dataset.
    \item  Figure~\ref{fig:appendix-zero-shot-detailed-img} provides detailed in-distribution and zero-shot out-of-distribution performance for all datasets grouped by network architectures and compares REMEDIS \vs the standard supervised baseline trained on ImageNet-1K dataset.
    
    \item  Figure~\ref{fig:appendix-label-eff-detailed-jft} provides detailed data-efficient generalization results using a common axes range for visualization and grouped by network architecture and compares REMEDIS \vs the strong supervised baseline trained on JFT-300M dataset. 
    \item  Figure~\ref{fig:appendix-label-eff-detailed-img} provides detailed data-efficient generalization results using a common axes range for visualization and grouped by network architecture and compares REMEDIS \vs the standard supervised baseline trained on ImageNet-1K dataset.     
    
    \item Tables~\ref{tab:performance-table-best-vs-best}-\ref{tab:relative-out-of-distribution-img} provide detailed performance values for zero-shot out-of-distribution gains using REMEDIS and the supervised baseline.
    \item Tables~\ref{tab:appendix-ablations-stats-all-task-zero-shot}~--~\ref{tab:appendix-ood-stats-img} list detailed $t$-test statistics for all experiments including REMEDIS, baselines, and multiple ablations studies. 
    
    \item  Figure~\ref{fig:appendix-in-vs-ood} shows a breakdown of in-distribution gains \vs zero-shot out-of-distribution gains using REMEDIS and the supervised baseline.
\end{itemize}

Specifically, Fig.~\ref{fig:appendix-zero-shot-detailed-jft} and Fig.~\ref{fig:appendix-zero-shot-detailed-img} provides in-distribution performance gains \vs zero-shot out-of-distribution performance gains using REMEDIS and supervised baselines. Unlike previous visualizations in Figure~\ref{fig:main-results}, the results are grouped based on the base network architecture, not the best overall performing model for each task. In all plots, the 95\% confidence intervals were calculated by running each experiment ten times, and are shown using the error bars.

\vspace{6pt}
In addition, we also provide additional zero-shot out-of-distribution results for multiple tasks using additional out-of-distribution datasets in these figures. This includes: (1) an additional out-of-distribution dataset for diabetic macular edema classification ($T_2$) which includes 323 de-identified fundus images collected in India, (2) a non-overlapping fraction of the CAMELYON-17 dataset for pathology metastases detection ($T_4$), which includes 273 pathology slides that do not appear in CAMELYON-16 pathology, or the original CAMELYON-17 dataset. These datasets are considered small-scale and are not suitable for data-efficient generalization evaluations; for example they contain only 2-3 examples at 1\% data fraction which is not enough to capture all possible variability in the corresponding tasks and also not relevant in real world deployment settings. Due to computational limits we were not able to train models using the ResNet-152 (2$\times$) architecture for pathology tasks. These results also suggest that, the ResNet-152 (2$\times$) architecture often leads to the highest performance.

\vspace{6pt} 
Figure~\ref{fig:appendix-in-vs-ood} shows the relationship between in-distribution \vs zero-shot out-of-distribution performance using REMEDIS and the supervised baseline. As discussed, 95\% confidence intervals of our experiments were calculated by running each experiment ten times. Each point in this plot corresponds to one of these repeated runs and the coordinates were obtained by calculating the in-distribution and zero-shot out-of-distribution for the target task. These plots confirm that dataset shift greatly impacts the performance of models when evaluated on the out-of-distribution dataset. However, our results suggest that REMEDIS improves out-of-distribution performance without decreasing in-distribution performance, and REMEDIS has higher performance for both in-distribution and out-of-distribution data.

\vspace{6pt}
Lastly, Fig.~\ref{fig:appendix-label-eff-detailed-jft} and Fig.~\ref{fig:appendix-label-eff-detailed-img} show results demonstrating data-efficient generalization of our method \vs the strong supervised baseline pretrained on JFT-300M and also standard supervised baseline pretrained on ImageNet-1K. Unlike the previous visualizations in Fig.~\ref{fig:main-results} and Fig.~\ref{fig:appendix-main-results-img}, these graphs were scaled based on a unified performance range axes and the results are grouped based on the base network architecture, not the best overall results for a given task. In particular, each graph depicts performance (measured by top-3 accuracy or area under the curve (AUC)) when using different data fractions/data counts of out-of-distribution data to fine-tune the model for the dermatology condition classification ($T_1$), diabetic macular edema classification ($T_2$), chest X-ray condition classification ($T_3$), pathology metastases detection ($T_4$), pathology colorectal survival prediction ($T_5$), and mammography classification ($T_6$) as well as two architectures ResNet-50 (1$\times$) and ResNet-152 (2$\times$). We also calculate a 95\% confidence interval by running each label fraction and experiment ten times and intervals are shown using the shaded area and error bars. A two-sided $t$-test was also calculated for each label fraction as well as in-distribution results and $p$-value for several thresholds comparing any significant improvement of our method against these baselines. We observe significantly ($p<0.05$) improved out-of-distribution generalization and a significant reduction in the need for labeled medical data when using REMEDIS.

\begin{table}[tbh!]
\centering
\caption{\small{\textcolor{black}{\textbf{Detailed in-distribution results.}} The table contains the numeric results displayed in Fig.~\ref{fig:main-results} and Fig.~\ref{fig:appendix-main-results-img}, specifically the average in-distribution performance values, with 95\% confidence intervals in parentheses.}}
\centering
\vspace{+3pt}
\label{tab:performance-table-best-vs-best}
\footnotesize
\begin{tabular}{l|l|c}
\toprule
\rowcolor{ourlightgray}
          Task and Metric &                                 Method &               Metric \\
\midrule
  Task 1 (Top-3 Accuracy) &         Baseline Supervised (ImageNet) &    0.900 (0.897,0.903) \\ [3pt]
                          &              Baseline Supervised (JFT) &    0.923 (0.922,0.925) \\ [3pt]  
                          &                                REMEDIS &    0.926 (0.925,0.928) \\ [3pt]\midrule
             Task 2 (AUC) &         Baseline Supervised (ImageNet) &    0.887 (0.886,0.887) \\ [3pt] 
                          &              Baseline Supervised (JFT) &    0.883 (0.880,0.886) \\ [3pt] 
                          &                                REMEDIS &    0.902 (0.900,0.902) \\ [3pt]\midrule
             Task 3 (AUC) &         Baseline Supervised (ImageNet) &    0.818 (0.818,0.819) \\
\bottomrule
\end{tabular}
\end{table}

\begin{table}[tbh!]
\centering
\caption{\small{\textcolor{black}{\textbf{In-distribution improvement between REMEDIS and the strong supervised baseline.}} The absolute and relative improvement in the main task metrics between REMEDIS and the strong supervised baseline (JFT) for the in-distribution dataset. This data is represented in Fig.~\ref{fig:main-results}.}}
\centering
\vspace{+3pt}
\label{tab:relative-in-distribution-jft}
\footnotesize

\begin{tabular}{lcc}
\toprule
\rowcolor{ourlightgray}
   Task (Metric) &  Absolute Improvement & Relative Improvement (\%) \\
\midrule
 Task 1 (Top-3 Accuracy)    &  0.003 (0.000, 0.005) &       0.3 (0.0, 0.6) \\
 Task 2 (AUC)               &  0.024 (0.018, 0.028) &       2.7 (2.1, 3.2) \\
 Task 3 (AUC)               &  0.017 (0.016, 0.018) &       2.1 (1.9, 2.3) \\
 Task 4 (AUC)               &  0.038 (0.032, 0.044) &       4.1 (3.5, 4.8) \\
 Task 5 (AUC)               &  0.049 (0.047, 0.051) &       7.0 (6.8, 7.2) \\
 Task 6 (AUC)               &  0.001 (-0.005, 0.006) &       0.1 (-0.5, 0.7) \\
\bottomrule
\end{tabular}
\end{table}

\begin{table}[tbh!]
\centering
\caption{\small{\textcolor{black}{\textbf{In-distribution improvement between REMEDIS and the standard supervised baseline.}} The absolute and relative improvement in the main task metrics between REMEDIS and the standard supervised baseline (ImageNet) for the in-distribution dataset. This data is represented in Fig.~\ref{fig:appendix-main-results-img}.}}
\centering
\vspace{+3pt}
\label{tab:relative-in-distribution-img}
\footnotesize

\begin{tabular}{lcc}
\toprule
\rowcolor{ourlightgray}
 Task (Metric) &   Absolute Improvement & Relative Improvement (\%)          \\\midrule
 Task 1 (Top-3 Accuracy)    &  0.026 (0.022, 0.03) &        2.9 (2.4, 3.4) \\
 Task 2 (AUC)               &  0.015 (0.013, 0.017) &       1.7 (1.5, 1.9) \\
 Task 3 (AUC)               &  0.014 (0.013, 0.016) &       1.8 (1.6, 1.9) \\
 Task 4 (AUC)               &  0.098 (0.085, 0.109) &     11.5 (9.9, 12.8) \\
 Task 5 (AUC)               &  0.034 (0.031, 0.036) &       4.7 (4.4, 5.1) \\
 Task 6 (AUC)               &  0.018 (0.012, 0.024) &       2.1 (1.4, 2.8) \\
\bottomrule
\end{tabular}
\end{table}

\begin{table}[!tbh]
\centering
\caption{\small{\textcolor{black}{\textbf{Out-of-distribution data efficiency metrics \vs Clinician.}} This table contains the percentage and absolute numbers of the training set necessary for REMEDIS to match the strong supervised baseline (JFT) performance, as shown in Fig.~\ref{fig:main-results}, as well as the estimated clinician hours saved.}}
\centering
\vspace{+3pt}
\label{tab:cost-saved-table-best-vs-best-jft}
\footnotesize
\begin{tabular}{l|c|c|c}
\toprule
\rowcolor{ourlightgray}
          Task &               Proportion required to meet baseline performance &   Count of samples saved &      Estimated clinician hours saved \\
\midrule
  Task 1 &                    33.2\%  (25.7\%, 39.3\%)        &    11,578 (10510, 12862)                      & 193 (175, 214)        \\ [3pt]
  Task 2 &                    7.2\%  (4.0\% , 10.3\%)         &    2,342 (2263, 2423)                         & 224 (217, 232)        \\ [3pt] 
  Task 3 &                    16.8\% (12.0\%, 22.8\%)         &    23,278 (21588, 24620)                      & 789 (732, 834)        \\ [3pt]
  Task 4 &                    5.7\%  (2.3\%  7.3\%)           &    16,872 (16596, 17482)                      & 2,812 (2766, 2914)    \\ [3pt]
  Task 5 &                    14.1\% (9.0\% , 18.8\% )        &    3,325 (3145, 3522)                         & 554 (524, 587)        \\ [3pt]
  Task 6 &                    8.7\% (4.6\%, 15.1\%)           &    15,689 (14575, 16390)                      & 1,569 (1457, 1639)    \\ [3pt]
\bottomrule
\end{tabular}
\end{table}

\begin{table}[!tbh]
\centering
\caption{\small{\textcolor{black}{\textbf{Out-of-distribution data efficiency metrics \vs Clinician.}} This table contains the percentage and absolute numbers of the training set necessary for REMEDIS to match the standard supervised (ImageNet) performance, as shown in Fig.~\ref{fig:appendix-main-results-img}, as well as the estimated clinician hours saved.}}
\centering
\vspace{+3pt}
\label{tab:cost-saved-table-best-vs-best-img}
\footnotesize
\begin{tabular}{l|c|c|c}
\toprule
\rowcolor{ourlightgray}
          Task &               Proportion required to meet baseline performance &   Count of samples saved &      Estimated clinician hours saved \\
\midrule
  Task 1 &                    30.7\% (25.14\%, 36.9\%)        &    11,990 (10920, 12967)                      & 200 (182, 216)        \\ [3pt]
  Task 2 &                     8.9\%  (6.6\%, 12.9\%)         &    2,297  (2196, 3356)                        & 220 (210, 226)        \\ [3pt] 
  Task 3 &                     5.9\% (2.8\%, 8.9\%)           &    26,315 (25467, 25467)                      & 219 (212, 226)        \\ [3pt]
  Task 4 &                    3.5\% (0.6\%, 4.5\%)            &    17,269 (17095, 17788)                      & 2,878 (2849, 2965)    \\ [3pt]
  Task 5 &                     3.7\%  (1.2\%, 10.7\%)         &    3,727  (3460, 3826)                        & 621 (577, 638)        \\ [3pt] 
  Task 6 &                     1.4\% (0.0\%, 4.9\%)           &    16,924 (16323, 17178)                      & 1,692 (1632, 1718)    \\ [3pt]
\bottomrule
\end{tabular}
\end{table}

\begin{table}[!tbh]
\centering
\caption{\small{\textcolor{black}{\textbf{Out-of-distribution Best-vs-Best Metrics - Part 1}} The table contains the exact metrics for Fig.~\ref{fig:main-results} and Fig.~\ref{fig:appendix-main-results-img}, specifically the out-of-distribution data efficiency metrics values for REMEDIS \vs the strong and standard supervised baselines for tasks $T_1$, $T_2$, $T_3$.}}
\centering
\vspace{+3pt}
\label{tab:performance-table-data-efficiency-1}
\footnotesize
\renewcommand{\arraystretch}{1.0}
\begin{tabular}{l|l|cc}
\toprule
\rowcolor{ourlightgray}
             Task (Metric)&               Method &  Percentage &               Metric \\\midrule
  Task 1 (Top-3 Accuracy) &  Baseline Supervised &           0 &  0.738 (0.734,0.743) \\
                          &           (ImageNet) &          10 &  0.816 (0.814,0.819) \\
                          &                      &          20 &  0.826 (0.823,0.830) \\
                          &                      &          50 &  0.837 (0.836,0.838) \\
                          &                      &         100 &  0.839 (0.838,0.840) \\  \cmidrule{2-4}
                          &  Baseline Supervised &           0 &  0.755 (0.750,0.760) \\
                          &                (JFT) &          10 &  0.817 (0.814,0.819) \\
\bottomrule
\end{tabular}
\end{table}

\begin{table}[!tbh]
\centering
\caption{\small{\textcolor{black}{\textbf{Out-of-distribution Best-vs-Best Metrics - Part 2}} The table contains the exact metrics for Fig.~\ref{fig:main-results} and Fig.~\ref{fig:appendix-main-results-img}, specifically the out-of-distribution data efficiency metrics values for REMEDIS \vs the strong and standard supervised baselines for tasks $T_4$, $T_5$, $T_6$.}}
\centering
\vspace{+3pt}
\label{tab:performance-table-data-efficiency-2}
\footnotesize
\renewcommand{\arraystretch}{1.0}
\begin{tabular}{l|l|cc}
\toprule
\rowcolor{ourlightgray}
            Task (Metric) &               Method &  Percentage &               Metric \\   \midrule
            Task 4 (AUC)  &  Baseline Supervised &           0 &  0.757 (0.755,0.758) \\
                          &           (ImageNet) &          10 &  0.870 (0.869,0.872) \\
                          &                      &          20 &  0.877 (0.873,0.882) \\
                          &                      &          50 &  0.892 (0.885,0.895) \\
                          &                      &         100 &  0.892 (0.886,0.895) \\  \cmidrule{2-4}
                          &  Baseline Supervised &           0 &  0.791 (0.790,0.792) \\
                          &                (JFT) &          10 &  0.879 (0.868,0.891) \\
\bottomrule
\end{tabular}
\end{table}

\begin{table}[!tbh]
\centering
\caption{\small{\textcolor{black}{\textbf{Out-of-distribution relative improvement between REMEDIS and the strong supervised baseline.}} The absolute and relative improvement in the metrics between REMEDIS and the strong supervised baseline (JFT), using 0\% and 100\% of the Out-of-distribution dataset. This data is represented in Figure~\ref{fig:main-results}.}}
\centering
\vspace{+3pt}
\label{tab:relative-out-of-distribution-jft}
\footnotesize

\begin{tabular}{lc|cc}
\toprule
\rowcolor{ourlightgray}
   Task (Metric)&           Percentage &     Absolute Improvement&       Relative Improvement (\%) \\
\midrule
 Task 1 (Top-3 Accuracy) &           0 &      0.009 (0.000, 0.018) &  1.2 (0.0, 2.5)   \\
              &                    100 &      0.020 (0.017, 0.023) &  2.4 (2.1, 2.8)   \\ \midrule
 Task 2 (AUC) &                      0 &      0.014 (0.007, 0.021) &  1.9 (0.9, 3.0)   \\
              &                    100 &      0.060 (0.050, 0.071) &  8.0 (6.6, 9.5)   \\ \midrule
 Task 3 (AUC) &                      0 &      0.014 (0.008, 0.019) &  1.7 (1.1, 2.4)   \\
              &                    100 &      0.009 (0.007, 0.011) &  1.1 (0.9, 1.3)   \\ \midrule
 Task 4 (AUC) &                      0 &      0.085 (0.084, 0.086) & 10.7 (10.6, 10.9) \\
              &                    100 &      0.053 (0.045, 0.063) &  5.8 (5.0, 7.0)   \\ \midrule
 Task 5 (AUC) &                      0 &      0.048 (0.043, 0.052) &  7.2 (6.5, 7.9)   \\
              &                    100 &      0.038 (0.029, 0.047) &  5.0 (3.8, 6.2)   \\ \midrule
 Task 6 (AUC) &                      0 &      0.014 (0.009, 0.018) &  1.9 (1.2, 2.5)   \\
              &                    100 &      0.016 (0.012, 0.019) &  2.2 (1.7, 2.6)   \\
\bottomrule
\end{tabular}

\end{table}

\begin{table}[!tbh]
\centering
\caption{\small{\textcolor{black}{\textbf{Out-of-distribution relative improvement between REMEDIS and the standard supervised baseline.}} The absolute and relative improvement in the metrics between REMEDIS and the standard supervised baseline (ImageNet), using 0\% and 100\% of the Out-of-distribution dataset. This data is represented in Figure~\ref{fig:appendix-main-results-img}.}}
\centering
\vspace{+3pt}
\label{tab:relative-out-of-distribution-img}
\footnotesize

\begin{tabular}{lc|cc}
\toprule
\rowcolor{ourlightgray}
   Task (Metric)&           Percentage &     Absolute Improvement&       Relative Improvement (\%) \\
\midrule
 Task 1 (Top-3 Accuracy) &           0 &  0.025 (0.016, 0.035) &       3.4 (2.2, 4.7) \\
              &                    100 &  0.025 (0.022, 0.028) &       3.0 (2.7, 3.4) \\ \midrule
 Task 2 (AUC) &                      0 &  0.047 (0.039, 0.054) &       6.8 (5.7, 7.9) \\
              &                    100 &  0.054 (0.046, 0.062) &       7.1 (6.1, 8.2) \\ \midrule
 Task 3 (AUC) &                      0 &  0.013 (0.009, 0.017) &       1.6 (1.1, 2.2) \\
              &                    100 &  0.022 (0.017, 0.028) &       2.7 (2.0, 3.5) \\ \midrule
 Task 4 (AUC) &                      0 &  0.119 (0.117, 0.121) &    15.8 (15.5, 16.1) \\
              &                    100 &  0.066 (0.061, 0.074) &       7.4 (6.8, 8.4) \\ \midrule
 Task 5 (AUC) &                      0 &  0.063 (0.055, 0.068) &      9.8 (8.4, 10.5) \\
              &                    100 &  0.074 (0.064, 0.084) &     10.2 (8.8, 11.7) \\ \midrule
 Task 6 (AUC) &                      0 &  0.025 (0.021, 0.029) &       3.6 (3.0, 4.2) \\
              &                    100 &  0.023 (0.02, 0.026) &       3.2 (2.8, 3.6) \\
\bottomrule
\end{tabular}

\end{table}

\begin{table}[!tbh]
\centering
\caption{\small{\textbf{Ablation Study Statistics} The table contains the corresponding two-sided $t$-test statistics for Fig.~\ref{fig:appendix-ablations}, specifically the metrics produced for REMEDIS \vs each of the different techniques. This $t$-test was done without the assumption that the variances are equal.}}
\centering
\vspace{+3pt}
\label{tab:appendix-ablations-stats-all-task-zero-shot}
\footnotesize

\begin{tabular}{llll|ccc}
\toprule
\rowcolor{ourlightgray}
   Task &         Distribution &          Architecture &         Comparison Method &  T-Statistic &       $p$-value &  Degrees of Freedom \\
\midrule
 Task 1 &      In-distribution &             R-152(2x) &                     BiT-L &         1.83 &  8.848414e-02 &               14.06 \\
 Task 1 &      In-distribution &             R-152(2x) &                    SimCLR &        10.29 &  6.815352e-09 &               17.69 \\
 Task 1 &      In-distribution &              R-50(1x) &                     BiT-L &         7.22 &  2.571443e-06 &               15.40 \\
 Task 1 &      In-distribution &              R-50(1x) &                    SimCLR &         5.36 &  7.154525e-05 &               15.46 \\
 Task 1 &  Out-of-distribution &             R-152(2x) &                     BiT-L &         1.54 &  1.411247e-01 &               17.58 \\
 Task 1 &  Out-of-distribution &             R-152(2x) &                    SimCLR &         6.95 &  2.933571e-06 &               16.34 \\
 Task 1 &  Out-of-distribution &              R-50(1x) &                     BiT-L &         5.59 &  5.035536e-05 &               15.06 \\
\bottomrule
\end{tabular}
\end{table}

\begin{table}[!tbh]
\centering
\caption{\small{\textcolor{black}{\textbf{Task Statistics REMEDIS \vs the strong supervised baseline (JFT).}} The table contains the corresponding two-sided $t$-test statistics for Fig.~\ref{fig:appendix-zero-shot-detailed-jft}, specifically the metrics produced for REMEDIS \vs the strong supervised baseline (JFT). This $t$-test was done without the assumption that the variances are equal.}}
\centering
\vspace{+3pt}
\label{tab:appendix-zeroshot-detailed-stats-jft}
\footnotesize
\begin{tabular}{lll|ccc}
\toprule
\rowcolor{ourlightgray}
   Task &         Distribution &           Architecture &  T-Statistic &       $p$-value &  Degrees of Freedom \\
\midrule
 Task 1 &      In-distribution &              R-152(2x) &         2.74 &  1.333964e-02 &               17.97 \\
 Task 1 &      In-distribution &               R-50(1x) &         6.53 &  1.478805e-05 &               13.71 \\
 Task 1 &  Out-of-distribution &              R-152(2x) &         2.76 &  1.331786e-02 &               16.98 \\
 Task 1 &  Out-of-distribution &               R-50(1x) &         7.97 &  2.626604e-07 &               17.98 \\
 Task 2 &      In-distribution &              R-152(2x) &        11.91 &  8.600512e-08 &               11.42 \\
 Task 2 &      In-distribution &               R-50(1x) &         9.91 &  8.610633e-08 &               14.31 \\
 Task 2 &  Out-of-distribution &  Dataset 1 (R-152(2x)) &         6.27 &  4.792115e-05 &               11.61 \\
\bottomrule
\end{tabular}
\end{table}

\begin{table}[!tbh]
\centering
\caption{\small{\textcolor{black}{\textbf{Task Statistics REMEDIS \vs the standard supervised baseline (ImageNet).}} The table contains the corresponding two-sided $t$-test statistics for Fig.~\ref{fig:appendix-zero-shot-detailed-img}, specifically the metrics produced for REMEDIS \vs the standard supervised baseline (ImageNet). This $t$-test was done without the assumption that the variances are equal.}}
\centering
\vspace{+3pt}
\label{tab:appendix-zeroshot-detailed-stats-img}
\footnotesize
\begin{tabular}{lll|ccc}
\toprule
\rowcolor{ourlightgray}
   Task &         Distribution &           Architecture &  T-Statistic &       $p$-value &  Degrees of Freedom \\
\midrule
 Task 1 &      In-distribution &              R-152(2x) &        14.60 &  4.509936e-09 &               12.16 \\
 Task 1 &      In-distribution &               R-50(1x) &        12.70 &  2.120881e-10 &               17.93 \\
 Task 1 &  Out-of-distribution &              R-152(2x) &         6.89 &  2.967478e-06 &               16.64 \\
 Task 1 &  Out-of-distribution &               R-50(1x) &        10.51 &  4.458786e-09 &               17.86 \\
 Task 2 &      In-distribution &              R-152(2x) &         5.10 &  8.239890e-05 &               17.48 \\
 Task 2 &      In-distribution &               R-50(1x) &        23.57 &  6.420586e-13 &               14.42 \\
 Task 2 &  Out-of-distribution &  Dataset 1 (R-152(2x)) &         1.47 &  1.609929e-01 &               16.13 \\
\bottomrule
\end{tabular}
\end{table}

\begin{table}[tbh!]
\centering
\caption{\small{\textcolor{black}{\textbf{Self-Training Statistics}} The table contains the corresponding two-sided $t$-test statistics for Fig.~\ref{fig:appendix-self-training}, specifically the metrics produced for REMEDIS \vs strong supervised baseline (JFT) and standard supervised baseline (ImageNet) further improved using the self-training strategy. This $t$-test was done without the assumption that the variances are equal.}}
\centering
\vspace{+3pt}
\label{tab:appendix-self-training-stats}
\footnotesize
\renewcommand{\arraystretch}{1.25}
\begin{tabular}{lll|ccc}
\toprule
\rowcolor{ourlightgray}
        Distribution & Architecture &                    Comparison Method &  T-Statistic &       $p$-value &  Degrees of Freedom \\
\midrule
     In-distribution &    R-152(2x) &  Baseline (ImageNet) + Self-training &        20.11 &  3.165953e-12 &               14.93 \\
     In-distribution &     R-50(1x) &  Baseline (ImageNet) + Self-training &        26.10 &  1.893045e-15 &               17.48 \\ \midrule
     In-distribution &    R-152(2x) &       Baseline (JFT) + Self-training &         3.83 &  1.242247e-03 &               17.74 \\
     In-distribution &     R-50(1x) &       Baseline (JFT) + self-training &         9.22 &  2.136911e-07 &               14.30 \\ \midrule
 Out-of-distribution &    R-152(2x) &  Baseline (ImageNet) + Self-training &        11.24 &  1.445689e-09 &               17.98 \\
 Out-of-distribution &     R-50(1x) &  Baseline (ImageNet) + Self-training &        23.11 &  7.005811e-12 &               12.90 \\ \midrule
 Out-of-distribution &    R-152(2x) &       Baseline (JFT) + Self-training &         2.54 &  2.067761e-02 &               17.92 \\
 Out-of-distribution &     R-50(1x) &       Baseline (JFT) + Self-training &         9.00 &  3.058183e-07 &               14.19 \\
\bottomrule
\end{tabular}
\end{table}

\begin{table}[tbh!]
\centering
\caption{\small{\textcolor{black}{\textbf{Data Efficiency Statistics.}} The table contains the corresponding two-sided $t$-test statistics for Fig.~\ref{fig:appendix-label-eff-detailed-jft}, specifically the metrics produced for REMEDIS \vs the Supervised Baseline (JFT). This $t$-test was done without the assumption that the variances are equal.}}
\centering
\vspace{+3pt}
\label{tab:appendix-data-efficiency-stats}
\footnotesize

\begin{tabular}{llr|ccc}
\toprule
\rowcolor{ourlightgray}
 Task &   Architecture &  Data Percentage &  T-Statistic &       $p$-value &  Degrees of Freedom \\ \midrule
 Task 1 &    R-152(2x) &                0 &         2.76 &  1.331786e-02 &               16.98 \\
 Task 1 &    R-152(2x) &               10 &         3.80 &  1.334593e-03 &               17.76 \\
 Task 1 &    R-152(2x) &               20 &         3.63 &  1.933656e-03 &               17.92 \\
 Task 1 &    R-152(2x) &               50 &        12.77 &  2.331362e-10 &               17.68 \\
 Task 1 &    R-152(2x) &              100 &        17.72 &  7.793245e-13 &               17.99 \\
 Task 1 &     R-50(1x) &                0 &         7.97 &  2.626604e-07 &               17.98 \\
 Task 1 &     R-50(1x) &               10 &         2.18 &  4.994976e-02 &               11.99 \\
\bottomrule
\end{tabular}
\end{table}

\begin{table}[tbh!]
\centering
\caption{\small{\textcolor{black}{\textbf{Data Efficiency Statistics.}} The table contains the corresponding two-sided $t$-test statistics for Fig.~\ref{fig:appendix-label-eff-detailed-img}, specifically the metrics produced for REMEDIS \vs the Supervised Baseline (ImageNet). This $t$-test was done without the assumption that the variances are equal.}}
\centering
\vspace{+3pt}
\label{tab:appendix-data-efficiency-stats-img}
\footnotesize

\begin{tabular}{llr|ccc}
\toprule
\rowcolor{ourlightgray}
 Task & Architecture &  Data Percentage &  T-Statistic &       $p$-value &  Degrees of Freedom   \\\midrule
 Task 1 &    R-152(2x) &                0 &         6.89 &  2.967478e-06 &               16.64 \\
 Task 1 &    R-152(2x) &               10 &         4.24 &  5.149853e-04 &               17.69 \\
 Task 1 &    R-152(2x) &               20 &         3.65 &  1.907543e-03 &               17.61 \\
 Task 1 &    R-152(2x) &               50 &        13.12 &  1.832681e-10 &               17.43 \\
 Task 1 &    R-152(2x) &              100 &        23.52 &  7.950471e-15 &               17.75 \\ 
 Task 1 &     R-50(1x) &                0 &        10.51 &  4.458786e-09 &               17.86 \\
 Task 1 &     R-50(1x) &               10 &         3.58 &  2.722524e-03 &               14.97 \\
\bottomrule
\end{tabular}
\end{table}

\begin{table}[!tbh]
\centering
\caption{\small{\textcolor{black}{\textbf{In-distribution Best-vs-Best Statistics.}} The table contains the corresponding two-sided $t$-test statistics for Fig.~\ref{fig:main-results}, specifically the metrics produced for REMEDIS \vs the Baseline Supervised (JFT) for in-distribution. This $t$-test was done without the assumption that the variances are equal.}}
\centering
\vspace{+3pt}
\label{tab:appendix-in-dist-stats-jft}
\footnotesize

\begin{tabular}{l|ccc}
\toprule
\rowcolor{ourlightgray}
   Task &  T-Statistic &       $p$-value &  Degrees of Freedom \\\midrule
 Task 1 &         2.74 &  1.333964e-02 &               17.97 \\
 Task 2 &        11.91 &  8.600512e-08 &               11.42 \\
 Task 3 &        31.20 &  2.842220e-14 &               13.91 \\
 Task 4 &        12.74 &  8.643737e-04 &                3.11 \\
 Task 5 &        71.75 &  1.512674e-09 &                5.62 \\
 Task 6 &         8.71 &  8.879738e-08 &               17.53 \\
\bottomrule
\end{tabular}
\end{table}

\begin{table}[!tbh]
\centering
\caption{\small{\textcolor{black}{\textbf{In-distribution Best-vs-Best Statistics.}} The table contains the corresponding two-sided $t$-test statistics for Fig.~\ref{fig:main-results}, specifically the metrics produced for REMEDIS \vs the Baseline Supervised (ImageNet) for in-distribution. This $t$-test was done without the assumption that the variances are equal.}}
\centering
\vspace{+3pt}
\label{tab:appendix-in-dist-stats-img}
\footnotesize

\begin{tabular}{l|ccc}
\toprule
\rowcolor{ourlightgray}
   Task &  T-Statistic &       $p$-value &  Degrees of Freedom \\\midrule
 Task 1 &         9.56 &  5.179113e-07 &               12.18 \\
 Task 2 &         5.10 &  8.239890e-05 &               17.48 \\
 Task 3 &        33.52 &  7.022141e-17 &               16.87 \\
 Task 4 &        19.97 &  2.152548e-06 &                5.56 \\
 Task 5 &        34.87 &  2.018287e-06 &                4.28 \\
 Task 6 &         9.31 &  1.612389e-07 &               14.57 \\
\bottomrule
\end{tabular}
\end{table}

\begin{table}[tbh!]
\centering
\caption{\small{\textcolor{black}{\textbf{Out-of-distribution Best-vs-Best Statistics.}} The table contains the corresponding two-sided $t$-test statistics for Fig.~\ref{fig:main-results}, specifically the metrics produced for REMEDIS \vs the Baseline Supervised (JFT) for out-of-distribution. This $t$-test was done without the assumption that the variances are equal.}}
\centering
\vspace{+3pt}
\label{tab:appendix-ood-stats-jft}
\footnotesize

\begin{tabular}{lr|ccc}
\toprule
\rowcolor{ourlightgray}
   Task &  Percentage &  T-Statistic &       $p$-value &  Degrees of Freedom \\
\midrule
 Task 1 &           0 &        2.44 &  2.661145e-02 &              15.99 \\
 Task 1 &          10 &        3.80 &  1.334593e-03 &              17.76 \\
 Task 1 &          20 &        4.29 &  5.605730e-04 &              15.99 \\
 Task 1 &          50 &       12.77 &  2.331362e-10 &              17.68 \\
 Task 1 &         100 &       17.72 &  7.793245e-13 &              17.99 \\ \midrule
 Task 2 &           0 &        4.60 &  4.244625e-04 &              13.83 \\
 Task 2 &          10 &        8.47 &  1.155962e-05 &               9.27 \\
\bottomrule
\end{tabular}
\end{table}

\begin{table}[tbh!]
\centering
\caption{\small{\textcolor{black}{\textbf{Out-of-distribution Best-vs-Best Statistics.}} The table contains the corresponding two-sided $t$-test statistics for Fig.~\ref{fig:main-results}, specifically the metrics produced for REMEDIS \vs the Baseline Supervised (ImageNet) for out-of-distribution. This $t$-test was done without the assumption that the variances are equal.}}
\centering
\vspace{+3pt}
\label{tab:appendix-ood-stats-img}
\footnotesize

\begin{tabular}{lr|ccc}
\toprule
\rowcolor{ourlightgray}
   Task &  Percentage &  T-Statistic &       $p$-value &  Degrees of Freedom \\
\midrule
 Task 1 &           0 &        7.02 &  2.936831e-06 &              15.98 \\
 Task 1 &          10 &        4.20 &  6.026055e-04 &              16.92 \\
 Task 1 &          20 &        4.30 &  5.412103e-04 &              16.15 \\
 Task 1 &          50 &       13.12 &  1.832681e-10 &              17.43 \\
 Task 1 &         100 &       23.52 &  7.950471e-15 &              17.75 \\  \midrule
 Task 2 &           0 &       15.63 &  1.932572e-10 &              14.42 \\
 Task 2 &          10 &       10.63 &  3.205746e-06 &               8.56 \\
\bottomrule
\end{tabular}
\end{table}

\setlength\bibitemsep{0pt}
\printbibliography[heading=subbibliography]
\clearpage
\end{refsection}

\end{document}